\RequirePackage{rotating} 

\documentclass{clv3}

\usepackage{hyperref}
\usepackage{nameref}
\usepackage{xcolor}
\definecolor{darkblue}{rgb}{0, 0, 0.5}
\hypersetup{colorlinks=true,citecolor=darkblue, linkcolor=darkblue, urlcolor=darkblue}

\usepackage{times}
\usepackage{latexsym}
\usepackage{graphicx}
\usepackage{amsmath}
\usepackage{booktabs}
\usepackage{xcolor,colortbl}
\usepackage{hhline}
\usepackage{afterpage}
\usepackage[T1]{fontenc}
\usepackage[utf8]{inputenc}
\usepackage{rotating}
\usepackage{inconsolata}

\usepackage{tabularx}

\usepackage{fmtcount}
\usepackage{CJKutf8}

\usepackage{pifont}
\definecolor{darkpastelgreen}{rgb}{0.01, 0.75, 0.24}
\definecolor{darkpastelred}{rgb}{0.76, 0.23, 0.13}
\newcommand{\cmark}{\textcolor{darkpastelgreen}{\ding{51}}}%
\newcommand{\xmark}{\textcolor{darkpastelred}{\ding{55}}}%

\usepackage{microtype}

\newcommand{\tabitem}{~~\llap{\textbullet}~~}

\GetTitleStringSetup{expand}

\begin{document}

\dochead{}

\runningtitle{Machine Translation Meta Evaluation through Translation Accuracy Challenge Sets}


\title{Machine Translation Meta Evaluation through Translation Accuracy Challenge Sets}




\author{Nikita Moghe}
\affil{School of Informatics, University of Edinburgh\thanks{Corresponding author: nikita.moghe@ed.ac.uk}}
\author{Arnisa Fazla}
\affil{Department of Computational Linguistics, University of Zurich}
\author{Chantal Amrhein}
\affil{Textshuttle, Zurich}
\author{Tom Kocmi}
\affil{Microsoft}
\author{Mark Steedman}
\affil{School of Informatics, University of Edinburgh}
\author{Alexandra Birch}
\affil{School of Informatics, University of Edinburgh}
\author{Rico Sennrich}
\affil{Department of Computational Linguistics, University of Zurich}
\author{Liane Guillou}
\affil{Department of Computer Science, RISE Research Institutes of Sweden}

\maketitle

\begin{abstract}

Recent machine translation (MT) metrics calibrate their effectiveness by correlating with human judgement. However, these results are often obtained by averaging predictions across large test sets without any insights into the strengths and weaknesses of these metrics across different error types. Challenge sets are used to probe specific dimensions of metric behaviour but there are very few such datasets and they either focus on a limited number of phenomena or a limited number of language pairs. 
We introduce \textsc{ACES}, a contrastive challenge set spanning  146 language pairs, aimed at discovering whether metrics can identify 68 translation accuracy errors.
These phenomena range from basic alterations at the word/character level to more intricate errors based on discourse and real-world knowledge. We conducted a large-scale study by benchmarking \textsc{ACES} on 50 metrics submitted to the WMT 2022 and 2023 metrics shared tasks. We benchmark metric performance, assess their incremental performance over successive campaigns, and measure their sensitivity to a range of linguistic phenomena. We also investigate claims that Large Language Models (LLMs) are effective as MT evaluators, addressing the limitations of previous studies by providing a more holistic evaluation that covers a range of linguistic phenomena and language pairs and includes both low- and medium-resource languages. 
Our results demonstrate that different metric families struggle with different phenomena and that LLM-based methods fail to demonstrate reliable performance. Our analyses indicate that most metrics ignore the source sentence, tend to prefer surface-level overlap and end up incorporating properties of base models which are not always beneficial. To further encourage detailed evaluation beyond singular scores, we expand \textsc{ACES} to include error span annotations, denoted as \textsc{SPAN-ACES} and we use this dataset to evaluate span-based error metrics showing these metrics also need considerable improvement. 

Finally, we provide a set of recommendations for building better MT metrics, including focusing on error labels instead of scores, ensembling, designing strategies to explicitly focus on the source sentence, focusing on semantic content rather than relying on the lexical overlap, and choosing the right base model for representations.


\end{abstract}

\section{Introduction}

Machine Translation (MT) metrics are a fundamental component of the development of high-quality MT systems as most state-of-the-art models claim their effectiveness through such metrics \citep{kocmi-etal-2021-ship}. While human evaluation of these MT systems is ideal, it is labour-intensive, time-consuming, and expensive. Development of automatic metrics has thus received significant interest over the past years \citep{koehn-monz-2006-manual, freitag-etal-2023-results} resulting in a surge of new metrics. These metrics are typically judged by their ability to distinguish the quality of one machine translation system over another (system-level) on large test sets. This type of evaluation only provides an overview and it is difficult to identify whether these metrics are robust to specific MT errors.

To systematically study the advantages and shortcomings of MT metrics, and to identify broad trends in metric development, we rely on the construction of challenge sets for MT metrics. Challenge sets are a useful tool in measuring the performance of systems or metrics on one or more specific phenomena of interest. They may be used to compare the performance of a range of \textit{different} systems or to identify performance improvement/degradation between successive iterations of the \textit{same} system. 

The WMT 2021 Metrics shared task \citep{freitag-etal-2021-results} introduced a shared task on constructing contrastive challenge sets for the evaluation of MT metrics. Contrastive challenge sets aim to assess how well a given metric can discriminate between a \textit{good} and \textit{incorrect} translation of the \textit{source} text where the incorrect translation consists of a translation error of interest.
Providing a \textit{reference} translation allows for flexibility: it may be included to assess reference-based (i.e. MT) metrics or excluded to assess reference-free (i.e. Quality Estimation (QE)) metrics. Benchmarking metrics on such challenge sets provides insights into their strengths while simultaneously uncovering their weaknesses on different translation errors.

In this work, we describe the Translation \textbf{A}ccuracy \textbf{C}halleng\textbf{E} \textbf{S}et (\textsc{ACES}) dataset submitted to the challenge sets subtask of the WMT 2022 and 2023 Metrics shared task and its subsequent expansion to include error span annotations (\textsc{span-ACES}). The \textsc{ACES} dataset\footnote{The \textsc{ACES} dataset is available at \url{https://huggingface.co/datasets/nikitam/ACES}} \citep{amrhein-etal-2022-aces} consists of 36,476 examples covering 146 language pairs and representing challenges from 68 phenomena. Most MT metric challenge sets \citep{avramidis-etal-2018-fine,alves-etal-2022-robust,karpinska-etal-2022-demetr} either focus on a small number of phenomena or a small number of languages. Our datasets are large scale in coverage of phenomena as well as language pairs providing a comprehensive challenge set for MT metrics.
\begin{figure*}
    \centering
    \includegraphics[width=\textwidth]{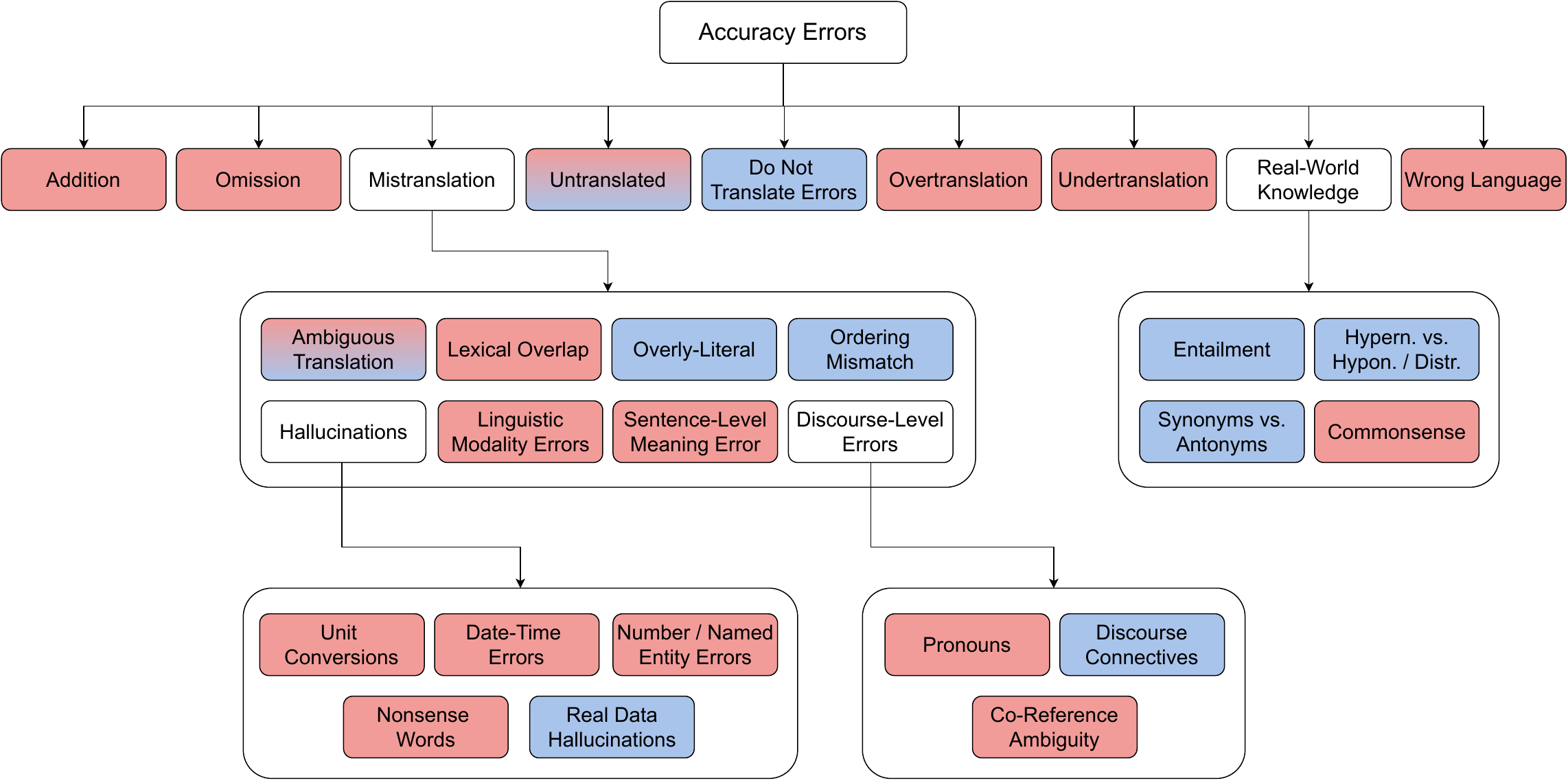}
    \caption{Diagram of the error categories on which our collection of challenge sets is based. Red means challenge sets are created automatically, blue means challenge sets are created manually.}
    \label{fig:diagram}
\end{figure*}

We focus on translation accuracy errors because in recent years, machine translation outputs have become increasingly fluent \citep{bentivogli-etal-2016-neural,toral-sanchez-cartagena-2017-multifaceted,article}. Further, accuracy errors can have dangerous consequences in certain contexts, for example in the medical and legal domains \citep{doi:10.1080/1369118X.2020.1776370}.

\textsc{ACES} uses the hierarchy of errors under the class \textit{Accuracy} from the Multidimensional Quality Metrics (MQM) ontology \citep{lommel2014} to design the \textsc{ACES} challenge sets. We extend this ontology by two error classes (translations defying real-world knowledge and translations in the wrong language) and specify several more specific subclasses such as discourse-level errors or ordering mismatches.  We include phenomena ranging from simple perturbations involving the omission/addition of characters or tokens to more complex examples involving mistranslation e.g. ambiguity and hallucinations in translation, untranslated elements of a sentence, discourse-level phenomena, and real-world knowledge.
A full overview of all error classes can be seen in Figure~\ref{fig:diagram}. Our challenge set consists of synthetically generated adversarial examples, examples from re-purposed contrastive MT test sets (both marked in red), and manually annotated examples (marked in blue). 

We use \textsc{ACES} to benchmark the metrics that participated in the WMT 2022 and 2023 metrics shared tasks. We also investigate whether Large Language Models can perform MT evaluation \citep{kocmi-federmann-2023-large, xu-etal-2023-instructscore}. We conduct several analyses on these results revealing: 

\begin{enumerate}
    \item There is no \textit{winning} metric as conducting granular evaluation reveals different metrics have different strengths and weaknesses.
    \item Most metrics tend to disregard information present in the source.
    \item Reference-based neural metrics still rely on surface-level overlap.
    \item Some properties of the base model in neural metrics may cause undesirable effects on evaluation.
\end{enumerate}

A metric that can, in addition to providing scores, accurately label errors in MT output provides many clear advantages over one that only provides scores \citep{freitag-etal-2021-experts}. Observations by \citet{moghe-etal-2023-extrinsic} suggest that interpreting the quality of MT output based on scores is both unreliable and uninformative. Instead, they recommend the development of metrics that predict labels for error spans in the MT output. Similarly, \citet{lommel2014,freitag-etal-2021-experts} and the recent WMT challenges \citep{freitag-etal-2021-results, freitag-etal-2022-results, freitag-etal-2023-results} also advocate the use of labelled error spans for MT evaluation.  When considering whether to deploy an MT system (or which of several systems to deploy), system developers can take into consideration the type, frequency, and severity of the errors that the system is likely to make, coupled with information about what types of errors may be tolerated/not for a given downstream task. 

With these motivations, we extend the \textsc{ACES} dataset into \textsc{Span-ACES}, where we include error span annotations for each example. These annotations indicate the location of error spans present in the \textit{incorrect} translation and pertaining to the specific linguistic phenomenon in focus. Whilst some currently available MT metrics are already able to mark error spans including MATESE \citep{perrella-etal-2022-matese}, COMET-22 \citep{COMET:WMT22} trained on MQM \citep{lommel2014}, GEMBA-MQM \citep{kocmi-federmann:2023:WMT}, AutoMQM \citep{fernandes-EtAl:2023:WMT} that prompt LLMs to obtain the corresponding error span, we believe that error-span labelling is an important next step in MT metric evolution. Independent challenge sets such as \textsc{span-ACES} will be essential in driving development forward. We benchmark GEMBA-MQM \citep{kocmi-federmann:2023:WMT}), XCOMET-XL \citep{guerreiro2023xcomet}, and adapted versions of COMET-22 \citep{COMET:WMT22} and UniTE \citep{wan-etal-2022-unite} on \textsc{Span-ACES}.

In this article, we provide an overview of the \textsc{ACES} challenge set and its participation at the WMT 2022 and 2023 Metrics shared task - Challenge Sets subtask \citep{amrhein-etal-2022-aces,amrhein-moghe-guillou:2023:WMT}. We list our contributions and also report on extensions to our previously published work, including novel contributions (listed from   4):

\begin{itemize}
    \item We briefly present the construction of \textsc{ACES}, containing 36k examples across 146 language pairs and 68 phenomena.  
    \item We evaluate \textsc{ACES} on the metrics submitted to the WMT 2022 and WMT 23 Metrics shared task providing an overview of the performance of 50 different metrics.
    \item We conduct several analyses on these metrics revealing their drawbacks and also providing recommendations to mitigate them.
    \item We describe the construction of \textsc{span-ACES}, an extended version \textsc{ACES} which includes error span annotations.
    \item Using \textsc{span-ACES}, we benchmark the performance of currently available metrics for the task of labelling errors in MT output. 
    Our results suggest that these methods show some success on the error labelling task with the highest span-F1 score reaching 26.9. However, these results and corresponding poor results on the contrastive task also raise new questions in labelling MT errors as evaluation. 
    \item We present the results of analyses aimed at determining how \textit{sensitive} metrics are to different phenomena. This is grounded in our assertion that an ideal metric should assign comparatively high scores to good translations and low scores to incorrect translations.
    \item We investigate claims that Large Language Models (LLMs) may be used as MT evaluators and describe experiments on LLMs from three different families. Benchmarking these LLMs on \textsc{ACES} reveals that these models perform worse than the string-overlap metrics. These results degrade further in the reference-free setting where all of the LLMs have a negative correlation across all of the \textsc{ACES} categories. 
\end{itemize}

We advocate steering metric development towards methods that produce error labels in addition to the scores. Based on our analyses, we also recommend that metric developers consider: a) combining metrics with different strengths, e.g. in the form of ensemble models, b) paying more attention to the source and avoiding over-reliance on surface-overlap with the reference, and c) check the base model properties prior to their use in developing new metrics

We propose the adoption of both \textsc{ACES} and \textsc{span-ACES} by the MT community, as a benchmark for developing Machine Translation metrics. We envisage several use cases in which the challenge sets may be employed: to profile and compare metric performance across a range of error categories, and to identify improvement/degradation in performance of successive development iterations of the same metric. Similarly, MT models can also be evaluated using this dataset by calculating  sentence-level perplexity of the two translations.  Furthermore, we propose the use of \textsc{span-ACES} to aid in advancing the development of the next generation of MT metrics which aim to provide error-span labels over MT output in addition to scores.

\section{Related Work}
\label{sec:related_work}

Challenge sets have been used for a range of NLP tasks to investigate the behaviour of these tasks under a specific phenomenon rather than the standard test distribution \citep{popovic-castilho-2019-challenge}. Challenge sets aim to provide insights on whether state-of-the-art models are robust to domain shifts, simple textual perturbations, whether they have some understanding of linguistic phenomena such as negation/commonsense or simply rely on shallow heuristics, to name a few. The earliest introduction of challenge sets was by \citet{king-falkedal-1990-using} who probed the acceptability of machine translations for different domains. Since then challenge sets have been developed for different fields within NLP including parsing \citep{rimell-etal-2009-unbounded}, NLI \citep{mccoy2019non,Rocchietti2021FANCYAD}, question answering \citep{ravichander-etal-2021-noiseqa}, reading comprehension \citep{khashabi-etal-2018-looking} and sentiment analysis \citep{li-etal-2017-bibi,mahler-etal-2017-breaking,staliunaite-bonfil-2017-breaking}.  Challenge sets are also referred as  ``adversarial datasets'' which also create examples by perturbing the standard test set to fool the model (\citet{DBLP:journals/corr/abs-1207-0245, jia-liang-2017-adversarial},  \textit{inter-alia}).

Challenge sets for evaluating MT systems have focused on the translation models' ability to generate the correct translation given a phenomenon of interest. These include word sense ambiguity \citep{rios-etal-2018-word,campolungo-etal-2022-dibimt}, gender bias \citep{rudinger-etal-2017-social, zhao-etal-2018-gender, stanovsky-etal-2019-evaluating}, structural divergence \citep{isabelle-etal-2017-challenge} and discourse level phenomena \citep{guillou-hardmeier-2016-protest,emelin-sennrich-2021-wino}. While such challenge sets focus on evaluating specific MT models, it is necessary to identify whether the existing MT evaluation metrics also perform well under these and related phenomena. Following the success of neural MT metrics, which have been shown to correlate well with human judgements \citep{freitag-etal-2021-results, kocmi-etal-2021-ship}, the development of challenge sets designed to examine their strengths and weaknesses has received considerable interest. However, metric weaknesses remain relatively unknown and only a small number of works (e.g. \citet {hanna-bojar-2021-fine} and \citet{amrhein2022identifying}) have proposed systematic analyses to uncover them. 

Early work on constructing challenge sets for metric evaluation typically focused on a small range of phenomena \citep{specia-etal-2020-findings}, synthetic perturbations \citep{freitag-etal-2021-results}, or manual perturbations for high-resource language pairs \citep{avramidis-etal-2018-fine}. These limitations have been addressed in the development of the DEMETR \citep{karpinska-etal-2022-demetr} and \textsc{ACES} datasets.

DEMETR \citep{karpinska-etal-2022-demetr}, which comprises 31K English examples translated from ten languages, was developed for evaluating MT metric sensitivity to a range of 35 different types of linguistic perturbations, belonging to semantic, syntactic, and morphological error categories. These were divided into minor, major, and critical errors according to the type of perturbation, similar to the grading of error categories to compute the weighted \textsc{ACES}-Score. As in \textsc{ACES}, example generation was carefully designed to form minimal pairs such that the perturbed translation only differs from the actual translation in one aspect. The application of DEMETR in evaluating a suite of baseline metrics revealed a similar pattern to the analyses in \citet{amrhein-etal-2022-aces} - that metric performance varies considerably across the different error categories, often with no clear winner. It is worth noting that DEMETR and \textsc{ACES} each have their respective advantages: all examples in DEMETR have been verified by human annotators; \textsc{ACES} provides broader coverage in terms of both languages and linguistic phenomena.

In addition to \textsc{ACES}, three other datasets were submitted to the WMT 2022 challenge sets shared task \citep{freitag-etal-2022-results}: SMAUG 
\citep{alves-etal-2022-robust}, the HWTSC challenge set \citep{chen-etal-2022-exploring}, and the DFKI challenge set \citep{avramidis-macketanz-2022-linguistically}. These datasets differ from \textsc{ACES} in terms of their size, and the languages and phenomena/categories they cover. 
Both SMAUG and HWTSC are relatively small datasets (<1000 examples) focusing on a small set of five phenomena, each pertaining to a single category of critical error for meaning change. In comparison, the DFKI challenge set is much larger -- it contains 19,347 examples and covers over 100 linguistically motivated phenomena, which are organised into 14 categories. Whereas the aim of \textsc{ACES} was to provide a broad coverage of language pairs, the other datasets provide an in-depth focus on specific high-resource language pairs: SMAUG (pt$\leftrightarrow$en and es$\rightarrow$en), DFKI (de$\leftrightarrow$en), and HWTSC (zh$\leftrightarrow$en). Whilst there is a clear overlap between the \textsc{ACES} phenomena and those in SMAUG and HWTSC, many of the phenomena in the DFKI dataset are complementary such that in the case of evaluating metrics for the German-English pair, metric developers might consider benchmarking on both datasets.

The WMT 2023 Challenge Sets submissions included \textsc{ACES}, MSLC23 \citep{lo-larkin-knowles:2023:WMT}, and an extended version of the DFKI challenge set to include the en$\rightarrow$ru language pair plus additional examples and phenomena for the en$\rightarrow$de language pair \citep{avramidis-etal-2023-challenging}. The MSLC23 dataset covers four language pairs (zh$\rightarrow$en, he$\leftrightarrow$en, and en$\rightarrow$de) and includes examples of low-, medium- and high-quality output designed to provide an interpretation of metric performance across a range of different levels of translation quality. The motivation for this is that whilst metric performance may be evaluated on high-quality MT output, these same metrics may later be used to evaluate low-quality MT output, and it is, therefore, important to understand their performance in the lower-quality setting.

We note that our contrastive challenge set with error span annotations is the first of its kind.

\section{Challenge Sets}
\label{sec:challengesets}
Creating a contrastive challenge set for evaluating a machine translation evaluation metric requires a source sentence, a reference translation, and two translation hypotheses: one which contains an error or phenomenon of interest (the ``incorrect'' translation) and one which is a correct translation in that respect (the ``good'' translation). One possible way to create such challenge sets is to start with two alternative references (or two identical copies of the same reference) and insert errors into one of them to form an incorrect translation while the uncorrupted version can be used as the good translation. This limits the full evaluation scope to translation hypotheses that only contain a single error. To create a more realistic setup, we also create many challenge sets where the good translation is not free of errors, but it is a better translation than the incorrect translation. For automatically created challenge sets, we put measures in place to ensure that the incorrect translation is indeed a worse translation than the good translation.

\subsection{Datasets}

The examples in \textsc{ACES} are based on several academic datasets designed to test particular properties in Machine Translation or other multilingual NLP tasks. The majority of the examples in our challenge set were based on data extracted from three main datasets: FLORES-101, PAWS-X, and XNLI (with additional translations from XTREME). \textbf{FLORES-101} \citep{goyal-etal-2022-flores}  and \textbf{FLORES-200} \citep{flores-200} are low resource MT evaluation benchmarks with parallel data in 101 and 200 languages respectively. \textbf{PAWS-X} \citep{yang-etal-2019-paws} is a cross-lingual dataset for paraphrase identification in seven languages that consists of pairs of sentences that are labelled as true or adversarial paraphrases. \textbf{XNLI} \citep{conneau-etal-2018-xnli} is a multilingual Natural Language Inference (NLI) dataset consisting of premise-hypothesis pairs with their corresponding inference label for 14 languages. The other datasets used in the development of \textsc{ACES} serve specific challenges.  \textbf{WinoMT} \citep{stanovsky-etal-2019-evaluating}, a challenge set developed for analysing gender bias in MT with examples exhibiting an equal balance of male and female genders, and of stereotypical and non-stereotypical gender-role assignments (e.g., a female nurse vs. a female doctor). \textbf{MuCoW} \citep{raganato-etal-2019-mucow} is a multilingual contrastive word sense disambiguation test suite for machine translation. The \textbf{WMT 2018 English-German pronoun translation evaluation test suite} \citep{guillou-etal-2018-pronoun} contains examples of the ambiguous English pronouns \textit{it} and \textit{they} extracted from the TED talks portion of ParCorFull \citep{lapshinova-koltunski-etal-2018-parcorfull}.  The \textbf{Europarl ConcoDisco} corpus \citep{laali-kosseim-2017-improving} comprises the English-French parallel texts from Europarl \citep{koehn-2005-europarl} over which automatic methods were used to perform discourse connective annotation of their sense types.  \textbf{Wino-X} \citep{emelin-sennrich-2021-wino} is a parallel dataset of German, French, and Russian Winograd schemas, aligned with their English counterparts used to test commonsense reasoning and coreference resolution of MT models.

We will now discuss the different categories of challenge sets. We list some examples from \textsc{ACES} in Table \ref{tab:ACES_top_level_examples}.

\subsection{Addition and Omission}
\label{sec:addition-omission}
We create a challenge set for addition and omission errors which are defined in the MQM ontology as ``target content that includes content not present in the source'' and ``errors where content is missing from the translation that is present in the source'', respectively. 
We focus on the level of constituents and use an implementation by \citet{vamvas-sennrich-2022-little} to create synthetic examples of addition and omission errors using the likelihood of tokens for a given MT model. To generate examples, we use the concatenated dev and devtest sets from the FLORES-101 evaluation benchmark for 46 languages. We focus on the 46 languages for which there exists a stanza parser\footnote{\url{https://stanfordnlp.github.io/stanza/available_models.html}} and create datasets for all languages paired with English plus ten additional language pairs that we selected randomly. For translation, we use the M2M100\footnote{\url{https://huggingface.co/facebook/m2m100_1.2B}} model with 1.2B parameters \citep{fan2021beyond}.
\subsection{Mistranslation}

The mistranslation phenomenon is broadly defined as the target translation not accurately containing the information in the source content. 

\subsubsection{Mistranslation - Ambiguous Translation}
\label{sec:source-disambig}
This error type is defined in the MQM ontology as a case where ``an unambiguous source text is translated ambiguously''. For this error type, we create challenge sets where MT metrics are presented with an unambiguous source and an ambiguous reference. The metrics then need to choose between two disambiguated translation hypotheses where only one meaning matches the source sentence. Therefore, these challenge sets test whether metrics consider the source when the reference is not expressive enough to identify the better translation. Since many reference-based metrics, by design, do not include the source to compute evaluation scores, we believe that this presents a challenging test set.

Our method for creating examples is inspired by \citet{vamvas-sennrich-2021-contrastive} who score a translation against two versions of the source sentence, one with an added correct disambiguation cue and one with a wrong disambiguation cue to determine whether a translation model produced the correct translation or not. Instead of adding the disambiguation cues to the source, we use an unambiguous source and add disambiguation cues to an ambiguous reference to create two contrasting translation hypotheses. We create three separate challenge sets of this type: 
\noindent{\textbf{Occupation Name Gender}} using the WinoMT dataset where the source language is a gendered language and the target language is English.
The cues added to the reference to form the ``good'' and ``incorrect'' translations are ``female'' and ``male''. 

\noindent \textbf{Word Sense Disambiguation} using the MuCoW dataset where the ambiguity lies in homographs in the target language that are unambiguous in the source sentence. The cues added to the reference to form the contrastive translations are sense-specific.

\noindent \textbf{Discourse Connectives} using the Europarl ConDisco corpus where the ambiguity lies in the English discourse connective ``since'' which can have both causal and temporal meanings.

\subsubsection{Mistranslation - Hallucinations}
\label{sec:hallucination}
In this category, we group several subcategories of mistranslation errors that happen at the word level and could occur due to hallucination by an MT model. Hallucinations are erroneous generations where the output is partially related or unrelated to the source sentence \citep{dale-etal-2023-detecting}.
These challenge sets test whether the machine translation evaluation metrics can reliably identify hallucinations when presented with a correct alternative translation. 

\noindent We create five different challenge sets based on hallucination errors:

\noindent{\textbf{Date-Time Errors}}: \label{p:date-time} using the FLORES-101 data where a month name in the reference (e.g. November) is replaced with a corresponding abbreviation in the ``good'' translation (e.g. Nov.) and a different month name in the ``incorrect'' translation (e.g. August). 


\noindent{\textbf{Numbers and Named Entities}}: We create a challenge set for numbers and named entities where we perform character-level edits (adding, removing or substituting digits in numbers or characters in named entities) as well as word-level edits (substituting whole numbers or named entities). In the 2021 WMT metrics shared task, number differences were not a big issue for most neural metrics \citep{freitag-etal-2021-results}. However, we believe that simply changing a number in an alternative translation and using this as an incorrect translation as done by \citet{freitag-etal-2021-results} is an overly simplistic setup and does not cover the whole translation hypothesis space. To address this shortcoming, we propose a three-level evaluation. The first, easiest level follows \citet{freitag-etal-2021-results} and applies a change to an alternative translation to form an incorrect translation. The second level uses an alternative translation that is lexically very similar to the reference as the good translation and applies a change to the reference to form an incorrect translation. The third, and hardest level, uses an alternative translation that is lexically very different from the reference as the good translation and applies a change to the reference to form an incorrect translation. In this way, our challenge set tests whether the number and named entity differences can still be detected as the surface similarity between the two translation candidates decreases and the surface similarity between the incorrect translation and the reference increases.
We use cross-lingual paraphrases from the PAWS-X dataset as a pool of alternative translations to create this challenge set. We only consider language pairs for which we can use a spacy NER model on the target side, which results in 42 language pairs.

\noindent{\textbf{Unit Conversion}}: \label{p:unit-conversion} using FLORES-101 dataset, where we replace unit mentions in the reference (e.g. 100 feet) with a different unit and corresponding amount in the ``good'' translation (e.g. 30.5 metres) and either the wrong amount (e.g. 100 metres) or wrong unit (30.5 feet) compared to the reference in the ``incorrect'' translation. 


\noindent{\textbf{Nonsense Words}}: We develop a challenge set for evaluating hallucinations at subword level \citep{sennrich-etal-2016-neural}.
To create this challenge set, we consider tokens which are broken down into at least two subwords and then randomly swap those subwords with other subwords to create nonsense words by using the multilingual BERT tokenizer \citep{devlin-etal-2019-bert}. We use the paraphrases from the PAWS-X dataset as good translations and randomly swap one subword in the reference to generate an incorrect translation. 

\noindent{\textbf{Real Data Hallucinations}}: To also create a more realistic hallucination benchmark, we manually check some machine translations of the FLORES-101 dev and devtest sets for four language pairs: de$\rightarrow$en, en$\rightarrow$de, fr$\rightarrow$de and en$\rightarrow$mr. We consider both cases where a more frequent, completely wrong word occurs and cases where the MT model started with the correct subword but then produced random subwords as hallucinations. Translations with a hallucination are used as incorrect translations. We manually replace the hallucination part with its correct translation to form the good translation.

\subsubsection{Mistranslation - Lexical Overlap}
\label{subsec:lexical-overlap}
Language models trained with the masked language modelling objective are successful on downstream tasks because they model higher-order word co-occurrence statistics instead of syntactic structures \citep{sinha-etal-2021-masked}. Similarly, existing surface-level metrics rely on n-gram matching between the hypothesis and the reference. We create this challenge set to test if metrics can reliably identify an incorrect translation especially when it shares a high degree of lexical overlap with the reference. To create such examples, we use the PAWS-X dataset for which adversarial paraphrase examples were constructed by changing the word order and/or the syntactic structure while maintaining a high degree of lexical overlap.

\subsubsection{Mistranslation - Linguistic Modality}
Modal auxiliary verbs signal the function of the main verb that they govern. For example, they may be used to denote possibility (``could''), permission (``may''), the giving of advice (``should''), or necessity (``must''). We are interested in whether MT evaluation metrics can identify when modal auxiliary verbs are incorrectly translated. We focus on the English modal auxiliary verbs: ``must'' (necessity), and ``may'', ``might'', ``could'' (possibility).  We then translate the source sentence using Google Translate to obtain the ``good'' translation and manually replace the modal verb with an alternative with the same meaning where necessary (e.g. ``have to'' denotes necessity as does ``must''; also ``might'', ``may'' and ``could'' are considered equivalent). For the incorrect translation, we manually substitute the modal verb that conveys a different meaning or \textit{epistemic strength} e.g. in the example above ``might'' (possibility) is replaced with ``will'', which denotes (near) certainty. We use a combination of the FLORES-200 and PAWS-X datasets as the basis of the challenge sets.

\subsubsection{Mistranslation - Overly Literal Translations}
\label{sec:overly_literal}
MQM defines this error type as translations that are overly literal, for example, literal translations of figurative language.  We create two challenge sets based on this error type:

\noindent {\textbf{Idioms}}: We create this challenge set based on the PIE\footnote{\url{https://github.com/zhjjn/MWE_PIE}} parallel corpus of English idiomatic expressions and literal paraphrases \citep{zhou-etal-2021-pie}. We manually translate 102 parallel sentences into German for which we find a matching idiom that is not a word-by-word translation of the original English idiom. Further, we create an overly literal translation of the English and German idioms. We use either the German or English original idiom as the source sentence. Then, we either use the correct idiom in the other language as the reference and the literal paraphrase as the good translation, or vice versa. The incorrect translation is always the overly literal translation of the source idiom.

\noindent{\textbf{Real Data Errors}}: For this challenge set, we manually check MT translations of the FLORES-101 datasets. If we find an overly-literal translation, we manually correct it to form the good translation and use the overly-literal translation as the incorrect translation.

\subsubsection{Mistranslation - Sentence-Level Meaning Error}
\label{subsec:lexically-similar}
We also consider a special case of sentence-level semantic error that arises due to the nature of the task of Natural Language Inference (NLI). The task of NLI requires identifying where the given hypothesis is an entailment, contradiction, or neutral, for a given premise. Thus, the premise and hypothesis have substantial overlap but they vary in meaning. We use the XNLI dataset to create such examples where there is at least a 0.5 chrF score between the English premise and hypothesis only for the neutral and contradiction examples. We use either the premise/hypothesis as the reference, an automatic translation as the ``good translation'', premise/hypothesis from the remaining non-English languages, and hypothesis/premise as the ``incorrect translation''.

\subsubsection{Mistranslation - Ordering Mismatch}
\label{subsec:ordering-mismatch}
We also investigate the effects of changing word order in a way that changes meaning. For example, ``I like apple pie and fried chicken'' is changed to `` I like chicken pie and fried apple'' to form the incorrect translation. 
This challenge set is created manually by changing translations from the FLORES-101 dataset and covers de$\rightarrow$en, en$\rightarrow$de and fr$\rightarrow$de.

\subsection{Mistranslation - Discourse-level Errors}
\label{sec:discourse}
We introduce a new subclass of mistranslation errors that specifically cover discourse-level phenomena. We create several challenge sets based on discourse-level errors:

\noindent{\textbf{Pronouns}}: To create these challenge sets, we use the English-German pronoun translation evaluation test suite from the WMT 2018 shared task as the basis for our examples. We focus on the following six categories derived from the manually annotated pronoun function and attribute labels: pleonastic \textit{it}, anaphoric subject and non-subject position \textit{it}, anaphoric \textit{they}, singular \textit{they}, and group \textit{it/they}. We use the MT translations as the ``good'' translations and automatically generate ``incorrect'' translations using one of the following strategies: \textit{omission} - the translated pronoun is deleted from the MT output, \textit{substitution} - the ``correct'' pronoun is replaced with an ``incorrect'' form.

\noindent {\textbf{Discourse Connectives}}: We leverage the Europarl ConcoDisco corpus of parallel English/French sentences with discourse connectives marked and annotated for sense, and select examples with ambiguity in the French source sentence. We construct the good translation by replacing instances of ``while'' (temporal) with ``as'' or ``as long as'' and instances of ``while'' (comparison) as ``whereas'' (ensuring grammaticality is preserved). For the incorrect translation, we replace the discourse connective with one with the alternative sense of ``while'' e.g. we use ``whereas'' (comparison) where a temporal sense is required.

\noindent{\textbf{Commonsense Co-Reference Disambiguation}}: We use the English sentences in the Wino-X challenge set which were sampled from the Winograd schema. All contain the pronoun \textit{it} and were manually translated into two contrastive translations for de, fr, and ru. Based on this data, we create our challenge sets covering two types of examples: For the first, the good translation contains the pronoun referring to the correct antecedent, while the incorrect translation contains the pronoun referring to the incorrect antecedent. For the second, the correct translation translates the instance of \textit{it} into the correct disambiguating filler, while the second translation contains the pronoun referring to the incorrect antecedent.

\subsection{Untranslated}
\label{sec:untranslated}
MQM defines this error type as ``errors occurring when a text segment that was intended for translation is left untranslated in the target content''. We create two challenge sets based on untranslated content errors:

\noindent{\textbf{Word-Level}}: We manually annotate real errors in translations of the FLORES-101 dev and devtest sets. We count complete copies as untranslated content as well as content that comes from the source language but was only adapted to look more like the target language. 

\noindent{\textbf{Sentence-Level}}: We create a challenge set for untranslated sentences by simply copying the entire source sentence as the incorrect translation. We used a combination of examples from the FLORES-200, XNLI, and PAWS-X datasets to create these examples.

\subsection{Do Not Translate Errors}
\label{sec:do-not-translate}
This category of errors is defined in MQM as content in the source that should be copied to the output in the source language but was mistakenly translated into the target language. Common examples of this error type are company names or slogans. Here, we manually create a challenge set based on the PAWS-X data which contains many song titles that should not be translated. To construct the challenge set, we use one paraphrase as the good translation and manually translate an English sequence of tokens (e.g. a song title) into German to form the incorrect translation.

\subsection{Overtranslation and Undertranslation}
\label{sec:overtranslation_undertranslation}

Hallucinations from a translation model can often produce a term which is either more generic than the source word or more specific. Within the MQM ontology, the former is referred to as undertranslation while the latter is referred to as overtranslation. 
For example, ``car'' may be substituted with ``vehicle'' (undertranslation) or ``BMW'' (overtranslation). A randomly selected noun from the reference translation is replaced by its corresponding hypernym or hyponym (by using Wordnet) to simulate undertranslation or overtranslation errors, respectively.

\subsection{Real-world Knowledge}
\label{sec:real-world-knowledge}

We propose a new error category where translations disagree with real-world knowledge in addition to the accuracy categories in MQM. 
We create five challenge sets based on this error type. For the first four, we manually construct examples each for en$\rightarrow$de and de$\rightarrow$en. We used German-English examples from XNLI, plus English translations from XTREME as the basis for our examples. Typically, we select a single sentence, either the premise or hypothesis from XNLI, and manipulate the MT translations.

\noindent{\textbf{Textual Entailment}}: We construct examples for which the good translation entails the meaning of the original sentence (and its reference). For example, we use the entailment \textit{was murdered} $\rightarrow$ \textit{died} (i.e. if a person is murdered then they must have died) to construct the good translation in the example above. We construct the incorrect translation by replacing the entailed predicate (\textit{died}) with a related but non-entailed predicate (here \textit{was attacked}) -- a person may have been murdered without being attacked, i.e. by being poisoned for example.

\noindent{\textbf{Hypernyms and Hyponyms}}: We consider a translation that contains a \textit{hypernym} of a word to be better than one that contains a \textit{hyponym}. For example, whilst translating ``Hund'' (``dog'') with the broader term ``animal'' results in some loss of information, this is preferable over hallucinating information by using a more specific term such as ``labrador'' (i.e. an instance of the hyponym class ``dog''). We used Wordnet and WordRel.com\footnote{\url{https://wordrel.com/}} (an online dictionary of words’ relations) to identify hypernyms and hyponyms of nouns within the reference sentences, and used these as substitutions in the MT output: hypernyms are used in the ``good'' translations and hyponyms in the ``incorrect'' translations.

\noindent{\textbf{Hypernyms and Distractors}}: Similar to above, we construct examples in which the good translation contains a hypernym (e.g. ``pet'') of the word in the reference (e.g. ``dog''). We form the incorrect translation by replacing the original word in the source/reference with a different member from the same class (e.g. ``cat''; both cats and dogs belong to the class of pets). Note the techniques in Section ~\ref{sec:overtranslation_undertranslation} manipulate the reference only to create an incorrect translation with the respective error. 

\noindent{\textbf{Antonyms}}: We also construct incorrect translations by replacing words with their corresponding antonyms from Wordnet. We construct challenge sets for both nouns and verbs. For nouns, we automatically constructed incorrect translations by replacing nouns in the reference with their antonyms. 
In the case of verbs, we manually constructed a more challenging set of examples intended to be used to assess whether the metrics can distinguish between translations that contain a synonym versus an antonym of a given word. 

\noindent{\textbf{Commonsense}}: We are also interested in whether evaluation metrics prefer translations that adhere to common sense. To test this, we remove explanatory subordinate clauses from the sources and references in the dataset described in Section~\ref{sec:discourse}. This guarantees that when choosing between a good and incorrect translation, the metric cannot infer the correct answer from looking at the source or the reference. We then pair the shortened source and reference sentences with the full translation that follows commonsense as the good translation and the full translation with the other noun as the incorrect translation.

\subsection{Wrong Language}
\label{sec:wrong_language}
Most of the representations obtained from large multilingual language models do not explicitly use the language identifier (id) as an input while encoding a sentence. Here, we are interested in checking whether sentences which have similar meanings are closer together in the representation space of neural MT evaluation metrics, irrespective of their language.  We create a challenge set for embedding-based metrics using the FLORES-200 dataset where the incorrect translation is in a similar language (same typology/same script) to the reference (e.g. a Catalan translation may be used as the incorrect translation if the target language is Spanish).

\begin{table*}[ht!]
    \small
    \centering
    \resizebox{\textwidth}{!}{
    \begin{tabular}{rl}
        \toprule
        & \textbf{Addition}\\
        & \textit{target includes content not present in the source}\\\\
        SRC (de): & In den letzten 20 Jahren ist die Auswahl in Uptown Charlotte exponentiell gewachsen.\\
        REF (en): & In the past 20 years, the amount in Uptown Charlotte has grown exponentially.\\
        \cmark: & Over the past 20 years, the selection in Uptown Charlotte has grown exponentially.\\
        \xmark: & Over the past 20 years, the selection of \textbf{child-friendly options} in Uptown Charlotte has grown exponentially.\\
        \midrule
        
        & \textbf{Omission}\\
        & \textit{errors where content is missing from the translation that is present in the source}\\\\
        SRC (fr): & Une tornade est un tourbillon d'air à basse-pression en forme de colonne, l'air alentour est aspiré vers l'intérieur et le haut.\\
        REF (en): & A tornado is a \textbf{spinning column} of very low-pressure air, which sucks the surrounding air inward and upward.\\
        \cmark: & A tornado is a \textbf{column-shaped} low-pressure air turbine, the air around it is sucked inside and up.\\
        \xmark: & A tornado is a low-pressure air turbine, the air around it is sucked inside and up.\\
        \midrule
        
        & \textbf{Untranslated - Word Level} \\
        & \textit{errors occurring when a text segment that was intended for translation is left untranslated in the target content} \\\\
      SRC (fr): & À l'origine, l'émission mettait en scène des \textbf{comédiens de doublage} amateurs, originaires de l'est du Texas. \\
     REF (de): & Die Sendung hatte ursprünglich lokale Amateur\textbf{synchronsprecher} aus Ost-Texas. \\
     \cmark{} (copy): & Ursprünglich spielte die Show mit Amateur\textbf{synchronsprechern} aus dem Osten von Texas. \\
     \cmark{} (syn.): & Ursprünglich spielte die Show mit Amateur-\textbf{Synchron-Schauspielern} aus dem Osten von Texas. \\
     \xmark: & Ursprünglich spielte die Show mit Amateur-\textbf{Doubling-Schauspielern} aus dem Osten von Texas. \\
        \midrule
        
        & \textbf{Mistranslation - Ambiguous Translation} \\
        & \textit{an unambiguous source text is translated ambiguously} \\\\
        SRC (de): &  Der Manager feuerte \textbf{die} Bäcker\textbf{in}. \\
        REF (en): & The manager fired the baker. \\
        \cmark: & The manager fired the \textbf{female} baker. \\
        \xmark: & The manager fired the \textbf{male} baker. \\
        \midrule

        & \textbf{Do Not Translate} \\
        & \textit{content in the source that should be copied to the output in the source language, but was mistakenly translated into the target language.} \\\\

       SRC (en): & Dance was one of the inspirations for the exodus - song \textbf{``The Toxic Waltz''}, from their 1989 album ``Fabulous Disaster''. \\
     REF (de): & Dance war eine der Inspirationen für das Exodus-Lied \textbf{„The Toxic Waltz“} von ihrem 1989er Album „Fabulous Disaster“. \\
     \cmark: & Der Tanz war eine der Inspirationen für den Exodus-Song \textbf{„The Toxic Waltz“}, von ihrem 1989er Album „Fabulous Disaster''. \\
     \xmark: & Der Tanz war eine der Inspirationen für den Exodus-Song \textbf{„Der Toxische Walzer“}, von ihrem 1989er Album „Fabulous Disaster''.\\
        \midrule

        & \textbf{Undertranslation} \\
        & \textit{erroneous translation has a meaning that is more generic than the source} \\\\

        SRC (de): & Bob und Ted waren Brüder. Ted ist der \textbf{Sohn} von John. \\
        REF (en): & Bob and Ted were brothers. Ted is John's \textbf{son}. \\
        \cmark: & Bob and Ted were brothers, and Ted is John's \textbf{son}. \\
       \xmark: & Bob and Ted were brothers. Ted is John's \textbf{male offspring}. \\
        \midrule
        
        & \textbf{Overtranslation} \\
        & \textit{erroneous translation has a meaning that is more specific than the source} \\\\

        SRC (ja): & \begin{CJK}{UTF8}{min}その 40 分の\textbf{映画}はアノーがアラン・ゴダードと協力して脚本を書いた。\end{CJK} \\
        REF (en): & The 40-minute \textbf{film} was written by Annaud with Alain Godard. \\
        \cmark: &The 40-minute \textbf{film} was written by Annaud along with Alain Godard. \\
       \xmark: & The 40-minute \textbf{cinema verite} was written by Annaud with Alain Godard. \\
        \midrule

         & \textbf{Real-world Knowledge - Textual Entailment} \\
         & \textit{meaning of the source/reference is entailed  by the ``good'' translation} \\\\
         SRC (de): & Ein Mann \textbf{wurde ermordet}.\\
         REF (en): & A man \textbf{was murdered}.\\
         \cmark: & A man \textbf{died}.\\
         \xmark : & A man \textbf{was attacked}.\\
         \midrule    
        
         & \textbf{Wrong Language} \\
         & \textit{incorrect translation is a perfect translation in a related language} \\\\
          SRC (en): & Cell comes from the Latin word cella which means small room. \\
         REF (es): & El término célula deriva de la palabra latina cella, que quiere decir «cuarto pequeño». \\
         \cmark\ (es): & La célula viene de la palabra latina cella que significa habitación pequeña. \\
         \xmark\ (ca): & Cèl·lula ve de la paraula llatina cella, que vol dir habitació petita. \\

         \bottomrule
    \end{tabular}}
    \caption{Examples from each top-level accuracy error category in \textsc{ACES}. An example consists of a source sentence (SRC), reference (REF), good (\cmark) and incorrect (\xmark) translations, language pair, and a phenomenon label. We also provide a description of the relevant phenomenon. en: English, de: German, fr: French, ja: Japanese, es: Spanish, ca: Catalan}
    \label{tab:ACES_top_level_examples}
\end{table*}

\subsection{Fluency}
Although the focus of \textsc{ACES} is on accuracy errors, we also include a small set of fluency errors for the punctuation category. 

\noindent\textbf{Punctuation}:
\label{sec:punctuation}
We assess the effect of deleting and substituting punctuation characters. We employ four strategies: 1) deleting all punctuation, 2) deleting only quotation marks (i.e. removing indications of quoted speech), 3) deleting only commas (i.e. removing clause boundary markers), 4) replacing exclamation points with question marks (i.e. statement $\rightarrow$ question). In strategies 1 and, especially, 3 and 4, some of the examples may also contain accuracy-related errors. For example, the meaning of the sentence could be changed in the incorrect translation if we remove a comma, e.g. in the (in)famous example ``Let's eat, Grandma!'' vs. ``Let's eat Grandma!''. We use the TED Talks from the WMT 2018 English-German pronoun translation evaluation test suite and apply all deletions and substitutions automatically.

See Appendix \nameref{app:language_pair_matrix} and \nameref{app:language_pair_phenomena} for further information on the distribution of examples and language pairs in ACES.


\section{Span Annotations}
\label{sec:spanannotations}
To support the development of Quality Estimation and MT evaluation metrics that predict error spans, we extended the original version of \textsc{ACES} (released at WMT 2022) to include error span annotations. Specifically, we annotated all error spans of the type denoted by the phenomenon category label, ignoring the presence of errors belonging to other categories. We therefore label only errors present in the incorrect translation, which by design contains errors of the phenomenon category denoted by the label. We annotate spans at the word/token level similar to the MQM format \citep{freitag-etal-2021-experts} and in line with recent developments in error span prediction metrics \citep{perrella-etal-2022-matese, COMET:WMT22}. Following the WMT 2022 MQM Human Evaluation span annotation format \citep{freitag-etal-2022-results}, error spans are enclosed in tags ($\textless$v$\textgreater$ error span $\textless$/ v$\textgreater$) denoting the start and end position of the error in the incorrect translation. Note that due to the formulation of the manual annotation guidelines (see Section~\ref{app:annotation_guidelines}) it is not possible for two spans to overlap.

We provide annotations for all \textsc{ACES} examples, using a combination of automated and manual methods. The annotation methods used for each phenomenon can be found in \nameref{app:automatic_annotations}. For many of the phenomena categories, we were able to automatically annotate examples using rule-based methods informed by the methodology that we followed to construct the examples. For the remaining phenomena, which we could not annotate automatically due to the manual methods used to generate the good and incorrect translations, we annotated the error spans manually (see Section~\ref{app:annotation_guidelines}). We also manually annotated a small number of examples (1959 from the mistranslation phenomena and 3 from the real-world knowledge phenomena) for which the automated annotation rules failed. 

\subsection{Automatic Annotations}
We automatically annotate the error spans in the incorrect translations for 34514 samples out of 36476, by deterministically comparing the incorrect translation to either the good translation or the reference sentence. The automatic annotation methods mainly depends on the way the challenge sets for each phenomena were constructed, and contain only word-level annotations following the annotation guidelines. The details about the automatic annotation methods are as following:
\vspace{-10pt}
\paragraph{\textbf{Annotation of addition, omission and substitutions}}
\label{p:span_annotate_word}
This method tokenises the good translation and incorrect translation, and compares the tokens to annotate word-level addition, omission and substitutions which may occur multiple times. It is only used to annotate the simpler cases of substitutions, when each word was replaced with another word.
\vspace{-10pt}
\paragraph{\textbf{Annotation of substitution of a variable-sized span comparing to the correct translation}}
\label{p:span_diff_flexible}
This method tokenises the good translation and the incorrect translation and then finds a single word-level error span with variable size.
\vspace{-10pt}
\paragraph{\textbf{Annotation of substitution of a variable-sized span comparing to the reference sentence}}
\label{p:span_REF_flexible}
Similar to ``Annotation of substitution of a variable-sized span comparing to the correct translation", this method tokenises the reference and the incorrect translation and then finds a single word-level error span with variable size.
\vspace{-10pt}
\paragraph{\textbf{Annotation of the date-time translation errors}}
\label{p:span_date}
In the \hyperref[p:date-time]{Hallucination - Date-Time} challenge set, the incorrect translations were built by substituting a month name in the reference with another month. This method finds the month names which are different in the incorrect translations and the reference, ignoring the months replaced with their corresponding abbreviations.
\vspace{-10pt}
\paragraph{\textbf{Annotation of the unit-conversion translation errors}}
\label{p:span_units}
In the \hyperref[p:unit-conversion]{Hallucination - Unit Conversion} phenomenon, the unit mentions in the reference (e.g. 100 feet) were replaced with either the wrong amount (e.g. 100 metres) or wrong unit (30.5 feet) in the incorrect translation. Using the Python package \texttt{quantulum3}\footnote{\url{https://github.com/nielstron/quantulum3}}, we detect the amount and units used in the incorrect translation, and annotate either the wrong amount or the wrong unit, according to the phenomenon category label (\hyperref[p:unit-conversion]{hallucination-unit-conversion-unit-matches-ref} and \hyperref[p:unit-conversion]{hallucination-unit-conversion-amount-matches-ref} respectively).
\vspace{-10pt}
\paragraph{\textbf{Annotation of the error where two words in the good translation were swapped}}
\label{p:span_swap}
In \hyperref[subsec:ordering-mismatch]{ordering-mismatch} challenge set, the incorrect sentence was generated by swapping the places of two words in the good translation. This method computes the annotations when two spans were swapped, and we manually annotated 4 samples which the method was not able to correctly annotate. 
\vspace{-10pt}
\paragraph{\textbf{Annotation of the whole sentence}}
\label{p:span_whole_sentence}
This method trivially annotates the whole incorrect translation as an error. For examples belonging to the following \hyperref[subsec:lexically-similar]{Mistranslation - Sentence-Level Meaning Error} phenomena, constructed using the XNLI dataset, we automatically mark the entire sentence as an error: xnli-addition-contradiction, xnli-addition-neutral, xnli-omission-contradiction, xnli-omission-neutral. Despite some degree of lexical overlap between the good- and incorrect-translation, the incorrect-translation is drawn from either a contradiction or neutral hypothesis in the XNLI dataset, and will therefore by definition \textit{not be a translation} of the premise (i.e. the sentence extracted as the good-translation).

\subsection{Manual Annotation}
\label{subsec:span_manual_annotation}
Automated annotation is suitable for many of the examples, e.g. where the good and incorrect translations only exhibit differences relevant to the particular phenomenon indicated by the category label. However, it is not suitable in all cases, for example where the good and incorrect translations contain additional differences (not related to the error phenomenon), which could result in the automatic annotation method introducing annotation errors. We identified four phenomena for which automated annotation was unsuitable, and submitted all examples from these categories for manual annotation. The table below lists the four \textsc{ACES} phenomenon labels and their corresponding category in the manual annotation guidelines.

\begin{table}[h!]
    \centering
    \small
    \begin{tabular}{ll}
        \toprule
        \textbf{\textsc{ACES} Phenomenon Label} & \textbf{Category in Annotation Guidelines} \\
        \midrule
        coreference-based-on-commonsense & coreference \\
        hallucination-real-data-vs-ref-word & hallucination \\
        hallucination-real-data-vs-synonym & hallucination \\
        lexical-overlap & word swap \\
        \bottomrule
    \end{tabular}
\end{table}

We extracted a total of 2,006 examples belonging to these phenomena (427 hallucination, 559 coreference, and 1020 word swap), with examples for the following languages: English (471), French (551), German (456), Japanese (322), Korean (4), Marathi (44), and Russian (158). The manual annotation of these examples was completed by a team of seven annotators (one per language), who are either professional translators or linguists. The annotators were provided with a set of general guidelines plus specific instructions for each of the different phenomena listed above. The annotation guidelines are summarised in the following sections and the complete set of guidelines given to the annotators is provided in \nameref{app:annotation_guidelines}.

Automated checks were carried out over the manual annotations to provide a basic validation. These checks were used to ensure that 1) each example had been annotated, i.e. contained at least one span of text within tags, 2) all spans were marked with an open and close tag (i.e. the number of open and close tags per example, should match), 3) no changes had been made to the example text other than the addition of the tags. Examples that failed these checks were sent to the annotators for re-annotation. We also automatically identified and resolved instances where additional whitespace was introduced (in error) at the start or end of an error span, ensuring that the annotated text and original (unannotated) text differed only in terms of the presence/absence of error tags.


\subsubsection{Overview of Annotation Guidelines}
\label{subsubsec:annotation_guidelines}

We split the annotation guidelines into a) general guidelines suitable for annotating all examples, and b) error type-specific guidelines intended for annotating specific categories. The annotators are presented with an \textsc{ACES} phenomenon label representing the type of error present, and two sentences: A and B, where B is the incorrect translation (i.e. contains one or more errors) and A is either the good translation or the reference (depending on the phenomenon). The annotators are asked to identify and mark \textit{all} error spans in sentence B that belong to the error type indicated by the phenomenon label. Error spans are marked with tags (<>) at the word level, i.e. in the case that the error is a \textit{misspelling} (e.g. ``combuter'' instead of ``computer'') the complete word (i.e. ``combuter'') should be marked.

\textbf{General guidelines.} The general guidelines may be applied for the annotation of any example in ACES. We begin by defining four possible operations to mark error spans: \textit{addition}, \textit{substitution}, \textit{deletion}, and \textit{reordering} (see Table~\ref{tab:manual-annotation-general-operations}). In simple scenarios, a single operation may be sufficient to annotate an example. In more complex scenarios multiple operations may be required. 

\begin{table}[h!]
    \centering
    \begin{tabular}{ll}
        \toprule
        \multicolumn{2}{l}{\textbf{\textit{Addition}}: a text span that is not present in sentence A is included in sentence B}\vspace{0.1cm}\\
        & Sentence A: The cat is a species of small carnivorous mammal.\\
        & Sentence B: The cat is a \textbf{<domestic>} species of small carnivorous mammal.\vspace{0.1cm}\\
        \midrule
        \multicolumn{2}{l}{\textbf{\textit{Substitution}}: a text span in sentence A is substituted with a different text span in sentence B}\vspace{0.1cm}\\
        & Sentence A: Female domestic cats can have kittens from spring to late autumn.\\
        & Sentence B: Female domestic cats can have kittens from \textbf{<May>} to \textbf{<December>}.\vspace{0.1cm}\\
        \midrule
        \multicolumn{2}{l}{\textbf{\textit{Deletion}}: a text span that is present in sentence A is omitted from sentence B}\vspace{0.1cm}\\
        & Sentence A: Feral cats are domestic cats that were born in or have reverted to a wild state.\\
        & Sentence B: Feral cats are domestic cats \textbf{<>}or have reverted to a wild state.\vspace{0.1cm}\\
        \midrule
        \multicolumn{2}{l}{\textbf{\textit{Reordering}}: a text span in sentence A that appears in a different position in sentence B}\vspace{0.1cm}\\
        & Sentence A: Montreal is the second most populous city in Canada and the most\\
        & \qquad\qquad\quad populous city in the province of Quebec.\\
        & Sentence B: Montreal is the \textbf{<>}most populous city in Canada and the \textbf{<second>} most \\
        & \qquad\qquad\quad populous city in the province of Quebec.\vspace{0.1cm}\\
    \end{tabular}
    \caption{Manual annotation guidelines: Operations for general guidelines}
    \label{tab:manual-annotation-general-operations}
\end{table}

\textbf{Error type-specific guidelines:} Additionally, we include specific guidelines for the annotation of three phenomenon categories: \textit{hallucination}, \textit{coreference}, and \textit{word swap} (see Table~\ref{tab:manual-annotation-type-specific}). The annotation of examples belonging to these categories may be achieved by marking the presence of one or more operations. For example, the hallucination example in Table~\ref{tab:manual-annotation-type-specific} contains both an ``addition'' (i.e. \textbf{<Welsh, French,>}) and a ``substitution'' (i.e. Gaelic $\rightarrow$ \textbf{<Garlic>}). The three categories, for which we provide \textit{error type-specific guidelines}, cover all of the examples submitted for manual annotation.

\begin{table}[h!]
    \centering
    \begin{tabular}{ll}
        \toprule
        \multicolumn{2}{l}{\textbf{\textit{Hallucination}}: text that is not present in sentence A is observed in sentence B or a word in}\\
        \multicolumn{2}{l}{sentence A is replaced by a more frequent or \textit{orthographically similar} word in sentence B} \vspace{0.1cm} \\
        & Sentence A: The official languages of Scotland are: English, Scots, and Scottish Gaelic. \\
        & Sentence B: The official languages of Scotland are: English, \textbf{<Welsh, French,>} Scots, and\\
        & \qquad\qquad\quad Scottish \textbf{<Garlic>}. \vspace{0.1cm}\\
        \midrule
        \multicolumn{2}{l}{\textbf{\textit{Coreference}}: a pronoun in sentence A is replaced with a (potentially) inappropriate}\\
        \multicolumn{2}{l}{noun-phrase in sentence B} \vspace{0.1cm}\\
        & Sentence A: The cat had caught the mouse and it was trying to wriggle free.\\
        & Sentence B: The cat had caught the mouse and \textbf{<the cat>} was trying to wriggle free.\vspace{0.1cm}\\
        \midrule
        \multicolumn{2}{l}{\textbf{\textit{Word swap}}: the position of a word or text span in sentence A appears swapped in sentence B} \vspace{0.1cm} \\
        & Sentence A: Their music is considered by many as an alternative metal with rap metal and\\
        & \qquad\qquad\quad industrial metal influences, which according to previous interviews call\\
        & \qquad\qquad\quad themselves ``murder - rock''. \\
        & Sentence B: Their music is considered by many as \textbf{<industrial>} metal with rap metal and\\
        & \qquad\qquad\quad \textbf{<alternative>} metal influences. According to previous interviews, they\\
        & \qquad\qquad\quad consider themselves ``murder rock''.\vspace{0.1cm}\\
    \end{tabular}
    \caption{Manual annotation guidelines: Error type-specific guidelines}
    \label{tab:manual-annotation-type-specific}
\end{table}

\subsubsection{Development of Manual Annotation Guidelines}
\label{subsubsec:annotation_guideline_development}
To aid in the development and refinement of the annotation guidelines, we conducted a two-phase annotation pilot. In the first phase, we drew up the set of formal guidelines (described in Section~\ref{subsubsec:annotation_guidelines}). In the second phase, we verified the guidelines and measured inter-annotator agreement. We then asked professional annotators to complete the manual annotation of the four \textsc{ACES} phenomena listed above, using the guidelines. 

In the first pilot phase, four of the authors of the paper\footnote{Two annotators for the first pilot phase are native English speakers; two are fluent English speakers} manually annotated error spans for a sample of 100 examples with English as the target language, randomly selected across all phenomena in ACES. The annotators had access to the source-language sentence, the three target-language translations: good- incorrect- and reference-translation, and the phenomenon label. We considered only the target language side and marked one or more error spans in the incorrect translation only. We then conducted an adjudication exercise in which all four annotators manually compared the four sets of annotations for each example and discussed our strategies for annotation. From this, we derived a set of general guidelines to accommodate the annotation of any example in ACES. We then added specific guidelines for examples belonging to the categories: \textit{hallucination}, \textit{coreference}, and \textit{word swap}.

In the second pilot phase, we verified the quality of the manual annotation guidelines. To verify the general guidelines, and provide a gold standard against which to measure the automated evaluation method, the same four annotators from the first pilot phase annotated another sample of 100 examples with English as the target language, randomly selected across all \textsc{ACES} phenomena. 
To verify the quality of the span annotations, we automatically measured inter-annotator agreement. We computed the percentage of exact matches\footnote{We ignore both leading and trailing whitespace when comparing spans} as total\_exact\_matches divided by total\_spans\_marked, i.e. where all four annotators agree on the same error span, as 81.82\% (examples=100, total spans=110, exact-match spans=90), indicating high agreement\footnote{Highest inter-annotator agreement with three annotators: 90.48\% (examples=100, total spans=105, exact-match spans=95)}. We also verified the type-specific guidelines for annotating \textit{hallucination}, \textit{coreference}, and \textit{word swap}. As the \textit{coreference} category requires manual annotation in German (\textsc{ACES} contains only en-de examples for the \textit{coreference-based-on-commonsense} phenomenon), and examples of the other phenomena exist for English, we asked two native German / fluent English speakers\footnote{One annotator for the second pilot phase was also an author of this paper} to annotate a randomly selected sample of 100 examples (25 examples from each of the relevant \textsc{ACES} phenomenon categories). We report inter-annotator agreement of 77.40\% (examples=100, total spans=146, exact-match spans=113).

In addition to measuring inter-annotator agreement, we also examined the examples where two or more annotators marked different spans. We concluded that the majority of differences arose from simple human errors as opposed to differing interpretations of the guidelines. For example, annotators sometimes accidentally marked longer spans than necessary, or marked the presence of a deletion in the wrong position. We concluded that many of these mistakes could have been avoided had the annotators carefully double-checked their annotations. We therefore added a note to the guidelines to this effect, but made no further changes to the instructions. It is also worth noting that for a handful of examples, the presence of Machine Translation led to annotators struggling to agree on a correct annotation -- an issue that is not easily resolved, but is infrequent in the \textsc{ACES} dataset.

\section{Evaluation Methodology}
\label{sec:eval_methodology}

\begin{table}[h]
\centering
\small
\setlength{\tabcolsep}{5pt}
\setlength{\fboxsep}{0.5pt} 
 \resizebox{\textwidth}{!}{

\begin{tabular}{lcccccc}
\toprule
 & supervised & surface & base- & LLM- & & \\
 &  & overlap & embedding & based & 2022 & 2023 \\
\midrule
BLEU & & \cmark & & & \cmark & \cmark \\
f101spBLEU & & \cmark & & & \cmark & \\
f200spBLEU & & \cmark & & & \cmark & \cmark \\
chrF & & \cmark & & & \cmark & \cmark \\
BERTScore & & & ? & & \cmark & \cmark \\
BLEURT20 & WMT human eval & & BERT & & \cmark & \cmark \\
COMET-20 & & & XML-R & & \cmark & \\
COMET-QE & & & XML-R? & & \cmark & \\
YiSi-1 & & & ? & & \cmark & \cmark \\
Random-sysname & & & & & & \cmark \\
\midrule
COMET-22\textbf{\textcolor{darkpastelred}{*}}\textbf{\textcolor{darkpastelred}{$\dagger$}} & DA+MQM & & & & \cmark & \cmark \\
MATESE & MQM & & & & \cmark & \\
metricx\_xl\_DA\_2019 & DA & & mt5 & & \cmark & \\
metricx\_xl\_MQM\_2020 & MQM & & mt5 & & \cmark & \\
metricx\_xxl\_DA\_2019 & DA & & mt5 & & \cmark & \\
metricx\_xxl\_MQM\_2020 & MQM & & mt5 & & \cmark & \\
MS-COMET-22 & human judgements & & mt5 & & \cmark & \\ 
UniTE & & & & & \cmark & \\
UniTE-ref \textbf{\textcolor{darkpastelred}{$\dagger$}} & & & & & \cmark & \\
eBLEU & & & & & & \cmark \\
embed\_llama & & & Llama 2 & \cmark & & \cmark \\
MetricX-23 & DA+MQM & & mT5 & & & \cmark \\
MetricX-23-b & DA+MQM & & mT5 & & & \cmark \\
MetricX-23-c & DA+MQM & & mT5 & & & \cmark \\
partokengram\_F & & \cmark? & & & & \cmark \\
tokengram\_F & & \cmark & & & & \cmark \\
XCOMET-Ensemble & DA+MQM & & XLM-R & & & \cmark \\
XCOMET-XL \textbf{\textcolor{darkpastelred}{$\dagger$}} & DA+MQM & & XLM-R & & & \cmark \\
XCOMET-XXL & DA+MQM & & XLM-R & & & \cmark \\
XLsim & WMT human eval & & XLM-R & & & \cmark \\
\midrule
COMETKiwi\textbf{\textcolor{darkpastelred}{*}} & DA & & InfoXLM & & \cmark & \cmark \\
Cross-QE & & & ? & & \cmark & \\
HWTSC-Teacher-Sim & & & paraphrase-multilingual & & \cmark & \\
 & & & -mpnet-base-v2 & & & \\
HWTSC-TLM & & & ? & & \cmark & \\
KG-BERTScore & & & & & \cmark & \cmark \\
MATESE-QE & MQM & & & & \cmark & \\
MS-COMET-QE-22\textbf{\textcolor{darkpastelred}{*}} & & & & & \cmark & \cmark \\
REUSE & & & BERT & & \cmark & \\
UniTE-src & & & & & \cmark & \\
cometoid22-wmt21 & ? & & InfoXLM & & & \cmark \\
cometoid22-wmt22 & ? & & InfoXLM & & & \cmark \\
cometoid22-wmt23 & ? & & InfoXLM & & & \cmark \\
CometKiwi-XL & & & XLM-R & & & \cmark \\
CometKiwi-XXL & & & XLM-R & & & \cmark \\
GEMBA-MQM \textbf{\textcolor{darkpastelred}{$\dagger$}} & & & & \cmark & & \cmark \\
MetricX-23-QE & DA+MQM & & mT5 & & & \cmark \\
MetricX-23-QE-b & DA+MQM & & mT5 & & & \cmark \\
MetricX-23-QE-c & DA+MQM & & mT5 & & & \cmark \\
XCOMET-QE-Ensemble & DA+MQM & & XLM-R & & & \cmark \\
XLsimQE & WMT human eval & & XLM-R & & & \cmark \\
\bottomrule
\end{tabular}}
\caption{Basline (top), reference-based (middle), and reference-free (bottom) metrics from WMT 2022 and 2023 Metrics shared tasks. \textbf{\textcolor{darkpastelred}{*}} denotes a participating metric from 2022 that was used as a baseline in 2023. \textbf{\textcolor{darkpastelred}{$\dagger$}} denotes that metrics were used as baselines for \textsc{Span-ACES}. ? indicates no information was made available.}
\label{tab:metrics_overview}
\end{table}

Table~\ref{tab:metrics_overview} lists the baseline, reference-based, and reference-free metrics from WMT 2022 and 2023 that provide segment-level judgements and cover all of the language pairs in ACES. We indicate whether metrics are embeddings-based with a subset of metrics using the supervision signal provided by Direct Assessment (DA) judgements from WMT \citep{bojar-etal-2016-findings} or MQM \citep{lommel2014} annotations, LLM-based, or rely on surface-level overlap with the reference. 

We briefly summarise the metrics here, grouping them into broad categories based on their design characteristics. The metrics that \textit{\textbf{rely on surface overlap with the reference}} include several baseline metrics: \textbf{BLEU} \citep{papineni-etal-2002-bleu}, \textbf{chrF} \citep{popovic-2017-chrf} and the \textbf{spBLEU} \citep{goyal-etal-2022-flores} metrics \textsc{f101spBLEU} and \textsc{f200spBLEU}, for which the SentencePiece tokeniser \citep{kudo-richardson-2018-sentencepiece} was trained using data from the FLORES-101 or -200 languages respectively. It also includes the 2023 participant metrics based on F-scores and inspired by chrF++: \textbf{Tokengram\_F} and \textbf{Partokengram\_F} \citep{dreano-molloy-murphy:2023:WMT1}. 

The largest group is \textit{\textbf{embedding-based metrics}}. 
Many are based on the \textbf{COMET} architecture: \textbf{COMET-20} and \textbf{COMET-QE} \citep{rei-etal-2020-comet}, Unbabel's WMT 2022 submission \textbf{COMET-22} \citep{COMET:WMT22}, and Microsoft's WMT 2022 submissions \textbf{MS-COMET-22} and \textbf{MS-COMET-QE-22} \citep{MS-COMET:WMT22}. The \textbf{XCOMET} family of metrics, trained to identify errors in sentences along with a final quality score, includes \textbf{XCOMET-XL}, \textbf{XCOMET-XXL}, and \textbf{XCOMET-QE}, and the two ensemble metrics: \textbf{XCOMET-Ensemble} and \textbf{XCOMET-QE-Ensemble}. The \textbf{COMET-Kiwi} \citep{COMET:WMT22} metric and the \textbf{COMETKiwi-XL} and \textbf{COMETKiwi-XXL} metrics from 2023 form another family. The \textbf{Cometoid22} \citep{gowda-kocmi-junczysdowmunt:2023:WMT} student metrics are trained to mimic teacher scores from COMET-22 without access to the reference. (The suffix [WMT-21,22,23] indicates the training data cut-off year.) The remaining metrics are based on a range of different architectures: \textbf{BERTScore} \citep{DBLP:conf/iclr/ZhangKWWA20}, \textbf{BLEURT20} \citep{sellam-etal-2020-learning}, \textbf{YiSi-1} \citep{lo-2019-yisi}, \textbf{UniTE} \citep{UNITE:WMT22}, \textbf{MATESE} and \textbf{MATESE-QE} \citep{MATESE:WMT22}, 
\textbf{REUSE} \citep{REUSE:WMT22}, \textbf{eBLEU} \citep{elnokrashy-kocmi:2023:WMT}, and \textbf{XLsim} \citep{mukherjee-shrivastava:2023:WMT2}. The \textbf{MetricX} family includes the \textbf{metricx\_*\_DA} and \textbf{metricx\_*\_MQM} metrics from 2022 and \textbf{MetricX-23} and \textbf{MetricX-23-QE} \citep{juraska-EtAl:2023:WMT} from 2023. The Huawei metrics include \textbf{Cross-QE}, \textbf{HWTSC-Teacher-Sim}, and \textbf{HWTSC-TLM} \citep{HWTSC-Metrics:WMT22}, and \textbf{KG-BERTScore} \citep{HWTSC-Metrics:WMT22,wu-EtAl:2023:WMT4} which incorporates a multilingual knowledge graph.

The \textit{\textbf{LLM-based}} metrics group comprises two WMT 2023 metrics: \textbf{Embed\_Llama} \citep{dreano-molloy-murphy:2023:WMT2} which uses pre-trained LLaMA2 embeddings without finetuning, and \textbf{GEMBA-MQM} \citep{kocmi-federmann:2023:WMT} -- a GPT-based metric for error quality span marking. Finally, \textbf{Random-sysname} is a random baseline which samples scores from a Gaussian distribution based on random mean value. It was included in 2023 to provide a context to scores and also to detect errors in metric meta-evaluations.
In addition to these metrics, we also conducted some experiments on using LLMs for evaluation as listed below.

\subsection{LLM Metrics}
\label{subsec:llm_metrics}
Following the rapid adoption of LLM-based approaches to address a range of NLP tasks, there has also been a steady increase in the use of LLMs for MT evaluation with apparently promising results \citep{xu-etal-2023-instructscore, lu2023error, kocmi-federmann:2023:WMT}. We note that these observations are often limited to system-level evaluation and also to high-resource language pairs.  
We thus intend to investigate the extent to which these LLMs can be used for MT evaluation more holistically through the \textsc{ACES} dataset.

We consider three variants of using LLMs for evaluation. The first one is \textsc{GEMBA-DA} \citep{kocmi-federmann:2023:WMT} where the model (GPT Davinci-003 a predecessor to GPT-4 model) is prompted using a zero-shot approach to produce a translation score between 0 and 100. Note that \textsc{GEMBA-DA} was the precursor of the \textsc{GEMBA-MQM} model, which was discussed previously. 
For the next two methods, we considered LLaMA2 (7B) \citep{llama2} and \textsc{Flan-Alpaca-XL} \citep{chia2023instructeval} (3B) which is Flan-T5 \citep{chung2022scaling} fine-tuned on the Alpaca dataset \citep{alpaca}. We chose LLaMA2 (7B), despite it being predominantly trained in English, to see if the accidental multilingual tokens are enough to provide multilingual evaluation. We included \textsc{Flan-Alpaca-XL} as it is a \textit{smaller} LLM and the base model was trained with multilingual data. \footnote{We also conducted experiments on BLOOM \citep{DBLP:journals/corr/abs-2211-05100} but found the majority of outputs produced by the BLOOM-7B model to be unintelligible which could not be converted into scores}

For these LLMs (\textsc{Flan-Alpaca-XL} and \textsc{LLaMA2}), we experimented with both zero-shot and five-shot prompting. In five-shot prompting, five examples of scored translations across varying scoring ranges and language pairs were provided with the prompt. However, we found that five-shot prompting performed poorly in our initial experiments and therefore we report only the zero-shot results. We provide the prompt templates in the Appendix. For the postprocessing of outputs from the above LLMs, we included the first rational number that appeared in the output from the respective models as the \textit{score} produced by that LLM. In the scenario in which no number was found, the example was given a score of 0. In such examples, the overgenerated text generally consisted of a hallucinated example of a source-reference-translation triplet.

As \textsc{ACES} is a contrastive dataset, we also experimented with providing a prompt that compares the two translations, labelled A and B respectively, and instructs the LLM to select the \textit{better} translation. However, in our initial experiments, we found that the models typically produce an option followed by the generation of both of the candidate translations. This copying of translations makes it hard to identify if the generation of the option was a result of the model actually performing the evaluation or an artefact of the overgeneration.

\subsection{Metrics with error spans}
\label{subsec:span_baselines}
In addition to the above metrics, we also conduct baseline experiments for \textsc{Span-ACES}. We include recently developed metrics that directly predict error spans while generating the scores, namely \textsc{XCOMET-XL} \citep{guerreiro2023xcomet} and \textsc{GEMBA-MQM} \citep{kocmi-federmann:2023:WMT}. These metrics also provide severity of the error for the predicted error span - minor, major, and critical.

Additionally, we derive baselines from existing metrics that were trained to only produce scores. We re-purpose the work in \citet{rei-etal-2023-inside}, which included the proposal of several neural explainability methods for interpreting state-of-the-art fine-tuned neural machine translation metrics such as \textsc{COMET} and \textsc{UniTE}. In one of these techniques, \textit{embed–align}, they calculate the maximum cosine similarity between each translation token embedding and the reference and/or source token embeddings \citep{tao-etal-2022-crossqe} and assign that scalar value to each translation token. Starting from embed-align scores attributed to each translation token, we generate error spans over the translations by marking any token which has an embed-align score higher than a constant threshold. We set the threshold that yields the span predictions with the highest Recall@K score on the WMT 2021 MQM annotations development dataset\footnote{threshold=0.1 for COMET-22, threshold=0.14 for UNiTE}. This method produces six different types of span predictions: embed–align[mt, src], embed–align[mt, ref] and embed–align[mt, src; ref] using the embeddings extracted from each of the \textsc{COMET-22} and \textsc{UniTE} models \footnote{We use the wmt22-comet-da version for \textsc{COMET-22} and \textsc{src+ref} version for \textsc{UniTE}}.

\subsection{Evaluation of Metrics}
\label{subsec:kendall_tau_formula}
For all phenomena in \textsc{ACES} where we generated more than 1,000 examples, we randomly subsample 1,000 examples according to the per language pair distribution to include in the final challenge set to keep the evaluation of new metrics tractable.

We follow the evaluation of the challenge sets from the 2021 edition of the WMT metrics shared task \citep{freitag-etal-2021-results} and report performance with Kendall's tau-like correlation\footnote{Evaluation scripts are available here: \url{https://github.com/EdinburghNLP/ACES}}. The Kendall's tau-like metric (see Equation~\ref{eq:kendal-tau}) measures the number of times a metric scores the good translation above the incorrect translation (concordant) and equal to or lower than the incorrect translation (discordant). Ties are considered as discordant. Note that a higher $\tau$ indicates a better performance and that the values can range between -1 and 1. 

\vspace{-20pt}
\begin{center}
\begin{equation}
\begin{aligned}
    \tau = \frac{concordant - discordant}{concordant + discordant}\\
\end{aligned}
\label{eq:kendal-tau}
\end{equation}
\end{center}

We discuss the evaluation on \textsc{Span-ACES} closer to its results section.

\section{Results}
\label{sec: Results}

\subsection{Phenomena-level Results}
\begin{sidewaystable*}[ht]
\small
\setlength{\tabcolsep}{3.75pt}
\centering
\begin{tabular}{@{}lccccccccccc@{}}
\toprule
 & \hyperref[sec:addition-omission]{\textbf{addition}} & \hyperref[sec:addition-omission]{\textbf{omission}} & \hyperref[sec:source-disambig]{\textbf{mistrans.}} & \hyperref[sec:untranslated]{\textbf{untranslated}} & \hyperref[sec:do-not-translate]{\textbf{do not}} & \hyperref[sec:overtranslation_undertranslation]{\textbf{overtrans.}} & \hyperref[sec:overtranslation_undertranslation]{\textbf{undertrans.}} & \hyperref[sec:real-world-knowledge]{\textbf{real-world}} & \hyperref[sec:wrong_language]{\textbf{wrong}} & \hyperref[sec:punctuation]{\textbf{punctuation}} & \textbf{ACES-}\\ 
&  &  &  &  & \hyperref[sec:do-not-translate]{\textbf{translate}} &  &  & \hyperref[sec:real-world-knowledge]{\textbf{knowledge}} & \hyperref[sec:wrong_language]{\textbf{language}} &  & \textbf{Score}\\ 
\midrule
\textit{\textbf{Examples}}     & \textit{999}                 & \textit{999}                 & \textit{24457}                     & \textit{1300}                    & \textit{100}                         & \textit{1000}                       & \textit{1000}                        & \textit{2948}                            & \textit{2000}                      & \textit{1673}                   \\ 
\midrule
BLEU                    & \phantom{-}0.748    & \phantom{-}0.435    & -0.229         & \phantom{-}0.353        & \phantom{-}0.600            & -0.838          & -0.856           & -0.768               & \phantom{-}0.661          & \phantom{-}0.638       & -2.7      \\
f101spBLEU              & \phantom{-}0.662    & \phantom{-}0.590    & -0.084         & \phantom{-}0.660        & \phantom{-}0.940            & -0.738          & -0.826           & -0.405               & \phantom{-}0.638          & \phantom{-}0.639       & -0.1      \\
f200spBLEU              & \phantom{-}0.664    & \phantom{-}0.590    & -0.082         & \phantom{-}0.687        & \phantom{-}0.920            & -0.752          & -0.794           & -0.394               & \phantom{-}0.658          & \phantom{-}0.648       & \phantom{-}0.1     \\
chrF                    & \phantom{-}0.642    & \phantom{-}0.784    & \phantom{-}0.162          & \colorbox[HTML]{B2EAB1}{\textbf{\phantom{-}0.781}}        & \colorbox[HTML]{B2EAB1}{\textbf{\phantom{-}0.960}}            & -0.696          & -0.592           & -0.294               & \colorbox[HTML]{B2EAB1}{\textbf{\phantom{-}0.691}}          & \phantom{-}0.743       & \phantom{1}3.7       \\
BERTScore               & \colorbox[HTML]{B2EAB1}{\textbf{\phantom{-}0.880}}    & \phantom{-}0.750    & \phantom{-}0.320          & \phantom{-}0.767        & \colorbox[HTML]{B2EAB1}{\textbf{\phantom{-}0.960}}            & -0.110          & -0.190           & \phantom{-}0.031                & \phantom{-}0.563          & \colorbox[HTML]{B2EAB1}{\textbf{\phantom{-}0.849}}       & 10.6      \\
BLEURT-20               & \phantom{-}0.437    & \phantom{-}0.810    & \phantom{-}0.429          & \phantom{-}0.748        & \phantom{-}0.860            & \phantom{-}0.200           & \phantom{-}0.014            & \phantom{-}0.401                & \phantom{-}0.533          & \phantom{-}0.649       & 12.0      \\
COMET-20                & \phantom{-}0.437    & \phantom{-}0.808    & \phantom{-}0.378          & \phantom{-}0.748        & \phantom{-}0.900            & \phantom{-}0.314           & \phantom{-}0.112            & \phantom{-}0.267                & \phantom{-}0.033          & \phantom{-}0.706       & 12.2      \\
COMET-QE                & -0.538   & \phantom{-}0.397    & \phantom{-}0.378          & \phantom{-}0.135        & \phantom{-}0.120            & \phantom{-}0.622           & \phantom{-}0.442            & \phantom{-}0.322                & -0.505         & \phantom{-}0.251       & \phantom{1}6.6       \\
YiSi-1                  & \phantom{-}0.770    & \phantom{-}0.866    & \phantom{-}0.356          & \phantom{-}0.730        & \phantom{-}0.920            & -0.062          & -0.076           & \phantom{-}0.110                & \phantom{-}0.431          & \phantom{-}0.734       & 11.5      \\
\midrule
COMET-22                & \phantom{-}0.333    & \phantom{-}0.806    & \phantom{-}0.566          & \phantom{-}0.536        & \phantom{-}0.900            & \phantom{-}0.690           & \phantom{-}0.538            & \phantom{-}0.574                & -0.318         & \phantom{-}0.539       & 16.4      \\
metricx\_xl\_DA\_2019   & \phantom{-}0.395    & \phantom{-}0.852    & \phantom{-}0.545          & \phantom{-}0.722        & \phantom{-}0.940            & \phantom{-}0.692           & \phantom{-}0.376            & \colorbox[HTML]{B2EAB1}{\textbf{\phantom{-}0.740}}                & \phantom{-}0.521          & \phantom{-}0.670       & 17.2      \\
metricx\_xl\_MQM\_2020  & -0.281   & \phantom{-}0.670    & \phantom{-}0.523          & \phantom{-}0.579        & {-}0.740            & \phantom{-}0.718           & \colorbox[HTML]{B2EAB1}{\textbf{\phantom{-}0.602}}            & \phantom{-}0.705                & -0.126         & \phantom{-}0.445       & 13.1      \\
metricx\_xxl\_DA\_2019  & \phantom{-}0.303    & \phantom{-}0.832    & \phantom{-}0.580          & \phantom{-}0.762        & \phantom{-}0.920            & \phantom{-}0.572           & \phantom{-}0.246            & \phantom{-}0.691                & \phantom{-}0.250          & \phantom{-}0.630       & 15.3      \\
metricx\_xxl\_MQM\_2020 & -0.099   & \phantom{-}0.534    & \phantom{-}0.578          & \phantom{-}0.651        & \phantom{-}0.880            & \colorbox[HTML]{B2EAB1}{\textbf{\phantom{-}0.752}}           & \phantom{-}0.552            & \phantom{-}0.712                & -0.321         & \phantom{-}0.369       & 13.5      \\
MS-COMET-22             & -0.219   & \phantom{-}0.686    & \phantom{-}0.397          & \phantom{-}0.504        & \phantom{-}0.700            & \phantom{-}0.548           & \phantom{-}0.290            & \phantom{-}0.230                & \phantom{-}0.041          & \phantom{-}0.508       & 10.0      \\
UniTE                   & \phantom{-}0.439    & \phantom{-}0.876    & \phantom{-}0.501          & \phantom{-}0.571        & \phantom{-}0.920            & \phantom{-}0.496           & \phantom{-}0.302            & \phantom{-}0.624                & -0.337         & \phantom{-}0.793       & 14.9      \\
UniTE-ref               & \phantom{-}0.359    & \phantom{-}0.868    & \phantom{-}0.535          & \phantom{-}0.412        & \phantom{-}0.840            & \phantom{-}0.640           & \phantom{-}0.398            & \phantom{-}0.585                & -0.387         & \phantom{-}0.709       & 15.5      \\
\midrule
COMETKiwi               & \phantom{-}0.361    & \phantom{-}0.830    & \colorbox[HTML]{B2EAB1}{\textbf{\phantom{-}0.631}}          & \phantom{-}0.230        & \phantom{-}0.780            & \phantom{-}0.738           & \phantom{-}0.574            & \phantom{-}0.582                & -0.359         & \phantom{-}0.490       & 16.9      \\
Cross-QE                & \phantom{-}0.163    & \phantom{-}0.876    & \phantom{-}0.546          & -0.094       & \phantom{-}0.320            & \phantom{-}0.726           & \phantom{-}0.506            & \phantom{-}0.446                & -0.374         & \phantom{-}0.455       & 14.4      \\
HWTSC-Teacher-Sim       & -0.031   & \phantom{-}0.495    & \phantom{-}0.406          & -0.269       & \phantom{-}0.700            & \phantom{-}0.552           & \phantom{-}0.456            & \phantom{-}0.261                & -0.021         & \phantom{-}0.271       & 10.1      \\
HWTSC-TLM               & -0.363   & \phantom{-}0.345    & \phantom{-}0.384          & \phantom{-}0.154        & -0.040           & \phantom{-}0.544           & \phantom{-}0.474            & \phantom{-}0.071                & -0.168         & \phantom{-}0.634       & \phantom{1}7.0       \\
KG-BERTScore            & \phantom{-}0.790    & \phantom{-}0.812    & \phantom{-}0.489          & -0.456       & \phantom{-}0.760            & \phantom{-}0.654           & \phantom{-}0.528            & \phantom{-}0.487                & \phantom{-}0.306          & \phantom{-}0.255       & \colorbox[HTML]{B2EAB1}{\textbf{17.5}}      \\
MS-COMET-QE-22          & -0.177   & \phantom{-}0.678    & \phantom{-}0.439          & \phantom{-}0.388        & \phantom{-}0.240            & \phantom{-}0.518           & \phantom{-}0.386            & \phantom{-}0.248                & -0.197         & \phantom{-}0.523       & \phantom{1}9.9       \\
UniTE-src               & \phantom{-}0.285    & \colorbox[HTML]{B2EAB1}{\textbf{\phantom{-}0.930}}    & \phantom{-}0.599          & -0.615       & \phantom{-}0.860            & \phantom{-}0.698           & \phantom{-}0.540            & \phantom{-}0.537                & -0.417         & \phantom{-}0.733       & 15.7      \\
\midrule
Average                 & \phantom{-}0.290    & \phantom{-}0.713    & \phantom{-}0.389          & \phantom{-}0.404        & \phantom{-}0.735            & \phantom{-}0.312           & \phantom{-}0.167            & \phantom{-}0.282                & \phantom{-}0.075          & \phantom{-}0.578       & 10.9  \\   
\bottomrule
\end{tabular}
\caption{2022 Results. Average Kendall’s tau-like correlation results for the nine top level categories in the \textsc{ACES} ontology, plus the additional fluency category: punctuation.  The horizontal lines delimit baseline metrics (top), participating reference-based metrics (middle) and participating reference-free metrics (bottom). The best result for each category is denoted by bold text with a green highlight. Note that \textit{Average} is an average over averages. The last column shows the \textsc{ACES}-Score, a weighted sum of the correlations. The \textsc{ACES}-Score ranges from -29.1 (all phenomena have a correlation of -1) to 29.1 (all phenomena have a correlation of +1).}
\label{tab:analysis_overview_2022}
\end{sidewaystable*}
\afterpage{\clearpage}

\begin{sidewaystable*}[ht]
\small
\setlength{\tabcolsep}{3.75pt}
\centering
\begin{tabular}{@{}lccccccccccc@{}}
\toprule
 & \hyperref[sec:addition-omission]{\textbf{addition}} & \hyperref[sec:addition-omission]{\textbf{omission}} & \hyperref[sec:source-disambig]{\textbf{mistrans.}} & \hyperref[sec:untranslated]{\textbf{untranslated}} & \hyperref[sec:do-not-translate]{\textbf{do not}} & \hyperref[sec:overtranslation_undertranslation]{\textbf{overtrans.}} & \hyperref[sec:overtranslation_undertranslation]{\textbf{undertrans.}} & \hyperref[sec:real-world-knowledge]{\textbf{real-world}} & \hyperref[sec:wrong_language]{\textbf{wrong}} & \hyperref[sec:punctuation]{\textbf{punctuation}} & \textbf{ACES-}\\ 
&  &  &  &  & \hyperref[sec:do-not-translate]{\textbf{translate}} &  &  & \hyperref[sec:real-world-knowledge]{\textbf{knowledge}} & \hyperref[sec:wrong_language]{\textbf{language}} &  & \textbf{Score}\\ 
\midrule
\textit{Examples}           & \textit{999}                          & \textit{999}                          & \textit{24457}                              & \textit{1300}                             & \textit{100}                                  & \textit{1000}                                & \textit{1000}                                 & \textit{2948}                                     & \textit{2000}                               & \textit{1673}                            &            \\
\midrule
BERTscore          & \colorbox[HTML]{B2EAB1}{\textbf{\phantom{-}0.872}}                        & \phantom{-}0.754                        & \phantom{-}0.318                              & \phantom{-}0.771                            & \phantom{-}0.940                                & -0.186                              & -0.288                               & \phantom{-}0.030                                    & \phantom{-}0.551                              & \colorbox[HTML]{B2EAB1}{\textbf{\phantom{-}0.844}}                           & \phantom{1}9.7                          \\
BLEU               & \phantom{-}0.742                        & \phantom{-}0.427                        & -0.227                             & \phantom{-}0.353                            & \phantom{-}0.580                                & -0.838                              & -0.856                               & -0.768                                   & \phantom{-}0.660                              & \phantom{-}0.704                           & -2.8                         \\
BLEURT-20          & \phantom{-}0.435                        & \phantom{-}0.812                        & \phantom{-}0.427                              & \phantom{-}0.743                            & \phantom{-}0.860                                & \phantom{-}0.202                               & \phantom{-}0.014                                & \phantom{-}0.388                                    & \phantom{-}0.536                              & \phantom{-}0.708                           & 12.0                         \\
chrF               & \phantom{-}0.644                        & \phantom{-}0.784                        & \phantom{-}0.162                              & \colorbox[HTML]{B2EAB1}{\textbf{\phantom{-}0.781}}                            & \colorbox[HTML]{B2EAB1}{\textbf{\phantom{-}0.960}}                                & -0.696                              & -0.592                               & -0.294                                   & \phantom{-}0.693                              & \phantom{-}0.773                           & \phantom{-}3.7                          \\
COMET-22           & \phantom{-}0.295                        & \phantom{-}0.822                        & \phantom{-}0.402                              & \phantom{-}0.718                            & \phantom{-}0.820                                & \phantom{-}0.502                               & \phantom{-}0.258                                & \phantom{-}0.382                                    & \phantom{-}0.078                              & \phantom{-}0.673                           & 13.458                         \\
CometKiwi          & \phantom{-}0.536                        & \colorbox[HTML]{B2EAB1}{\textbf{\phantom{-}0.918}}                        & \phantom{-}0.614                              & -0.105                           & \phantom{-}0.520                                & \phantom{-}0.766                               & \phantom{-}0.604                                & \phantom{-}0.577                                    & -0.307                             & \phantom{-}0.765                           & \colorbox[HTML]{B2EAB1}{\textbf{17.9}}                         \\
f200spBLEU         & \phantom{-}0.666                        & \phantom{-}0.584                        & -0.082                             & \phantom{-}0.680                            & \phantom{-}0.920                                & -0.752                              & -0.794                               & -0.394                                   & \phantom{-}0.657                              & \phantom{-}0.708                           & \phantom{1}0.041                          \\
MS-COMET-QE-22     & -0.179                       & \phantom{-}0.674                        & \phantom{-}0.440                              & \phantom{-}0.394                            & \phantom{-}0.300                                & \phantom{-}0.524                               & \phantom{-}0.382                                & \phantom{-}0.262                                    & -0.195                             & \phantom{-}0.632                           & 10.0                         \\
Random-sysname     & -0.117                       & -0.117                       & -0.116                             & -0.083                           & -0.100                               & -0.118                              & -0.152                               & -0.245                                   & -0.113                             & -0.074                          & -3.6                         \\
YiSi-1             & \phantom{-}0.766                        & \phantom{-}0.868                        & \phantom{-}0.354                              & \phantom{-}0.720                            & \phantom{-}0.940                                & -0.062                              & -0.076                               & \phantom{-}0.110                                    & \phantom{-}0.421                              & \phantom{-}0.763                           & 11.5                         \\
\midrule
eBLEU              & \phantom{-}0.674                        & \phantom{-}0.682                        & \phantom{-}0.197                              & \phantom{-}0.739                            & \phantom{-}0.880                                & -0.662                              & -0.684                               & -0.042                                   & \colorbox[HTML]{B2EAB1}{\textbf{\phantom{-}0.771}}                              & \phantom{-}0.270                           & \phantom{1}3.4                          \\
embed\_llama       & \phantom{-}0.211                        & \phantom{-}0.457                        & \phantom{-}0.016                              & \phantom{-}0.503                            & \phantom{-}0.400                                & -0.170                              & -0.492                               & -0.165                                   & \phantom{-}0.154                              & \phantom{-}0.476                           & \phantom{1}1.054                          \\
MetricX-23         & -0.027                       & \phantom{-}0.568                        & \phantom{-}0.578                              & \phantom{-}0.473                            & \phantom{-}0.800                                & \phantom{-}0.790                               & \phantom{-}0.586                                & \phantom{-}0.766                                    & -0.486                             & \phantom{-}0.636                           & 14.1                         \\
MetricX-23-b       & -0.135                       & \phantom{-}0.622                        & \phantom{-}0.572                              & \phantom{-}0.613                            & \phantom{-}0.860                                & \phantom{-}0.772                               & \phantom{-}0.568                                & \phantom{-}0.749                                    & -0.444                             & \phantom{-}0.532                           & 13.8                         \\
MetricX-23-c       & -0.015                       & \phantom{-}0.794                        & \phantom{-}0.617                              & \phantom{-}0.611                            & \phantom{-}0.800                                & \phantom{-}0.740                               & \phantom{-}0.526                                & \colorbox[HTML]{B2EAB1}{\textbf{\phantom{-}0.783}}                                    & -0.629                             & \phantom{-}0.527                           & 15.0                        \\
partokengram\_F    & \phantom{-}0.087                        & \phantom{-}0.191                        & -0.034                             & \phantom{-}0.310                            & \phantom{-}0.140                                & -0.042                              & -0.028                               & \phantom{-}0.032                                    & \phantom{-}0.508                              & \phantom{-}0.171                           & \phantom{1}1.9                          \\
tokengram\_F       & \phantom{-}0.698                        & \phantom{-}0.758                        & \phantom{-}0.160                              & \phantom{-}0.779                            & \colorbox[HTML]{B2EAB1}{\textbf{\phantom{-}0.960}}                                & -0.732                              & -0.632                               & -0.273                                   & \phantom{-}0.687                              & \phantom{-}0.830                           & \phantom{1}3.5                          \\
XCOMET-Ensemble    & \phantom{-}0.311                        & \phantom{-}0.786                        & \phantom{-}0.663                              & \phantom{-}0.379                            & \phantom{-}0.780                                & \colorbox[HTML]{B2EAB1}{\textbf{\phantom{-}0.794}}                               & \phantom{-}0.612                                & \phantom{-}0.708                                    & -0.423                             & \phantom{-}0.595                           & 17.3                         \\
XCOMET-XL          & \phantom{-}0.169                        & \phantom{-}0.542                        & \phantom{-}0.570                              & \phantom{-}0.222                            & \phantom{-}0.800                                & \phantom{-}0.656                               & \phantom{-}0.464                                & \phantom{-}0.582                                    & -0.367                             & \phantom{-}0.220                           & 13.3                         \\
XCOMET-XXL         & -0.119                       & \phantom{-}0.413                        & \phantom{-}0.547                              & \phantom{-}0.234                            & \phantom{-}0.600                                & \phantom{-}0.736                               & \phantom{-}0.568                                & \phantom{-}0.508                                    & -0.507                             & \phantom{-}0.509                           & 11.6                         \\
XLsim              & \phantom{-}0.429                        & \phantom{-}0.618                        & \phantom{-}0.153                              & \phantom{-}0.643                            & \phantom{-}0.820                                & -0.210                              & -0.290                               & -0.044                                   & \phantom{-}0.392                              & \phantom{-}0.753                           & \phantom{1}5.4                          \\
\midrule
cometoid22-wmt21   & -0.339                       & \phantom{-}0.658                        & \phantom{-}0.493                              & -0.076                           & \phantom{-}0.280                                & \phantom{-}0.670                               & \phantom{-}0.566                                & \phantom{-}0.362                                    & -0.454                             & \phantom{-}0.608                           & 10.4                        \\
cometoid22-wmt22   & -0.301                       & \phantom{-}0.674                        & \phantom{-}0.493                              & -0.119                           & \phantom{-}0.280                                & \phantom{-}0.686                               & \phantom{-}0.538                                & \phantom{-}0.340                                    & -0.472                             & \phantom{-}0.599                           & 10.534                         \\
cometoid22-wmt23   & -0.253                       & \phantom{-}0.702                        & \phantom{-}0.502                              & -0.046                           & \phantom{-}0.420                                & \phantom{-}0.750                               & \phantom{-}0.590                                & \phantom{-}0.362                                    & -0.319                             & \phantom{-}0.557                           & 11.9                        \\
CometKiwi-XL       & \phantom{-}0.239                        & \phantom{-}0.828                        & \phantom{-}0.624                              & \phantom{-}0.239                            & \phantom{-}0.440                                & \phantom{-}0.762                               & \phantom{-}0.560                                & \phantom{-}0.563                                    & -0.380                             & \phantom{-}0.630                           & 16.0                        \\
CometKiwi-XXL      & \phantom{-}0.361                        & \phantom{-}0.828                        & \phantom{-}0.653                              & \phantom{-}0.414                            & \phantom{-}0.320                                & \phantom{-}0.774                               & \phantom{-}0.560                                & \phantom{-}0.683                                    & -0.537                             & \phantom{-}0.503                           & 16.8                         \\
GEMBA-MQM          & \phantom{-}0.037                        & \phantom{-}0.281                        & \phantom{-}0.153                              & \phantom{-}0.094                            & \phantom{-}0.140                                & \phantom{-}0.466                               & \phantom{-}0.276                                & \phantom{-}0.268                                    & -0.150                             & \phantom{-}0.015                           & \phantom{1}6.4                          \\
KG-BERTScore       & \phantom{-}0.538                        & \phantom{-}0.912                        & \phantom{-}0.585                              & -0.206                           & \phantom{-}0.700                                & \phantom{-}0.772                               & \phantom{-}0.606                                & \phantom{-}0.594                                    & -0.307                             & \phantom{-}0.654                           & \colorbox[HTML]{B2EAB1}{\textbf{18.0}}                         \\
MetricX-23-QE      & \phantom{-}0.045                        & \phantom{-}0.678                        & \phantom{-}0.654                              & \phantom{-}0.379                            & \phantom{-}0.460                                & \phantom{-}0.772                               & \phantom{-}0.612                                & \phantom{-}0.654                                    & -0.702                             & \phantom{-}0.226                           & 14.6                         \\
MetricX-23-QE-b    & \phantom{-}0.027                        & \phantom{-}0.760                        & \phantom{-}0.663                              & \phantom{-}0.489                            & \phantom{-}0.480                                & \phantom{-}0.758                               & \colorbox[HTML]{B2EAB1}{\textbf{\phantom{-}0.620}}                                & \phantom{-}0.647                                    & -0.673                             & \phantom{-}0.256                           & 15.1                         \\
MetricX-23-QE-c    & -0.115                       & \phantom{-}0.664                        & \colorbox[HTML]{B2EAB1}{\textbf{\phantom{-}0.721}}                              & \phantom{-}0.384                            & \phantom{-}0.340                                & \phantom{-}0.726                               & \phantom{-}0.618                                & \phantom{-}0.753                                    & -0.712                             & \phantom{-}0.375                           & 13.8                         \\
XCOMET-QE-Ensemble & \phantom{-}0.277                        & \phantom{-}0.754                        & \phantom{-}0.644                              & \phantom{-}0.181                            & \phantom{-}0.720                                & \phantom{-}0.764                               & \phantom{-}0.582                                & \phantom{-}0.626                                    & -0.519                             & \phantom{-}0.449                           & 16.1                         \\

XLsimQE            & \phantom{-}0.205                        & \phantom{-}0.383                        & \phantom{-}0.087                              & -0.694                           & \phantom{-}0.940                                & \phantom{-}0.454                               & \phantom{-}0.352                                & \phantom{-}0.042                                    & \phantom{-}0.307                              & \phantom{-}0.671                           & \phantom{1}8.0                         \\
\midrule
Average            & \phantom{-}0.232                        & \phantom{-}0.639                        & \phantom{-}0.382                              & \phantom{-}0.349                            & \phantom{-}0.609                                & \phantom{-}0.314                               & \phantom{-}0.187                                & \phantom{-}0.289                                    & -0.069                             & \phantom{-}0.532                           & 10.0                        
\\
\bottomrule
\end{tabular}
\caption{2023 Results. Average Kendall’s tau-like correlation results for the \textsc{ACES} top-level categories and \textsc{ACES}-Scores (final column). Metrics are grouped into baseline (top), and participating reference-based (middle) and reference-free (bottom) metrics. Note that \textit{Average} is an average over averages. Best results are highlighted in green.}
\label{tab:analysis_overview_2023}
\end{sidewaystable*}
\afterpage{\clearpage}

\label{sec:aces_overview}
We begin by providing a broad overview of metric performance on the different phenomena categories, before conducting more detailed analyses in Section~\ref{sec:analysis}. We restrict the overview to the metrics which provide a) segment-level scores and b) scores for all language pairs and directions in \textsc{ACES}. After filtering according to these criteria, 24 metrics from 2022 remain: nine baseline, eight reference-based, and seven reference-free metrics. In 2023, 33 metrics fulfil these criteria: 10 baseline, 11 reference-based, and 12 reference-free metrics.

We first calculate Kendall’s tau-like correlation scores for all of the \textsc{ACES} examples (see Equation~\ref{eq:kendal-tau}). We then report the average score over all examples belonging to each of the nine top-level accuracy categories in \textsc{ACES}, plus the fluency category \textit{punctuation} (see Tables~\ref{tab:analysis_overview_2022} and~\ref{tab:analysis_overview_2023}). 
In addition, we calculate the \textsc{ACES}-Score, a weighted combination of the top-level categories, which allows us to identify high-level performance trends of the metrics (see Equation~\ref{eq:aces-score}). The weights correspond to the values under the MQM framework \citep{freitag-etal-2021-experts} for major (weight$=$5), minor (weight$=$1) and fluency/punctuation errors (weight$=$0.1). We categorise \hyperref[sec:untranslated]{untranslated}, \hyperref[sec:do-not-translate]{do not translate} and \hyperref[sec:wrong_language]{wrong language} as minor errors due to the ease with which they can be identified with automatic language detection tools or during post-editing. We also include \hyperref[sec:real-world-knowledge]{real-world knowledge} under minor errors since we do not generally expect MT evaluation metrics to have any notion of real-world knowledge and do not wish to punish them for this. Note that the \textsc{ACES}-Score ranges from -29.1 (all phenomena have a correlation of -1) to 29.1 (all phenomena have a correlation of +1).

\noindent
\begin{center}
    \begin{equation}
    \small
    \textsc{ACES} = sum \left\{
    \begin{aligned}
        \quad 5 * \tau_{\text{addition}}\\
       \quad  5 * \tau_{\text{omission}}\\
       \quad  5 * \tau_{\text{mistranslation}}\\
       \quad  1 * \tau_{\text{untranslated}}\\
       \quad  1 * \tau_{\text{do not translate}}\\
       \quad  5 * \tau_{\text{overtranslation}}\\
       \quad  5 * \tau_{\text{undertranslation}}\\
       \quad  1 * \tau_{\text{real-world knowledge}}\\
       \quad  1 * \tau_{\text{wrong language}}\\
       \quad  0.1 * \tau_{\text{punctuation}}\\
    \end{aligned}
    \right\} \vspace{0.5cm}
    \label{eq:aces-score}
    \end{equation}
\end{center}

We report an overview of the results for WMT 2022 in Table~\ref{tab:analysis_overview_2022} and the results for WMT 2023 in Table~\ref{tab:analysis_overview_2023}. Using the ACES-Score (the final column in each of the tables) we can see at a glance that the majority of the metrics submitted to the WMT 2022 shared task outperform the baseline metrics. The same is true of the WMT 2023 metrics -- except for CometKiwi, a successful submission from 2022 which was used as a baseline in 2023 -- the majority of the 2023 baseline metrics are outperformed by the metrics submitted by participants. Interestingly, in both years, many reference-free metrics performed on par with reference-based metrics. This is because our challenge sets are constructed to make the reference useless (ambiguous translation, discourse connectives, \textit{etc.,}), or misleading (hallucinations, lexical overlap, sentence-level meaning error). Note that we cannot directly compare the results from 2022 and 2023 -- for a small subset (2,659; approx. 7\%) of the \textsc{ACES} examples different results were returned in 2022 and 2023 for metrics where no changes had been made (e.g. baseline metrics such as BLEU or CometKiwi, etc.)\footnote{A subsequent investigation suggested that differences in the pre-processing steps by the shared task organisers in 2022 and 2023 may have led to the differences; in particular the handling of double quotes present in some of the \textsc{ACES} examples may be one of the main causes.}.

The best-performing metric in 2022 is a reference-free metric, namely \textsc{KG-BERTScore}, closely followed by the reference-based metric \textsc{metricx\_xl\_DA\_2019}. The best-performing metrics in 2023 are \textsc{COMETKiwi} (a reference-free baseline metric), and \textsc{KG-BERTScore}. Perhaps unsurprisingly, BLEU is one of the worst performing metrics, underperformed only by the random baseline, \textsc{Random-sysname}, in 2023. We caution that we developed \textsc{ACES} to investigate strengths and weaknesses of metrics on a phenomena level -- hence, we advise the reader not to draw any conclusions based solely on the \textsc{ACES}-Score.

Our observations regarding the metric performance were similar for both 2022 and 2023, and the following three points hold true for both years. Firstly, we observed that metric performance varies greatly and there is no clear winner in terms of performance across all of the categories. There is also a high degree of variation in terms of metric performance when each category is considered in isolation. Secondly, whilst each of the categories proves challenging for at least one metric, some categories are more challenging than others. For example, looking at the average scores in the last row of Table~\ref{tab:analysis_overview_2022}, and without taking outliers into account, we might conclude that \hyperref[sec:addition-omission]{addition}, \hyperref[sec:overtranslation_undertranslation]{undertranslation}, \hyperref[sec:real-world-knowledge]{real-world knowledge}, and \hyperref[sec:wrong_language]{wrong language} (all with average Kendall tau-like correlation of $<$ 0.3) present more of a challenge than the other categories. On the other hand, for \hyperref[sec:addition-omission]{omission} and \hyperref[sec:do-not-translate]{do not translate} (with an average Kendall tau-like correlation of $>$ 0.7 in 2022 and $>$ 0.6 in 2023) metric performance is generally rather high.

Thirdly, we also observe variation in terms of the performance of metrics belonging to the baseline, reference-based, and reference-free groups. For example, in both years, the baseline metrics generally appear to struggle more on the \hyperref[sec:overtranslation_undertranslation]{overtranslation and undertranslation} categories than the metrics belonging to the other groups. Reference-based metrics also appear to perform better overall on the \hyperref[sec:untranslated]{untranslated} category than the reference-free metrics. This makes sense as a comparison with the reference is likely to highlight tokens that ought to have been translated.
\begin{table}
\centering
\scriptsize
    \begin{minipage}[b]{0.5\textwidth}\centering
        \begin{tabular}{@{}lccc@{}}
\toprule
      -              & \hyperref[sec:discourse]{\textbf{disco.}} & \hyperref[sec:hallucination]{\textbf{halluci.}} & \textbf{other}         \\ \midrule
\textit{\textbf{Examples}}          & \textit{3698}   & \textit{10270} & \textit{10489} \\ \midrule
                    BLEU                    & -0.048                                         & -0.420                                             & -0.251                                     \\
f101spBLEU              & \phantom{-}0.105                                          & -0.206                                             & -0.153                                     \\
f200spBLEU              & \phantom{-}0.094                                          & -0.191                                             & -0.149                                     \\
chrF                    & \phantom{-}0.405                                          & -0.137                                             & \phantom{-}0.161                                      \\
BERTScore               & \phantom{-}0.567                                          & -0.058                                             & \phantom{-}0.362                                      \\
BLEURT-20               & \phantom{-}0.695                                          & \phantom{-}0.142                                              & \phantom{-}0.402                                      \\
COMET-20                & \phantom{-}0.641                                          & \phantom{-}0.016                                              & \phantom{-}0.399                                      \\
COMET-QE                & \phantom{-}0.666                                          & \phantom{-}0.303                                              & \phantom{-}0.208                                      \\
YiSi-1                  & \phantom{-}0.609                                          & \phantom{-}0.019                                              & \phantom{-}0.368                                      \\
\midrule
COMET-22                & \phantom{-}0.682                                          & \phantom{-}0.461                                              & \phantom{-}0.542                                      \\
metricx\_xl\_DA\_2019   & \phantom{-}0.701                                          & \phantom{-}0.493                                              & \phantom{-}0.458                                      \\
metricx\_xl\_MQM\_2020  & \phantom{-}0.573                                          & \phantom{-}0.677                                              & \phantom{-}0.394                                      \\
metricx\_xxl\_DA\_2019  & \phantom{-}0.768                                          & \phantom{-}0.541                                              & \phantom{-}0.463                                      \\
metricx\_xxl\_MQM\_2020 & \phantom{-}0.716                                          & \colorbox[HTML]{B2EAB1}{\textbf{\phantom{-}0.713}}                                              & \phantom{-}0.392                                      \\
MS-COMET-22             & \phantom{-}0.645                                          & \phantom{-}0.148                                              & \phantom{-}0.360                                      \\
UniTE                   & \phantom{-}0.746                                          & \phantom{-}0.322                                              & \phantom{-}0.424                                      \\
UniTE-ref               & \colorbox[HTML]{B2EAB1}{\textbf{\phantom{-}0.776}}                                          & \phantom{-}0.396                                              & \phantom{-}0.437                                      \\
\midrule
COMETKiwi               & \phantom{-}0.733                                          & \phantom{-}0.493                                              & \colorbox[HTML]{B2EAB1}{\textbf{\phantom{-}0.637}}                                      \\
Cross-QE                & \phantom{-}0.644                                          & \phantom{-}0.395                                              & \phantom{-}0.563                                      \\
HWTSC-Teacher-Sim       & \phantom{-}0.594                                          & \phantom{-}0.296                                              & \phantom{-}0.330                                      \\
HWTSC-TLM               & \phantom{-}0.756                                          & \phantom{-}0.306                                              & \phantom{-}0.151                                      \\
KG-BERTScore            & \phantom{-}0.593                                          & \phantom{-}0.387                                              & \phantom{-}0.472                                      \\
MS-COMET-QE-22          & \phantom{-}0.626                                          & \phantom{-}0.243                                              & \phantom{-}0.416                                      \\
UniTE-src               & \phantom{-}0.772                                          & \phantom{-}0.463                                              & \phantom{-}0.551                                      \\
\midrule
Average                 & \phantom{-}0.586                                          & \phantom{-}0.242                                              & \phantom{-}0.331                                     \\
\bottomrule
        \end{tabular}
        \caption{2022 Results. Average Kendall’s tau-like correlation results for the sub-level categories in mistranslation: \hyperref[sec:discourse]{\textbf{disco}urse-level}, \hyperref[sec:hallucination]{\textbf{halluci}nation}, and \textbf{other} errors.  The horizontal lines delimit baseline metrics (top), participating reference-based metrics (middle) and participating reference-free metrics (bottom). The best result for each category is denoted by bold text with a green highlight. Note that \textit{Average} is an average over averages.}
        \label{tab:analysis_mistranslation_2022}
    \end{minipage}
   \hfill
\begin{minipage}[b]{0.49\textwidth}\centering    
 \begin{tabular}{@{}lccc@{}}
\toprule
                    & \hyperref[sec:discourse]{\textbf{disco.}} & \hyperref[sec:hallucination]{\textbf{halluci.}} & \textbf{other}         \\
\midrule
\textbf{\textit{Examples}}           & \textbf{\textit{3698}}                                           & \textbf{\textit{10270}}                                              & \textbf{\textit{10489}}                                      \\
\midrule
BERTscore          & \phantom{-}0.563                                          & -0.062                                             & \phantom{-}0.361                                      \\
BLEU               & -0.042                                         & -0.418                                             & -0.250                                     \\
BLEURT-20          & \phantom{-}0.695                                          & \phantom{-}0.141                                              & \phantom{-}0.398                                      \\
chrF               & \phantom{-}0.406                                          & -0.138                                             & \phantom{-}0.160                                      \\
COMET-22           & \phantom{-}0.657                                          & \phantom{-}0.113                                              & \phantom{-}0.383                                      \\
CometKiwi          & \phantom{-}0.779                                          & \phantom{-}0.465                                              & \phantom{-}0.580                                      \\
f200spBLEU         & \phantom{-}0.095                                          & -0.190                                             & -0.150                                     \\
MS-COMET-QE-22     & \phantom{-}0.631                                          & \phantom{-}0.240                                              & \phantom{-}0.417                                      \\
Random-sysname     & -0.117                                         & -0.122                                             & -0.111                                     \\
YiSi-1             & \phantom{-}0.608                                          & \phantom{-}0.017                                              & \phantom{-}0.366                                      \\
\midrule
eBLEU              & \phantom{-}0.374                                          & -0.166                                             & \phantom{-}0.282                                      \\
embed\_llama       & -0.089                                         & -0.140                                             & \phantom{-}0.189                                      \\
MetricX-23         & \phantom{-}0.757                                          & \phantom{-}0.663                                              & \phantom{-}0.393                                      \\
MetricX-23-b       & \phantom{-}0.749                                          & \phantom{-}0.656                                              & \phantom{-}0.390                                      \\
MetricX-23-c       & \phantom{-}0.694                                          & \colorbox[HTML]{B2EAB1}{\textbf{\phantom{-}0.755}}                                              & \phantom{-}0.477                                      \\
partokengram\_F    & -0.062                                         & -0.101                                             & 0.027                                      \\
tokengram\_F       & \phantom{-}0.396                                          & -0.132                                             & \phantom{-}0.157                                      \\
XCOMET-Ensemble    & \colorbox[HTML]{B2EAB1}{\textbf{\phantom{-}0.791}}                                          & \phantom{-}0.566                                              & \phantom{-}0.626                                      \\
XCOMET-XL          & \phantom{-}0.706                                          & \phantom{-}0.482                                              & \phantom{-}0.521                                      \\
XCOMET-XXL         & \phantom{-}0.609                                          & \phantom{-}0.540                                              & \phantom{-}0.504                                      \\
XLsim              & \phantom{-}0.217                                          & -0.066                                             & \phantom{-}0.236                                      \\
\midrule
cometoid22-wmt21   & \phantom{-}0.782                                          & \phantom{-}0.286                                              & \phantom{-}0.400                                      \\
cometoid22-wmt22   & \phantom{-}0.748                                          & \phantom{-}0.290                                              & \phantom{-}0.423                                      \\
cometoid22-wmt23   & \phantom{-}0.758                                          & \phantom{-}0.223                                              & \phantom{-}0.478                                      \\
CometKiwi-XL       & \phantom{-}0.752                                          & \phantom{-}0.501                                              & \phantom{-}0.602                                      \\
CometKiwi-XXL      & \phantom{-}0.735                                          & \phantom{-}0.535                                              & \phantom{-}0.661                                      \\
GEMBA-MQM          & \phantom{-}0.076                                          & \phantom{-}0.291                                              & \phantom{-}0.127                                      \\
KG-BERTScore       & \phantom{-}0.685                                          & \phantom{-}0.466                                              & \phantom{-}0.580                                      \\
MetricX-23-QE      & \phantom{-}0.728                                          & \phantom{-}0.604                                              & \phantom{-}0.628                                      \\
MetricX-23-QE-b    & \phantom{-}0.694                                          & \phantom{-}0.617                                              & \phantom{-}0.666                                      \\
MetricX-23-QE-c    & \phantom{-}0.747                                          & \phantom{-}0.659                                              & \colorbox[HTML]{B2EAB1}{\textbf{\phantom{-}0.739}}                                      \\
XCOMET-QE-Ensemble & \phantom{-}0.702                                          & \phantom{-}0.558                                              & \phantom{-}0.651                                      \\
XLsimQE            & \phantom{-}0.053                                          & \phantom{-}0.050                                              & \phantom{-}0.134                                      \\
\midrule
Average            & \phantom{-}0.511                                          & \phantom{-}0.248                                              & \phantom{-}0.365     \\
\bottomrule
        \end{tabular}
        \caption{2023 Results. Average Kendall’s tau-like correlation results for the sub-level categories in mistranslation: \hyperref[sec:discourse]{\textbf{disco}urse-level}, \hyperref[sec:hallucination]{\textbf{halluci}nation}, and \textbf{other} errors.  The horizontal lines delimit baseline metrics (top), participating reference-based metrics (middle) and participating reference-free metrics (bottom). The best result for each category is denoted by bold text with a green highlight. Note that \textit{Average} is an average over averages.}
        \label{tab:analysis_mistranslation_2023}
\end{minipage}

\end{table}

Whilst there are many similarities in the performance trends observed in 2022 and 2023, there are also some differences. In 2023, we observe that the reference-free group exhibits overall stronger performance compared with the other groups, but in particular for the \textit{mistranslation}, \textit{overtranslation}, \textit{undertranslation}, and \textit{real-world knowledge} categories. We observe that (unlike in 2022) some of the 2023 metrics perform similarly to or worse than the baseline metrics. In particular, \textsc{embed\_llama} and \textsc{GEMBA-MQM} which are designed using Large Language Models (LLMs), struggle with this challenge set. This suggests that we need better design strategies in using the rich representations from LLMs for MT evaluation. 

\subsection{Mistranslation Results}

Next, we drill down to the fine-grained categories of the largest category: \textit{mistranslation}. We present metric performance on its sub-level categories (\textit{discourse}, \textit{hallucination}, and \textit{other}) in Table~\ref{tab:analysis_mistranslation_2022} (2022 results) and Table~\ref{tab:analysis_mistranslation_2023} (2023 results). The \textit{discourse} sub-category includes errors involving the mistranslation of discourse-level phenomena such as pronouns and discourse connectives. \textit{Hallucination} includes errors at the word level that could occur due to hallucination by an MT model, for example, the use of wrong units, dates, times, numbers or named entities, as well as hallucinations at the subword level that result in nonsensical words. The \textit{other} cub-category covers all other categories of mistranslation errors including overly literal translations of idioms and the introduction of ambiguities in the translation output. 

As for the results overview in Section~\ref{sec:aces_overview}, we find that performance on the different sub-categories is variable, with no clear winner among the metrics in either 2022 or 2023. The results from both years suggest that \hyperref[sec:hallucination]{hallucination} phenomena are generally more challenging than \hyperref[sec:discourse]{discourse-level} phenomena. Performance on the \hyperref[sec:hallucination]{hallucination} sub-category is poor overall, although it appears to be particularly challenging for the baseline metrics. We present additional, more fine-grained, performance analyses for individual phenomena in Section \ref{sec:analysis}.

\subsection{LLM Results}
\label{subsec:llm_results}
\begin{table}[]
\small
\centering
\begin{tabular}{@{}rcccccc@{}}
\toprule
 & \multicolumn{2}{c}{GEMBA-DA} & \multicolumn{2}{c}{LLAMA-2 (7B)} & \multicolumn{2}{c}{FLAN-T5-XL + Alpaca (3B)} \\
 \cmidrule{2-3}\cmidrule{4-5}\cmidrule{6-7}
 & REF & QE & REF & QE & REF & QE \\
addition & -0.235 & -0.794 & -0.607 & -0.587 & -0.834 & -0.922 \\
mistranslation & -0.031 & -0.322 & -0.58 & -0.552 & -0.656 & -0.832 \\
real-world knowledge & 0.366 & 0.157 & -0.58 & -0.6 & -0.280 & -0.739 \\
untranslated & -0.334 & -0.606 &  -0.650 & -0.626 & -0.529 & -0.631 \\
do not translate & -0.100 & -0.840 & -0.64  & -0.52 & -0.180 & -0.500 \\
undertranslation & 0.090 & -0.286 & -0.602 & -0.602 & 0.016 & -0.730 \\
overtranslation & 0.472 & -0.034 &  -0.564 & -0.524 & 0.026 & -0.744 \\
omission & -0.281 & -0.568 & -0.549 & -0.503 & -0.848 & -0.854 \\
punctuation & -0.306 & -0.355 & -0.646  & -0.650 & -0.875 & -0.924 \\
wrong language & 0.026 & -0.688 & -0.55 & -0.483 & -0.632 & -0.705 \\ \hline
\textsc{aces}-Score & -0.02 & -12.0 & -16.9 & -16.1 & -13.2 & -23.1
\end{tabular}
\caption{LLM results across three LLMs: GPT-4 through GEMBA-DA, \textsc{LLAMA-2}, and \textsc{FLAN-t5-XL} fine-tuned with Alpaca. REF: Reference based, QE: Quality Estimation/Reference-free. Using zero-shot prompting on LLMs for MT evaluation has results poorer than the surface overlap baselines in Table~\ref{tab:analysis_overview_2022}. This result worsens when the LLMs operate in a QE setting.}
\label{tab:llm_aces}
\end{table}
We report the results of the LLM experiments described in Section~\ref{subsec:llm_metrics} in Table~\ref{tab:llm_aces}. Overall, we find MT evaluation via LLMs a hard task in the zero-shot setup. This is also evident in the results in Section~\ref{sec:aces_overview} where we highlight the relatively low performance of \textsc{GEMBA-MQM} and \textsc{embed-LLaMA}. This is contrary to findings where LLMs show promising trends for evaluation at the system level \citep{fernandes-EtAl:2023:WMT,kocmi-federmann-2023-large}. 

We find that of the three LLMs, \textsc{GEMBA-DA} has better (though still poor) performance. These results worsen for the reference-less setting where most of the phenomena have a negative correlation. Despite the instructions for DA scores to be assigned using a continuous scale of 0--100, we find that the LLMs tend to produce a peaked distribution. For example, GEMBA-DA produces only seven different scores for the full set of examples. This results in a higher number of ties which get penalised in Equation~\ref{eq:kendal-tau}.
Even after instructing the LLMs to output scores within the range of 0--100, we observed instances where the LLMs produced scores beyond that range. 

These results suggest that while LLMs may perform well for MT evaluation under a specific setup like high-resource pairs or system-level evaluation, their zero-shot inference abilities for MT evaluation at segment-level are far from perfect. This can be attributed to a lack of multilingual training data \citep{kocmi-federmann:2023:WMT} as well as a limited numerical understanding of LLMs \citep{Dziri2023FaithAF}. We additionally express concerns over \textit{test-data leakage} as \textsc{ACES} is built on several other academic datasets (see Section~\ref{sec:aces_overview}) that may have been a part of the LLM training data \citep{Carlini2020ExtractingTD}. We also note that these models are quite slow at inference. It takes approximately six hours to make a pass over the entire dataset using \textsc{FLAN-T5-XL} on a 24GB GPU, while it takes five days with two 24GB GPUs for \textsc{LLaMA2} on 8bit precision.

\subsection{Span-based Results}

We first discuss the evaluation for \textsc{Span-ACES} and then report the results for the baseline methods discussed in Section~\ref{subsec:span_baselines}.

\subsubsection{Metrics for \textsc{Span-ACES}}
\label{sec: span_based_eval}

We consider two different types of evaluation for \textsc{Span-ACES} extraction and contrastive evaluation:

\noindent \textbf{Span Extraction}: We first measure how well the methods that produce spans perform the task of identifying erroneous span(s) in a translation. We evaluate the predicted spans for the incorrect translation against the gold annotation. We calculate sample F1 and average across the dataset where a span is considered to be a true positive if the span exactly matches its ground truth, denoted as \textit{Span-F1}. We also experimented with using partial matches between the gold error span and the predicted error span. However, using standardised tokenization based on words/sub-words/characters and then developing a threshold for partial match is not trivial and results in incorrect inflation of scores. 

\noindent \textbf{Contrastive Evaluation}: To evaluate these methods on \textsc{ACES} and compare their results, we obtain span predictions for the good translation as well. We use a length heuristic where we measure the number of times the metric produces fewer spans for the good translation compared with the incorrect translation (concordant) and greater than or equal to the incorrect translation (discordant) and calculate the correlation as described in Section~\ref{subsec:kendall_tau_formula}. If the severity of errors for the predicted spans is available, a use a weighted score based on the severity label.

\subsubsection{Results}
\label{subsec:span_based_results}

We now report the results of different models that produce error spans (and occasionally labels) from Section~\ref{sec: span_based_eval} on the \textsc{span-ACES} dataset in Table \ref{tab:span-based-results}. Overall, we find that these methods perform poorly on both the error span extraction and contrastive evaluation tasks.
\begin{table}[]
\small
    \resizebox{\linewidth}{!}{%

\begin{tabular}{@{}lrrrrrrrrrr@{}}
\toprule
                     & \multicolumn{3}{l}{\textsc{COMET-22}}                                                    & \multicolumn{3}{l}{\textsc{UniTE}}                                                 & \multicolumn{2}{l}{\textsc{XCOMET-XL}}                          & \multicolumn{2}{l}{\textsc{GEMBA-MQM}}                           \\ \midrule
                     & \multicolumn{1}{l}{src-ref} & \multicolumn{1}{l}{ref} & \multicolumn{1}{l}{src} & \multicolumn{1}{l}{src-ref} & \multicolumn{1}{l}{ref} & \multicolumn{1}{l}{src} & \multicolumn{1}{l}{length} & \multicolumn{1}{l}{weight} & \multicolumn{1}{l}{length} & \multicolumn{1}{l}{weight} \\ \hline
\multicolumn{11}{c}{\textbf{Span Extraction Evaluation}} \\

Span F1              & 26.9                        & 26.2                    & 4                       & 22.7                        & 22.7                    & 7.3                     & 10.6                       & 10.6                       & 8.67                       & 8.67                       \\ \hline

                     \multicolumn{11}{c}{\textbf{Contrastive Evaluation}} \\
addition             & 0.598                       & 0.477                   & -0.177                  & 0.522                       & 0.475                   & 0.317                   & -0.269                     & -0.191                     & -0.077                     & 0.103                      \\
mistranslation       & -0.313                      & -0.364                  & -0.482                  & -0.447                      & -0.431                  & -0.308                  & -0.222                     & -0.016                     & 0.005                      & 0.240                      \\
real-world knowledge & -0.470                      & -0.501                  & -0.417                  & -0.360                      & -0.377                  & -0.279                  & -0.202                     & 0.088                      & -0.330                     & 0.328                      \\
untranslated         & -0.641                      & -0.056                  & -0.689                  & -0.759                      & 0.260                   & -0.910                  & -0.239                     & -0.166                     & -0.152                     & 0.103                      \\
do not translate     & 0.500                       & 0.340                   & -0.380                  & 0.460                       & 0.520                   & 0.380                   & 0.060                      & 0.100                      & -0.080                     & 0.140                      \\
undertranslation     & -0.192                      & -0.206                  & -0.392                  & 0.110                       & 0.092                   & -0.220                  & -0.066                     & 0.250                      & 0.162                      & 0.368                      \\
overtranslation      & -0.144                      & -0.174                  & -0.362                  & 0.312                       & 0.284                   & -0.088                  & 0.008                      & 0.430                      & 0.236                      & 0.554                      \\
omission             & -0.770                      & -0.842                  & -0.838                  & -0.814                      & -0.784                  & -0.700                  & -0.381                     & -0.197                     & 0.165                      & 0.385                      \\
punctuation          & -0.385                      & -0.479                  & -0.609                  & -0.642                      & -0.574                  & -0.624                  & -0.593                     & -0.525                     & 0.039                      & 0.129                      \\
wrong language       & 0.406                       & 0.289                   & -0.212                  & 0.484                       & 0.387                   & 0.285                   & -0.225                     & -0.279                     & -0.132                     & -0.047                     \\ \hline
\textsc{ACES}-Score           & -4.3                        & -5.5                    & -13.0                   & -1.8                        & -1.1                    & -5.6                    & -5.3                       & 1.1                        & 1.8                        & 8.8                        \\ \bottomrule
\end{tabular}}
\caption{Results of span-based metrics on \textsc{Span-ACES} for the tasks of span extraction and then contrastive evaluation  on \textsc{ACES} using the predicted spans as outlined in Section~\ref{sec: span_based_eval}.  Under COMET-22 and UniTE, use of src and ref denotes if these components were used to obtain attention weights which were converted to spans. Span-F1 is only calculated for the incorrect translation. For the contrastive evaluation on \textsc{ACES}, all the above methods consider a candidate translation to be better than the other translation if the number of predicted spans in the former translation is less than the later, denoted by ``length''. For the ``weight'' version of \textsc{XCOMET-XL} and \textsc{GEMBA-MQM}, the labels denoting error severity of the predicted spans are converted to a weighted score. We note the derived metrics - \textsc{COMET-22} and \textsc{UniTE} have better results on the span extraction task than the metrics designed to predict the spans. This trend flips for the contrastive evaluation. Overall, all of the methods struggle on both tasks.}
\label{tab:span-based-results}
\end{table}

On the span extraction task, we find that the derived methods -- \textsc{COMET-22} and \textsc{UniTE} -- i.e. using attention maps over the source/reference sentences lead to higher Span-F1 scores than either \textsc{XCOMET} and \textsc{GEMBA-MQM} which were specifically designed to generate error spans. This adds some more evidence to the findings in \citet{rei-etal-2023-inside} that suggest metrics (\textsc{COMET-22} and \textsc{UniTE}) tend to use token-level information that can be associated with tangible translation errors. Within using attention maps over the source/reference sentences for \textsc{COMET-22} and \textsc{UniTE}, we find that the scores for the \textit{src} only version are the worst suggesting that these metrics use very limited information from the source (c.f. the similar observation made in Section~\ref{subsec:source-relevance}).

While using the length heuristic for the contrastive evaluation, \textsc{GEMBA-MQM} has better results followed by \textsc{UniTE}. As \textsc{GEMBA-MQM} and \textsc{XCOMET-XL} also provide labels to their predicted error spans, we also convert these labels into score based on the weights in \citet{guerreiro2023xcomet} (critical: 10, major: 5, minor: 1), then cap the error score per sentence at 25, and finally convert the score to a value between 0 and 1. We find that weighted label scores have a good improvement over the length heuristic suggesting that more sophisticated heuristics need to be developed in the future to obtain better meta-evaluation strategies. After using the label weighted score, we find that the performance for \textsc{XCOMET-XL} is still lower than the performance in Table~\ref{tab:analysis_overview_2023}, suggesting that the scores produced by the joint model may not necessarily rely on the error spans produced by that model.
In contrast, \textsc{GEMBA-MQM} improves on its performance in Tables~\ref{tab:analysis_overview_2023} and \ref{tab:span-based-results}. We attribute this to either a change in the underlying model powering GPT-4 between submissions to WMT and re-running for \textsc{span-ACES} or the use of a different weighting scheme. We also find it encouraging, that \textit{GEMBA-MQM} improves over \textit{GEMBA-DA}, providing us with some evidence that label-based evaluation can be helpful.

We speculate that these poor results may be attributed to (i) the unavailability of labelled MQM data during training (\textsc{COMET-22} and \textsc{UniTE}), (ii) the availability of labelled data for only a few language pairs (\textsc{XCOMET-XL}), (iii) the use of proprietary models, and thus no knowledge of underlying training data (\textsc{GEMBA-MQM}), (iv) the fact that these metrics are the earliest designs for span-based evaluation, and (v) the fact that our annotation schemes and evaluation regimes are also the first of their kind, potentially introducing new challenges for span-based evaluation metrics. 
We also caution the readers that our heuristics for contrastive evaluation only offer a starting point. Future work can include model confidence, different weighting schemes, POS tags \textit{etc.,} to compare the two translations.

\section{Analysis}
\label{sec:analysis}
Aside from high-level evaluations of which metrics perform best, we are mostly interested in weaknesses of metrics in general that we can identify using \textsc{ACES}. This section shows an analysis of some general questions that we aim to answer using \textsc{ACES}.

\subsection{How sensitive are the metrics to error types?}
\begin{sidewaystable*}[ht]
 \small
 \setlength{\tabcolsep}{3.75pt} 
 \centering 
 \resizebox{\textwidth}{!}{
 \begin{tabular}{@{}lccccccccccc@{}} 
 \\\toprule 
 & \hyperref[sec:addition-omission]{\textbf{addition}} & \hyperref[sec:addition-omission]{\textbf{omission}} & \hyperref[sec:source-disambig]{\textbf{mistranslation}} & \hyperref[sec:untranslated]{\textbf{untranslated}} & \hyperref[sec:do-not-translate]{\textbf{do not}} & \hyperref[sec:overtranslation_undertranslation]{\textbf{overtranslation}} & \hyperref[sec:overtranslation_undertranslation]{\textbf{undertranslation}} & \hyperref[sec:real-world-knowledge]{\textbf{real-world}} & \hyperref[sec:wrong_language]{\textbf{wrong}} & \hyperref[sec:punctuation]{\textbf{punctuation}} \\
 &  &  &  &  & \hyperref[sec:do-not-translate]{\textbf{translate}} &  &  & \hyperref[sec:real-world-knowledge]{\textbf{knowledge}} & \hyperref[sec:wrong_language]{\textbf{language}} & \\
\midrule
\textit{\textbf{Examples}}  & \textit{931} & \textit{951} & \textit{22530} & \textit{1187} & \textit{76} & \textit{962} & \textit{967} & \textit{2924} & \textit{1840}\\ 
 \midrule
BLEURT-20					&	\phantom{-}0.106	&	\phantom{-}0.355	&	\phantom{-}0.200	&	\phantom{-}1.743	&	\phantom{-}0.398	&	\phantom{-}0.142	&	-0.002	&	\phantom{-}0.055	&	\colorbox[HTML]{B2EAB1}{\textbf{\phantom{-}0.826}}	&	\phantom{-}0.318\\ 
COMET-20					&	\phantom{-}0.073	&	\phantom{-}0.410	&	\phantom{-}0.262	&	\phantom{-}1.486	&	\phantom{-}0.312	&	\phantom{-}0.150	&	\phantom{-}0.061	&	\phantom{-}0.051	&	-0.322	&	\phantom{-}0.229\\ 
YISI-1					&	\phantom{-}0.118	&	\phantom{-}0.293	&	\phantom{-}0.075	&	\colorbox[HTML]{B2EAB1}{\textbf{\phantom{-}2.575}}	&	\phantom{-}0.294	&	-0.036	&	-0.049	&	-0.044	&	\phantom{-}0.190	&	\colorbox[HTML]{B2EAB1}{\textbf{\phantom{-}0.376}}	\\ 
\midrule 
metricx\_xl\_DA\_2019					&	\phantom{-}0.100	&	\phantom{-}0.447	&	\phantom{-}0.496	&	\phantom{-}1.772	&	\phantom{-}0.559	&	\phantom{-}0.689	&	\phantom{-}0.366	&	\phantom{-}0.498	&	\phantom{-}0.752	&	\phantom{-}0.302	\\ 
metricx\_xl\_MQM\_2020					&	-0.056	&	\phantom{-}0.361	&	\colorbox[HTML]{B2EAB1}{\textbf{\phantom{-}0.651}}	&	\phantom{-}0.422	&	\colorbox[HTML]{B2EAB1}{\textbf{\phantom{-}0.697}}	&	\colorbox[HTML]{B2EAB1}{\textbf{\phantom{-}1.000}}	&	\colorbox[HTML]{B2EAB1}{\textbf{\phantom{-}0.654}}	&	\colorbox[HTML]{B2EAB1}{\textbf{\phantom{-}0.740}}	&	-0.560	&	\phantom{-}0.331 \\ 
metricx\_xxl\_MQM\_2020					&	-0.008	&	\phantom{-}0.294	&	\phantom{-}0.550	&	\phantom{-}0.649	&	\phantom{-}0.688	&	\phantom{-}0.826	&	\phantom{-}0.485	&	\phantom{-}0.629	&	-0.768	&	\phantom{-}0.225	 \\ 
\midrule 
COMETKiwi					&	\colorbox[HTML]{B2EAB1}{\textbf{\phantom{-}0.126}}	&	\phantom{-}0.441	&	\phantom{-}0.594	&	\phantom{-}0.699	&	\phantom{-}0.272	&	\phantom{-}0.572	&	\phantom{-}0.358	&	\phantom{-}0.337	&	-0.559	&	\phantom{-}0.247	 \\ 
Cross-QE					&	\phantom{-}0.104	&	\phantom{-}0.422	&	\phantom{-}0.599	&	-0.285	&	\phantom{-}0.055	&	\phantom{-}0.703	&	\phantom{-}0.456	&	\phantom{-}0.225	&	-0.510	&	\phantom{-}0.109	\\ 
UniTE-src					&	\phantom{-}0.096	&	\colorbox[HTML]{B2EAB1}{\textbf{\phantom{-}0.524}}	&	\phantom{-}0.484	&	-0.782	&	\phantom{-}0.318	&	\phantom{-}0.394	&	\phantom{-}0.254	&	\phantom{-}0.191	&	-0.552	&	\phantom{-}0.196	\\ [0.5ex]

\hline \hline & \\[-1.5ex]

COMETKiwi &	\phantom{-}0.196	&	\phantom{-}0.618	&	\phantom{-}0.721	&	-0.180	&	\phantom{-}0.285	&	\phantom{-}0.719	&	\phantom{-}0.439	&	\phantom{-}0.316	&	-0.767	&	\phantom{-}0.218	 \\ 
MS-COMET-QE-22					&	-0.040	&	\phantom{-}0.391	&	\phantom{-}0.258	&	\colorbox[HTML]{B2EAB1}{\textbf{\phantom{-}2.730}}	&	\phantom{-}0.126	&	\phantom{-}0.257	&	\phantom{-}0.185	&	\phantom{-}0.095	&	-0.654	&	\phantom{-}0.226 \\ 
BLEURT-20					&	\phantom{-}0.094	&	\phantom{-}0.314	&	\phantom{-}0.177	&	\phantom{-}1.545	&	\phantom{-}0.353	&	\phantom{-}0.126	&	-0.002	&	\phantom{-}0.048	&	\colorbox[HTML]{B2EAB1}{\textbf{\phantom{-}0.732}}	&	\colorbox[HTML]{B2EAB1}{\textbf{\phantom{-}0.281}}	\\ 
\midrule 
MetricX-23-c					&	\phantom{-}0.021	&	\phantom{-}0.413	&	\phantom{-}0.645	&	\phantom{-}0.334	&	\phantom{-}0.399	&	\phantom{-}0.593	&	\phantom{-}0.330	&	\phantom{-}0.766	&	-0.728	&	\phantom{-}0.131 \\ 
MetricX-23					&	\phantom{-}0.001	&	\phantom{-}0.184	&	\phantom{-}0.407	&	-0.022	&	\phantom{-}0.363	&	\phantom{-}0.675	&	\phantom{-}0.367	&	\phantom{-}0.491	&	-0.618	&	\phantom{-}0.151	 \\ 
XCOMET-Ensemble					&	\phantom{-}0.070	&	\phantom{-}0.342	&	\phantom{-}0.434	&	\phantom{-}0.208	&	\phantom{-}0.249	&	\phantom{-}0.462	&	\phantom{-}0.308	&	\phantom{-}0.358	&	-0.713	&	\phantom{-}0.151 \\ 
\midrule 
GEMBA-MQM					&	\colorbox[HTML]{B2EAB1}{\textbf{\phantom{-}0.584}}	&	\colorbox[HTML]{B2EAB1}{\textbf{\phantom{-}1.132}}	&	\colorbox[HTML]{B2EAB1}{\textbf{\phantom{-}1.566}}	&	\phantom{-}2.380	&	\colorbox[HTML]{B2EAB1}{\textbf{\phantom{-}0.719}}	&	\colorbox[HTML]{B2EAB1}{\textbf{\phantom{-}1.646}}	&	\colorbox[HTML]{B2EAB1}{\textbf{\phantom{-}0.976}}	&	\colorbox[HTML]{B2EAB1}{\textbf{\phantom{-}1.814}}	&	\phantom{-}0.328	&	\phantom{-}0.226	 \\ 
MetricX-23-QE					&	\phantom{-}0.001	&	\phantom{-}0.410	&	\phantom{-}0.903	&	\phantom{-}0.100	&	\phantom{-}0.309	&	\phantom{-}0.995	&	\phantom{-}0.660	&	\phantom{-}0.955	&	-1.163	&	\phantom{-}0.135	 \\ 
CometKiwi-XXL					&	\phantom{-}0.118	&	\phantom{-}0.519	&	\phantom{-}0.713	&	\phantom{-}2.038	&	\phantom{-}0.197	&	\phantom{-}0.550	&	\phantom{-}0.306	&	\phantom{-}0.611	&	-0.733	&	\phantom{-}0.187	 \\ 

\end{tabular}}
\caption{Metric sensitivity scores (scaled by WMT scores, then avg(good - bad)) for the nine top level categories in the \textsc{ACES} ontology, plus the additional fluency category: punctuation. The double horizontal line delimits the metrics submitted to WMT 2022 (top three groups) and the metrics submitted to WMT 2023 (bottom three groups). In each of these groups, the horizontal lines delimit baseline metrics (top), participating reference-based metrics (middle) and participating reference-free metrics (bottom), where we picked the top three metrics from each. The highest result for each category is denoted by bold text with a green highlight.}
\label{tab:analysis_sensitivities_top_metrics}
\end{sidewaystable*}
\afterpage{\clearpage}

One important quality of a reliable metric is its ability to assign sufficiently different scores to a good vs. an incorrect translation. To evaluate and compare the difference between the scores that the metrics assign to the good and incorrect translations, we normalise the metric scores to a common scale with an open-ended range, using the statistics from the metric scores submitted to the 2022 and 2023 editions of the WMT metrics task \citep{freitag-etal-2022-results,freitag-etal-2023-results}. We do that by scaling the metric scores based on the mean and IQR (Interquartile range) of the scores of that metric submitted to the WMT22/23 metric shared task (see Equation~\ref{eq:normalised_score}). 

\vspace{-10pt}
\begin{equation}
    \begin{aligned}
        score^*=\frac{score-Avg(score_{wmt})}{IQR_{wmt}}
    \end{aligned}
    \label{eq:normalised_score}
\end{equation}

We calculate the sensitivity score of the metric (see Equation~\ref{eq:sensitivity}) on a subset of the \textsc{ACES} samples, which is the average difference between the normalised scores assigned to good translations and to incorrect translations. Here $score_+^*$ is the normalised score assigned to the \textit{good translation} and $score_-^*$ is the normalised score assigned to the \textit{incorrect translation}. The value range of the sensitivity scores is open. \footnote{Evaluation scripts are available here: \url{https://github.com/EdinburghNLP/ACES}}

\vspace{-10pt}
\begin{equation}
    \begin{aligned}
        sensitivity=Avg(score_+^*-score_-^*)
    \end{aligned}
    \label{eq:sensitivity}
\end{equation}

Similarly to the Kendall’s tau-like correlation scores, we then report the average score overall examples belonging to each of the nine top-level accuracy categories in \textsc{ACES}, plus the fluency category \textit{punctuation}, calculated for the top three metrics from the baseline, reference-based and reference-free metrics each, submitted to WMT 2022, and WMT 2023 (see Table~\ref{tab:analysis_sensitivities_top_metrics}). The phenomena-level sensitivity scores for all the metrics submitted to WMT 2022 and WMT 2023 can be found in Appendix~\ref{app:phenomena-level-sensitivities}.

The average sensitivity scores of the metrics support the results reached by the analysis of the average Kendall’s tau-like correlation scores in most cases. One of the most significant exceptions to that is that \textsc{GEMBA-MQM} has significantly higher sensitivity scores across a majority of the high-level phenomena when evaluated according to the average sensitivity scores, unlike the Kendall’s tau-like correlation results. 

Looking at the average sensitivity scores of the metrics in the last row of Tables~\ref{tab:analysis_overview_sensitivities_2022} and ~\ref{tab:analysis_overview_sensitivities_2023} in Appendix~\ref{app:phenomena-level-sensitivities}, we can see that the metrics are more sensitive to the \hyperref[sec:untranslated]{untranslated} category than all the other categories by a margin, where the \hyperref[sec:untranslated]{untranslated} category is not one of the easier categories according to the average Kendall’s tau-like correlation scores.

Regarding the subcategories of mistranslation, discourse, previously considered the least challenging category based on Kendall’s tau-like correlation, emerges as the most difficult for the metrics according to sensitivity scores. It can be seen that across multiple 2022 and 2023 metrics, the average sensitivity scores of the metrics on the hallucination subcategory are higher compared to the average sensitivity scores on discourse, while the average Kendall’s tau-like correlation scores favour the discourse subcategory over hallucination. 



\subsection{How sensitive are metrics to the source?}
\label{subsec:source-relevance}

We designed our challenge sets for the type of \hyperref[sec:source-disambig]{ambiguous translation} in a way that the correct translation candidate given an ambiguous reference can only be identified through the source sentence. See the third example in Table~\ref{tab:ACES_top_level_examples}, where the reference is in non-gendered language, thus requiring the information in the source sentence about the female baker to disambiguate the sentence. We present a targeted evaluation intended to provide some insights into how important the source is for different metrics. 
For brevity, we include top three performing metrics in each category in 2022 and 2023, and a couple of baseline metrics.
Table~\ref{tab:src_disambig} shows the detailed results of each metric on the considered phenomena.

The most important finding is that the reference-free metrics generally perform much better on these challenge sets than the reference-based metrics. This indicates that reference-based metrics rely too much on the reference. Interestingly, most of the metrics that seem to ignore the source do not randomly guess the correct translation (which is a valid alternative choice when the correct meaning is not identified via the source) but rather they strongly prefer one phenomenon over the other. For example, several metrics show a gender bias either towards female \hyperref[sec:source-disambig]{occupation names} (female correlations are high, male low) or male occupation names (vice versa). Likewise, most metrics prefer translations with frequent senses for the \hyperref[sec:source-disambig]{word-sense disambiguation challenge sets}, although the difference between frequent and infrequent is not as pronounced as for gender.

Only metrics that look at the source and exhibit fewer such preferences can perform well on average on this collection of challenge sets. \textsc{XCOMET-Ensemble} performs best out of the reference-based metrics and \textsc{XCOMET-QE-Ensemble} performs best of all reference-free metrics. It is noteworthy that there is still a considerable gap between these two models across most of the  error categories, suggesting that reference-based models should pay more attention to the source when a reference is ambiguous to reach the performance of reference-free metrics.

This finding is also supported by our \hyperref[sec:real-world-knowledge]{real-world knowledge commonsense challenge set}. If we compare the scores on the examples where the subordinate clauses are missing from both the source and the reference to the ones where they are only missing from the reference, we can directly see the effect of disambiguation through the source. The corresponding correlation gains are shown in Table~\ref{tab:corr_gain} in the Appendix. All reference-based model correlation scores improve less than most reference-free correlations when access to the subordinate clause is given through the source. This highlights again that reference-based metrics do not give enough weight to the source sentence.
\begin{table*}[h]
    \centering
    \small
    \resizebox{\linewidth}{!}{%

    \begin{tabular}{lcccccccccc}
    \toprule
    & \multicolumn{2}{c}{\hyperref[sec:discourse]{\textbf{since}}} & \multicolumn{2}{c}{\hyperref[sec:source-disambig]{\textbf{female}}}  & \multicolumn{2}{c}{\hyperref[sec:source-disambig]{\textbf{male}}} & \multicolumn{2}{c}{\hyperref[sec:source-disambig]{\textbf{wsd}}} &\\
    \cmidrule(lr){2-3} \cmidrule(lr){4-5} \cmidrule(lr){6-7} \cmidrule(lr){8-9} \\
    & \textbf{causal} & \textbf{temp.} & \textbf{anti.} & \textbf{pro.} & \textbf{anti.} & \textbf{pro.} & \textbf{freq.} & \textbf{infreq.} & \textbf{AVG}\\
    \cmidrule(lr){2-2} \cmidrule(lr){3-3} \cmidrule(lr){4-4} \cmidrule(lr){5-5} \cmidrule(lr){6-6}  \cmidrule(lr){7-7} \cmidrule(lr){8-8} \cmidrule(lr){9-9} \cmidrule(lr){10-10} \\
    \textit{\textbf{Examples}} & \textit{106} & \textit{106} & \textit{1000} & \textit{806} & \textit{806} & \textit{1000} & \textit{471} & \textit{471} & \textit{4766}\\
    \midrule
   BERTScore & -0.434 & \phantom{-}0.434 & -0.614 & -0.216 & \phantom{-}0.208 & \phantom{-}0.618 & \phantom{-}0.214 & -0.223 & -0.001\\
   COMET-22 & -0.415 & \phantom{-}0.792 & \colorbox[HTML]{B2EAB1}{\textbf{\phantom{-}0.940}} & \colorbox[HTML]{B2EAB1}{\textbf{\phantom{-}1.000}} & -0.628 & \phantom{-}0.374 & \colorbox[HTML]{B2EAB1}{\textbf{\phantom{-}0.558}} & \phantom{-}0.040 & \phantom{-}0.333\\
   MS-COMET-22 & -0.604 & \phantom{-}0.623 & \phantom{-}0.296 & \phantom{-}0.640 & -0.342 & \phantom{-}0.046 & \phantom{-}0.316 & -0.155 & \phantom{-}0.102\\
   UniTE & \colorbox[HTML]{B2EAB1}{\textbf{\phantom{-}0.038}} & -0.075 & -0.890 & -0.213 & \phantom{-}0.377 & \phantom{-}0.934 & \phantom{-}0.270 & -0.223 & \phantom{-}0.027\\
   MetricX-23 & -1.000 & \colorbox[HTML]{B2EAB1}{\textbf{1.000}} & -0.864 & -0.062 & \phantom{-}0.062 & \phantom{-}0.870 & 0.227 & -0.222 & \phantom{-}0.001\\
MetricX-23-c & -0.849 & \phantom{-}0.849 & -0.998 & -0.581 & \colorbox[HTML]{B2EAB1}{\textbf{\phantom{-}0.576}} & \colorbox[HTML]{B2EAB1}{\textbf{\phantom{-}0.996}} &  \phantom{-}0.150 & -0.133 & \phantom{-}0.172  \\
XCOMET-Ensemble & -0.585 & \phantom{-}0.981 & \phantom{-}0.852 & \phantom{-}0.948 & \phantom{-}0.273 & \phantom{-}0.922 & \phantom{-}0.554 & \colorbox[HTML]{B2EAB1}{\textbf{\phantom{-}0.231}} &  \colorbox[HTML]{B2EAB1}{\textbf{\phantom{-}0.522}}\\
   \midrule
   Cross-QE & \colorbox[HTML]{B2EAB1}{\textbf{\phantom{-}0.208}} & \phantom{-}0.830 & \phantom{-}0.976 & \phantom{-}0.995 & -0.337 & \phantom{-}0.364 & \colorbox[HTML]{B2EAB1}{\textbf{\phantom{-}0.762}} & \phantom{-}0.355 & \phantom{-}0.519 \\
    MS-COMET-QE-22 & -0.283 & \phantom{-}0.792 & -0.194 & \phantom{-}0.320 & \phantom{-}0.246 & \phantom{-}0.694 & \phantom{-}0.465 & \phantom{-}0.002 & \phantom{-}0.255 \\
    UniTE-src & -0.321 & \phantom{-}0.906 & \phantom{-}0.976 & \phantom{-}0.980 & \phantom{-}0.171 & \phantom{-}0.736 & \phantom{-}0.622 & \phantom{-}0.346 & \phantom{-}0.552 \\
    CometKiwi & \phantom{-}0.075 & \colorbox[HTML]{B2EAB1}{\textbf{1.000}} & \colorbox[HTML]{B2EAB1}{\textbf{\phantom{-}0.990}} & \colorbox[HTML]{B2EAB1}{\textbf{\phantom{-}0.998}} & -0.171 & 0.440 & 0.740 & 0.384 & 0.557\\
    KG-BERTScore & \phantom{-}0.075 & \colorbox[HTML]{B2EAB1}{\textbf{1.000}} & \colorbox[HTML]{B2EAB1}{\textbf{\phantom{-}0.990}} & \colorbox[HTML]{B2EAB1}{\textbf{\phantom{-}0.998}} & -0.171 & \phantom{-}0.440 & 0.702 & 0.460 & 0.315\\
    MetricX-23-QE-b & -0.566 & \phantom{-}0.868 & \phantom{-}0.968 & \phantom{-}0.995 & \colorbox[HTML]{B2EAB1}{\textbf{\phantom{-}0.722}} & \colorbox[HTML]{B2EAB1}{\textbf{\phantom{-}0.968}} & \phantom{-}0.643 & \colorbox[HTML]{B2EAB1}{\textbf{\phantom{-}0.490}} & \phantom{-}0.643\\
    XCOMET-QE-Ensemble & -0.208 & \phantom{-}0.925 & \phantom{-}0.930 & \phantom{-}0.975 & \phantom{-}0.546 & \phantom{-}0.912 & \phantom{-}0.740 & \phantom{-}0.477 & \colorbox[HTML]{B2EAB1}{\textbf{\phantom{-}0.662}}\\
    
   \bottomrule
    \end{tabular}}
    \caption{Results on the challenge sets where the good translation can only be identified through the source sentence. Upper block: reference-based metrics, lower block: reference-free metrics. The best results for each phenomenon and each group of models are marked in bold and green and the average overall can be seen in the last column.}
    \label{tab:src_disambig}
\end{table*}

\subsection{How much do metrics rely on surface overlap with the reference?}
\label{subsec:surface-relevance}
Another question we are interested in is whether neural reference-based metrics still rely on surface-level overlap with the reference. For this analysis, we use the dataset we created for \hyperref[sec:hallucination]{hallucinated named entities and numbers}. We add an example about the three levels in Table~\ref{tab:levels}
Note that as the levels increase, the surface level similarity between the good translation and the reference decreases while the surface level overlap between the incorrect translation and the reference increases. 
\begin{small}
\vspace{1.5cm}
\label{tab:levels}
\begin{tabularx}{0.95\columnwidth}{lX}
    
     SRC (es): & Sin embargo, Michael Jackson, Prince y \textbf{Madonna} fueron influencias para el álbum. \\
     REF (en): & Michael Jackson, Prince and \textbf{Madonna} were, however, influences on the album. \\\\\hline\\
    Level-1 \cmark: & However, Michael Jackson, Prince, and \textbf{Madonna} were influences on the album. \\
    Level-1 \xmark: & However, Michael Jackson, Prince, and \textbf{Garza} were influences on the album. \\\\\hline\\
    Level-2 \cmark: & However, Michael Jackson, Prince, and \textbf{Madonna} were influences on the album. \\
    Level-2 \xmark: &  Michael Jackson, Prince and \textbf{Garza} were, however, influences on the album.\\\\\hline\\
    Level-3 \cmark: & The record was influenced by \textbf{Madonna}, Prince, and Michael Jackson  though. \\
    Level-3 \xmark: & Michael Jackson, Prince and \textbf{Garza} were, however, influences on the album.\vspace{0.35cm}
\end{tabularx}
\end{small}

\begin{figure}
    \centering
    \includegraphics[width=0.48\textwidth]{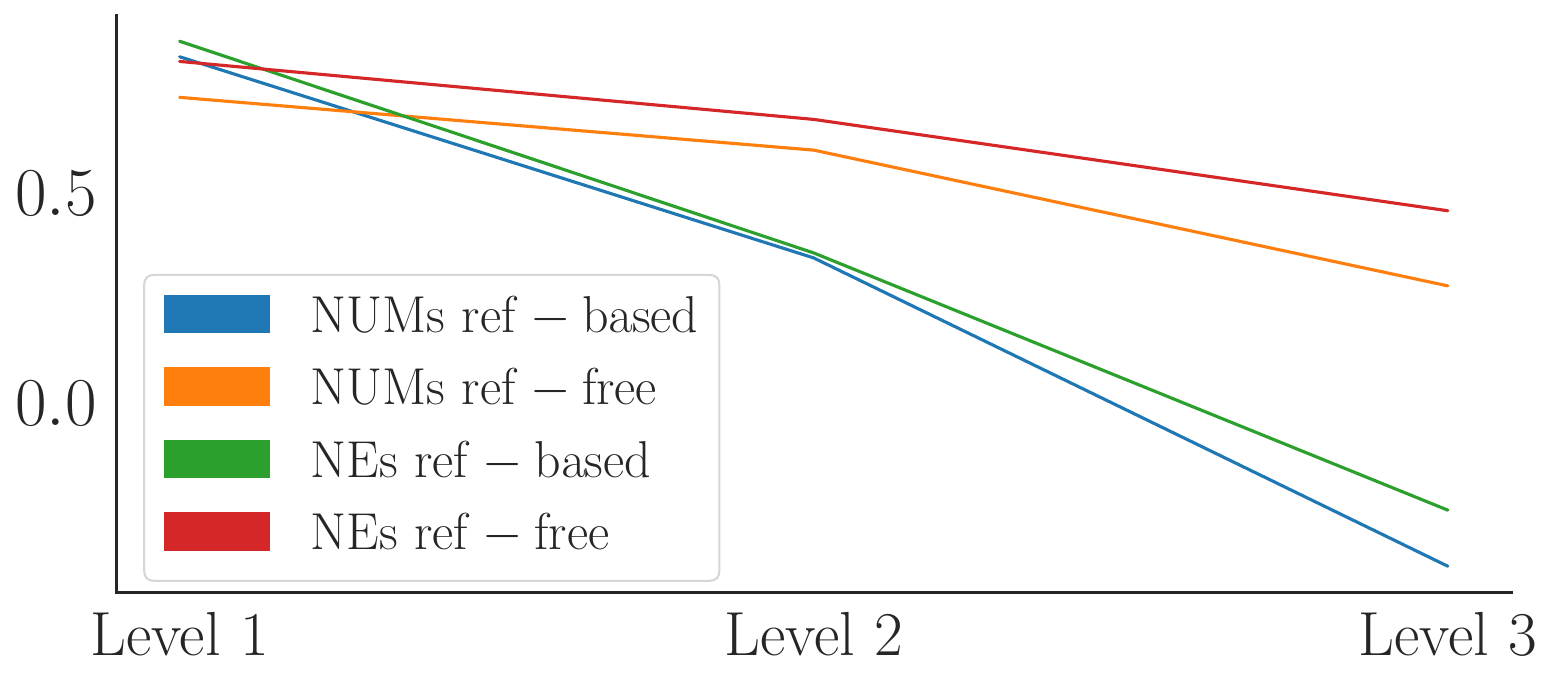}
    \caption{Decrease in correlation for reference-based and reference-free metrics on the \hyperref[sec:hallucination]{named entity and number hallucination challenge sets}.}
    \label{fig:corr_decrease}
\end{figure}

We take the average correlation for all reference-based metrics, (excluding lexical overlap metrics like \textsc{BLEU}) and the average correlation of all reference-free metrics that cover all languages across both the years and plot the decrease in correlation with increasing surface-level similarity of the incorrect translation to the reference. The result can be seen in Figure~\ref{fig:corr_decrease}.

We can see that on average reference-based metrics have a much steeper decrease in correlation than the reference-free metrics as the two translation candidates become more and more lexically diverse and the surface overlap between the incorrect translation and the reference increases. This indicates a possible weakness of reference-based metrics: If one translation is lexically similar to the reference but contains a grave error while others are correct but share less surface-level overlap with the reference, the incorrect translation may still be preferred.

We also show that this is the case for the challenge set where we use an adversarial paraphrase from PAWS-X that shares a high degree of \hyperref[subsec:lexical-overlap]{lexical overlap} with the reference but does not have the same meaning as an incorrect translation. On average, the reference-based metrics only reach a correlation of 0.05 ± 0.17 on this challenge set, whereas the reference-free metrics reach a correlation of 0.24 ± 0.17. This shows that reference-based metrics are less robust when the incorrect translation has high lexical overlap with the reference.


\begin{table}[]
    \centering
    \small
    \begin{tabular}{ccc}
    \toprule
         & reference-based & reference-free \\
        \midrule
        \hyperref[sec:hallucination]{hallucination} & -0.21 ± 0.15 & +0.01 ± 0.05 \\
        \hyperref[sec:overly_literal]{overly-literal} & -0.32 ± 0.16
        & +0.07± 0.09\\
        \hyperref[sec:untranslated]{untranslated} & -0.43 ± 0.15 & -0.00 ± 0.07\\
    \bottomrule
    \end{tabular}
    \caption{Average correlation difference and standard deviation between the challenge sets with reference-copied good translations and the challenge sets with the synonymous good translations.}
    \label{tab:corr-difference-syn-ref}
\end{table}

\subsection{Do multilingual embeddings help design better metrics?}
\label{subsec:mutlilingual embeddings}
As the community moves towards building metrics that use multilingual encoders, we investigate if some (un)desirable properties of multilingual embeddings or other base models are propagated in these metrics.

\begin{figure}
    \centering
    \includegraphics[width=0.46\textwidth]{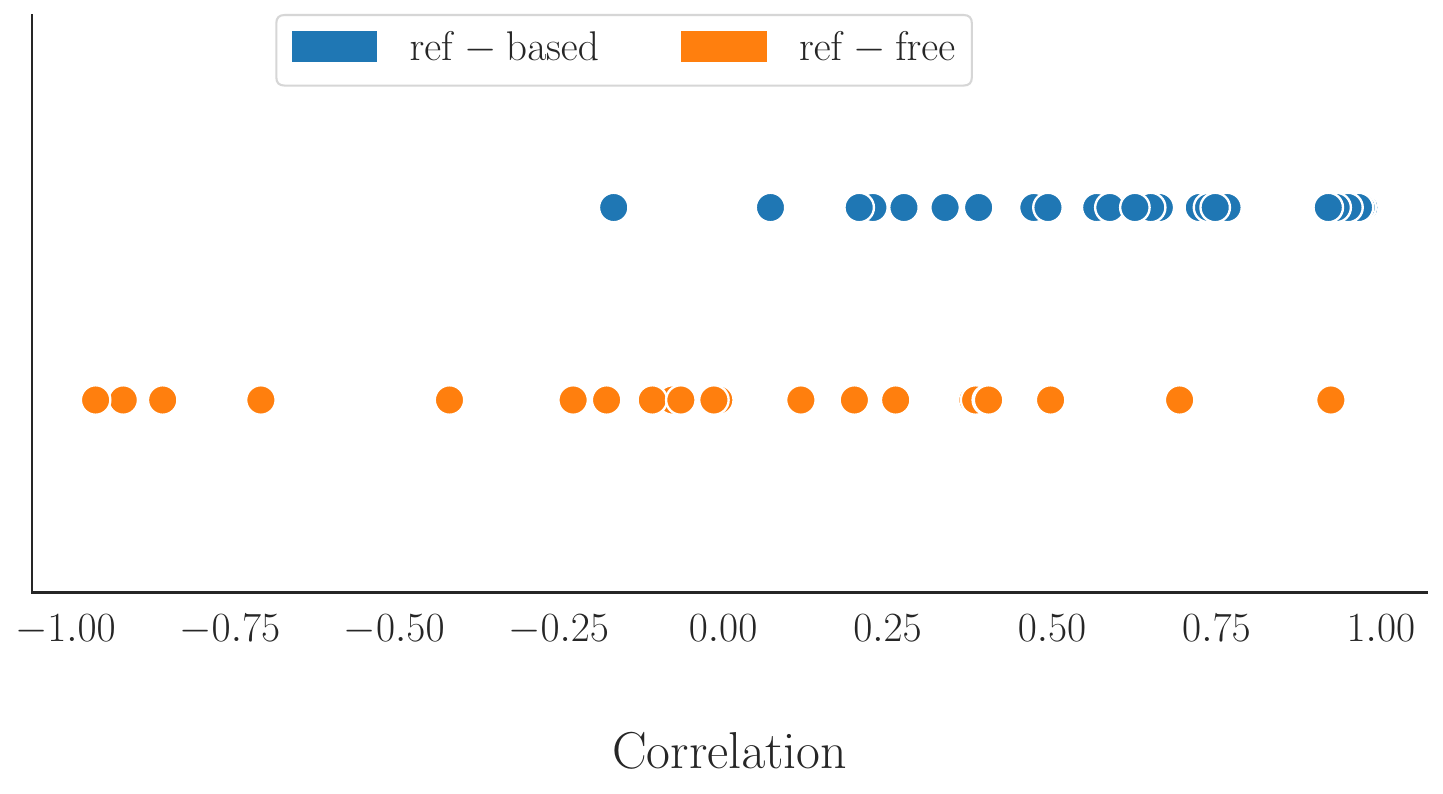}
    \caption{Correlation of reference-based metrics (blue) and reference-free metrics (orange) on the  \hyperref[sec:untranslated]{sentence-level untranslated test challenge set}}.
   \label{fig:corr_copy_src}
\end{figure}

Multilingual models often learn cross-lingual representations by abstracting away from language-specific information \citep{wu-dredze-2019-beto}. We are interested in whether the representations are still language-dependent in neural MT evaluation metrics which are trained on such models. For this analysis, we look at the \hyperref[sec:untranslated]{sentence-level untranslated text} challenge set (see Figure \ref{fig:corr_copy_src}) and \hyperref[sec:wrong_language]{wrong language phenomena} (see Table~\ref{tab:analysis_overview_2022}).

Figure \ref{fig:corr_copy_src} shows the correlations for all reference-based and reference-free metrics. Unsurprisingly, some 
reference-free metrics struggle considerably on this challenge set and almost always prefer the copied source to the real translation. The representations of the source and the incorrect translation are identical, leading to a higher surface and embedding similarity, and thus a higher score. We do, however, find some exceptions to this trend - \textsc{COMET-Kiwi} and \textsc{MS-COMET-QE-22} both have a high correlation on  \hyperref[sec:untranslated]{sentence-level untranslated text}. This suggests that these metrics could have learnt language-dependent representations. 

Most reference-based metrics have good to almost perfect correlation and can identify the copied source quite easily. As reference-based metrics tend to ignore the source (see Section~\ref{subsec:surface-relevance}), the scores are based on the similarity between the reference and the MT output. In this challenge set, the similarity between the good translation and the reference is likely to be higher than the incorrect translation and the reference. The former MT output is in the same language as the reference and will have more surface-level overlap. We believe the reference here acts as grounding.

However, this grounding property of the reference is only robust when the source and reference languages are dissimilar, as is the case with language pairs in the \hyperref[sec:untranslated]{sentence-level untranslated text} challenge set. We find that reference-based metrics struggle on \hyperref[sec:wrong_language]{wrong language phenomena}  (see Tables \ref{tab:analysis_overview_2022}, \ref{tab:analysis_mistranslation_2023}) where the setup is similar, but now the incorrect translation and the reference are from similar languages (e.g. one is in Hindi and the other is in Marathi). Naturally, there will be surface-level overlap between the reference and both the good translation and the incorrect translation. For example, both Marathi and Hindi use named entities with identical surface form, and so these will appear in the reference and also in both the good translation and the incorrect translation. Thus, the semantic content drives the similarity scores between the MT outputs and the references. The human translation in the similar language (labelled as the incorrect translation) may have a closer representation to the human reference because in the MT output (labelled as the good translation) some semantic information may be lost. We leave further investigation of this for future work.

\subsection{How does metric training data size affect MT evaluation?}
\label{subsec:data_effects}
\begin{figure}
    \centering
    \includegraphics[width=0.48\textwidth]{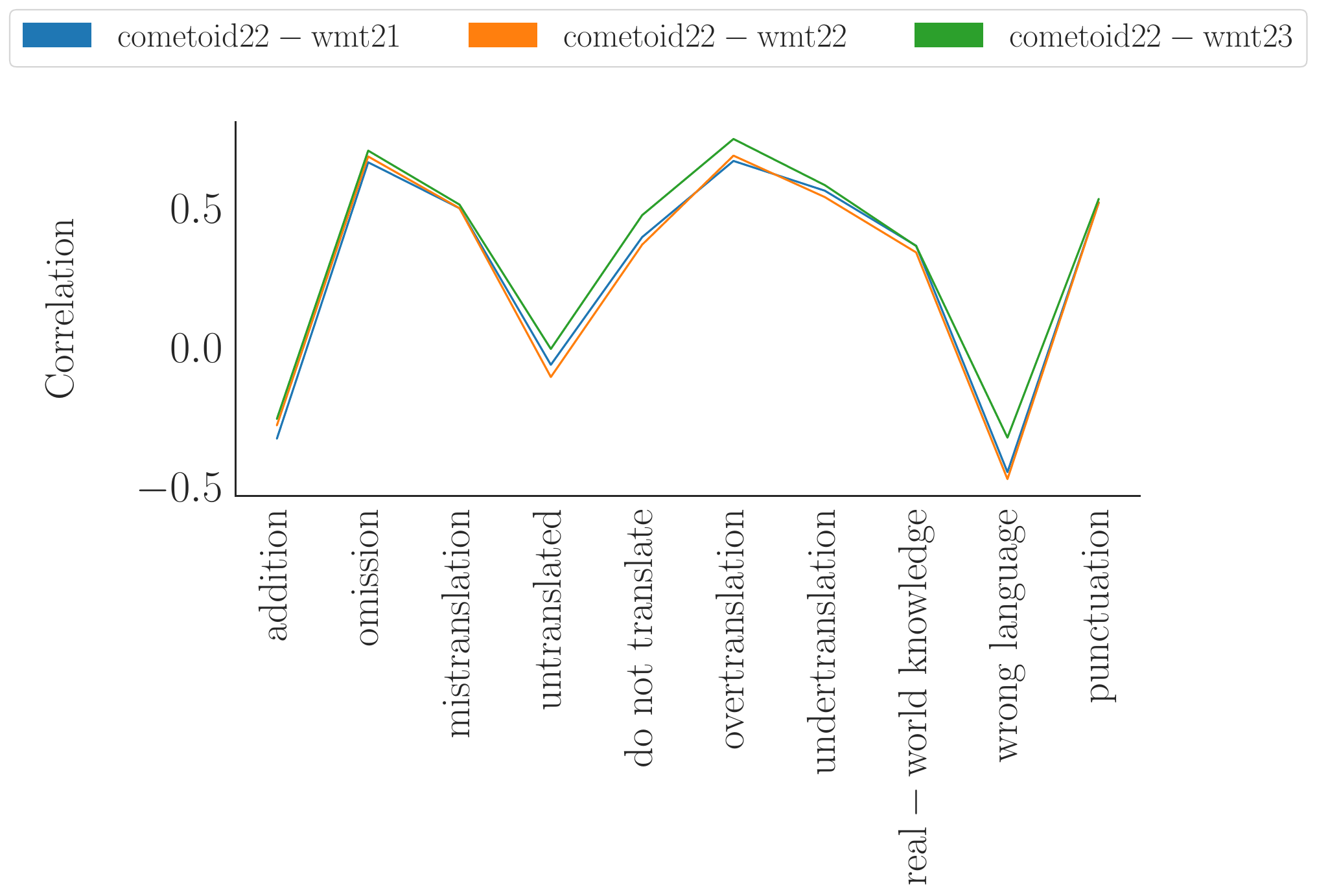}
    \caption{Correlations for different top-level phenomena categories with different models trained on successively more data.}
    \label{fig:data_effects}
\end{figure}

The \textsc{Cometoid22} submission in 2023 included three different reference-free metric versions, each trained on successively more data. This allows us to investigate the effects of the metric training data size\footnote{Note that for \textsc{Cometoid22} this is not human judgement labelled data but rather pseudo labelled data where labels come from the reference-based COMET-22 model.} on the performance on ACES. (Note that we cannot draw any conclusions about the training data size of the pretraining models that are used.) In Figure~\ref{fig:data_effects}, we can see the effect of training data size on the performance on the top-level phenomena categories. \textsc{Cometoid22-wmt23}, the model that has seen the most data outperforms the other two metrics on almost all top-level categories. The correlation gain is especially pronounced for the \textit{untranslated}, \textit{do not translate} (content in the source is erroneously translated into the target language), \textit{overtranslation} (the target translation contains more specific information than the source) and \textit{wrong language} categories (see Table~\ref{tab:ACES_top_level_examples} for examples for each of the phenomena). For clearer insights as to where the performance gain comes from, we would need to analyse the training data in depth. However, it is evident from these results that more training data is beneficial for metric development. In the next section, we look at metric score changes over metric implementation cycles - where likely more than just the training data changed.

\subsection{Changes between 2022 and 2023}
\label{subsec:results_progress}
We compare the results of metrics submitted by the same teams in both 2022 and 2023 in Table~\ref{tab:analysis_2022_2023_delta}. To make a valid comparison, we exclude the examples affected by the double quote pre-processing resulting in 33817 examples which are discussed below.


\begin{table*}[h!]
\centering
\small
\begin{tabular}{@{}lcccccc@{}}
\toprule
 & \multicolumn{2}{c}{COMETKiwi} & KG-BERTScore & \multicolumn{3}{c}{XCOMET} \\
\cmidrule{2-3}\cmidrule{5-7}
                     & -XL & -XXL & & -Ensemble & -XL & -XXL \\
\midrule
addition             & -0.120            & -0.004             & -0.251            & \phantom{-}0.595           & \phantom{-}0.455     & \phantom{-}0.142      \\
omission             & -0.004            & -0.002             & \phantom{-}0.103             & \phantom{-}0.118           & -0.126    & -0.254     \\
mistranslation       & -0.005            & \phantom{-}0.013              & \phantom{-}0.077             & \phantom{-}0.126           & \phantom{-}0.038     & \phantom{-}0.005      \\
untranslated         & \phantom{-}0.000             & \phantom{-}0.142              & \phantom{-}0.266             & -0.181          & -0.342    & -0.362     \\
do not translate     & -0.395            & -0.553             & \phantom{-}0.000             & \phantom{-}0.053           & \phantom{-}0.079     & -0.105     \\
overtranslation      & \phantom{-}0.027             & \phantom{-}0.035              & \phantom{-}0.119             & \phantom{-}0.073           & -0.067    & \phantom{-}0.017      \\
undertranslation     & -0.019            & -0.021             & \phantom{-}0.077             & \phantom{-}0.014           & -0.132    & -0.025     \\
real-world knowledge & -0.020            & \phantom{-}0.100              & \phantom{-}0.107             & \phantom{-}0.003           & -0.123    & -0.198     \\
wrong language       & -0.014            & -0.173             & -0.618            & -0.296          & -0.232    & -0.395     \\
punctuation          & -0.037            & \phantom{-}0.004              & \phantom{-}0.264             & \phantom{-}0.206           & -0.144    & \phantom{-}0.006      \\
\midrule
ACES-Score           & -1.04\phantom{1}             & -0.38\phantom{1}               & \phantom{-}0.40\phantom{1}               & 4.23\phantom{1}             & \phantom{-}0.21\phantom{1}       & -1.64\phantom{1}      \\
\bottomrule
\end{tabular}
\caption{Comparison of average Kendall’s tau-like correlation: delta calculated as 2023 score minus 2022 score.}
\label{tab:analysis_2022_2023_delta}
\end{table*}

We report changes in performance in terms of deltas, computed by subtracting the 2022 score from the 2023 score. We do this for the following pairs of metrics: \textsc{KG-BERTScore} (2022) and \textsc{KG-BERTScore} (2023); \textsc{COMETKiwi} (2022) paired with \textsc{COMETKiwi-XL} (2023) and \textsc{COMETKiwi-XXL} (2023); \textsc{COMET-22} (2022) paired with \textsc{XCOMET-Ensemble} (2023), \textsc{XCOMET-XL} (2023) and \textsc{XCOMET-XXL} (2023). 

We observe that the performance of \textsc{KG-BERTScore} improved in 2023. From the description provided by the metric developers, the main difference is that the 2023 version of \textsc{KG-BERTScore} metric uses COMET-QE instead of BERTScore \citep{DBLP:conf/iclr/ZhangKWWA20} to compute the similarity between the source and the hypothesis. Whilst we might therefore attribute the increase in performance to this change, a more systematic comparison of the two metric versions would be required to confirm whether this is the only contributing factor. 

The metrics in the \textsc{COMETKiwi} family exhibit: a slight drop in performance (\textsc{COMETKiwi-XL}) and a similar performance to that of last year (\textsc{COMETKiwi-XXL}). The difference can be attributed to changing the underlying encoder, XLM-R XL and XLM-R XXL \citep{DBLP:journals/corr/abs-2105-00572} respectively, and the use of additional fine-tuning data made available this year. We have seen that the addition of more training data helps in Section~\ref{subsec:data_effects}. Considering that there is no improvement in the performance, we question if an increase in the underlying model capacity of the encoder alone is useful for obtaining better MT evaluation.

Performance change for the XCOMET family is variable: there is a performance increase for \textsc{XCOMET-Ensemble} (compared to COMET-22), for \textsc{XCOMET-XL} the increase is smaller, and the performance of \textsc{XCOMET-XXL} is degraded. The XCOMET family is designed to provide both a quality score and an error span. Considering that the metric also provides an explanation of the scores without hurting the performance, this is indeed a positive change.
Finally, it is worth noting that for \textit{all} metrics in Table~\ref{tab:analysis_2022_2023_delta} a change in performance is observed for almost all \textsc{ACES} categories, for all metrics.

Whilst it is not possible to draw conclusions or make predictions about the future of metric development based solely on the observations from two consecutive metrics shared tasks, we highlight several high-level changes. Firstly, we note the participation of many more COMET-based metrics in 2023, compared with 2022. This is presumably based on the success of COMET at previous shared tasks and its adoption within the MT community. We find that three metrics from 2022 were used as baseline metrics in 2023: \textsc{COMET-22}, \textsc{CometKiwi}, and \textsc{MS-COMET-QE-22}. In contrast to the submissions in 2022, the 2023 submissions included some new metrics that use lexical overlap through text matching or embeddings (\textsc{Tokengram\_F}, \textsc{Partokengram\_F}, and \textsc{eBLEU}). However, their performance trend is similar to other surface overlap metrics. This year has also seen submissions based on large language models (\textsc{embed\_llama} and \textsc{GEMBA-MQM}). As seen in Section~\ref{sec:aces_overview}, their moderate performance indicates the need for more effective approaches. Additionally, we note an overall increase between 2022 and 2023 in the number of metrics submitted to WMT that a) provide segment-level scores and b) provide scores for all language pairs and directions in ACES. There were 37 segment-level metrics at WMT 2022, 24 of which covered the language pairs and directions in ACES, compared with 47 and 33, respectively in 2023. This suggests that the interest in metric development remains high, and could be increasing. In terms of metric sensitivities to error types, computed by subtracting the 2022 score from the 2023 score mostly support the observations about the changes in the metric performances evaluated using the Kendall’s tau-like correlation.

In our overview analysis in Section~\ref{sec:aces_overview} we highlight many similarities between metric performance trends in 2022 and 2023. However, there are a few differences. In 2023 we observe that the reference-free metric group performs strongly overall, compared with the baseline and reference-free groups. This could be attributed to the increase in metrics based on the COMET architecture in 2023. WMT 2023 also saw the submission of two LLM-based metrics: \textsc{embed\_llama} and \textsc{GEMBA-MQM}. Despite the success of LLMs across various tasks \citep{DBLP:journals/corr/abs-2005-14165}, the performance of both \textsc{embed\_llama} and \textsc{GEMBA-MQM} highlights that leveraging LLMs to evaluate translated outputs on segment-level still requires some improved design strategies. All of these observations suggest that evaluating MT outputs is indeed a hard problem \citep{neubig2022is}. While we do have a good suite of metrics to provide a proxy for evaluation, there are indeed several interesting challenges that need to be tackled before we find an ideal evaluation regime. And even then, we need to continuously monitor this to ensure that we do not optimise towards metric weaknesses that we have not yet discovered.

\section{Recommendations}
\label{sec:recommendations}
Based on the metrics results on \textsc{ACES}, \textsc{Span-ACES} and our analyses, we first make some recommendations for MT evaluation in general and then provide some more specific suggestions for metric development.

\textbf{Informative Evaluation}: From our results in Section~\ref{sec: Results}, we find that \textit{a single score is not enough} to identify if a metric has superior performance. By evaluating on \textsc{ACES}, we can obtain a profile for the metric showcasing its strengths and weaknesses across different MT errors, supporting metric developers in making more informed choices. To further deter the development of metrics that produce a single score, we also recommend predicting error spans (ideally with labels) instead of scores. We propose \textsc{Span-ACES} as an additional test suite for the development of metrics that produce error spans. 

\textbf{Building metric ensembles:} Both the evaluation on phenomena and language pair categories in Section~\ref{sec: Results} showed that there is no single best-performing metric. This divergence is likely to become even larger if we evaluate metrics on different domains. In future work on MT evaluation, it may be worthwhile thinking about how different metrics can be \textit{combined} to make more robust decisions as to which is the best translation. Recent submissions to the WMT Metrics shared task include ensemble models (such as) \textsc{COMET-Kiwi}, \textsc{KG-BERTSCore}, \textsc{XCOMET-Ensemble}, \textit{etc.,}), which suggests that our recommendations are aligned with the efforts of the community.

\textbf{The source matters:} Our analysis in Section~\ref{subsec:source-relevance} highlighted that many reference-based metrics that take the source as input do not consider it enough. Cases where the correct translation can only be identified through the source, are currently better handled by reference-free metrics. This is a serious shortcoming of reference-based metrics which should be addressed in future research, also considering that many reference-based metrics choose to exclude source information by design. 

\textbf{Surface overlap prevails:} In Section~\ref{subsec:surface-relevance}, we showed that despite moving beyond a purely surface-level comparison with the reference, most reference-based metrics are still considerably influenced by surface-level overlap. We thus recommend including paraphrases in the training regime as well as designing loss functions that explicitly discourage surface-level overlap. 

\textbf{Check the base model properties:} Some properties of multilingual representations, like the representation space being language-agnostic, can result in undesirable effects on MT evaluation (Section ~\ref{subsec:mutlilingual embeddings}). We also find that LLMs are not effective segment-level MT evaluators just yet (see Section~\ref{subsec:llm_metrics}), hence, better design strategies must be employed to make LLMs useful in evaluation.
Simple strategies to model language-specific information in the metrics could also improve the robustness of the metrics to adversarial language pair attacks.

\section{Conclusion}
In this work, we identify and address some of the  shortcomings of MT metrics. A single segment-level (or system-level) score for a metric does not provide an overview of that metric's strengths and weaknesses.
To address this, we developed \textsc{ACES}: a translation accuracy challenge set based on the MQM ontology, which consists of 36,476 examples covering 146 language pairs and representing challenges from 68 phenomena. \textsc{ACES} can be used to provide a profile of metric performance over a range of phenomena, and to measure incremental performance between multiple versions of the same metric. We used \textsc{ACES} to evaluate the baseline and submitted metrics from the WMT 2022 and 2023 metrics shared tasks, to measure the incremental performance of those metrics submitted in both years, to measure how sensitive metrics are to certain phenomena, and to provide fine-grained analyses of metric performance to reveal the extent to which metrics rely on the source and on surface-level overlap with the reference, and to assess whether multilingual embeddings are a helpful component in metric design.

Our overview of metric performance at the phenomena and language levels in Section~\ref{sec: Results} reveals that there is no single best-performing metric. The more fine-grained analyses in Section~\ref{sec:analysis} highlight that 1) metric sensitivity is correlated with score prediction for most of the metrics 2) many reference-based metrics that take the source as input do not consider it enough, 3) most
reference-based metric scores are still considerably influenced by surface overlap with the reference,  4) the use of multilingual embeddings can have undesirable effects on MT evaluation and 5) the addition of metric-specific data improves the quality of the metric. We find that LLM-based evaluation methods have mediocre results and in some cases even worse than the surface overlap-based metrics. 

We recommend that these shortcomings of existing metrics be addressed in future research and that metric developers should consider a) combining metrics with different strengths, e.g. in the form of ensemble models, b) developing metrics that give more weight to the source and less to surface-level overlap with the reference, and c) incorporating strategies to explicitly model additional language-specific information (rather than simply relying on multilingual embeddings). We also recommend the community develop evaluation methods that produce error types and error spans as singular scores are not informative. To that end, we released \textsc{Span-ACES} where every incorrect translation in \textsc{ACES} contains span-level annotations for the erroneous text corresponding to the phenomenon label. We also provided baseline results on \textsc{Span-ACES}. We have made \textsc{ACES} and \textsc{Span-ACES} publicly available and hope that it will provide a useful benchmark for MT evaluation metric developers in the future.

\appendix

\appendixsection{Language Codes}

\begin{table}[ht]
\begin{tabular}{ll|ll|ll|ll}
\hline
\textbf{Code} & \textbf{Language} & 
\textbf{Code} & \textbf{Language} &
\textbf{Code} & \textbf{Language} &
\textbf{Code} & \textbf{Language} \\ \hline
af   &  Afrikaans   & fa   & Persian    & ja   & Japanese   & sl   & Slovenian  \\
ar   &  Arabic      & fi   & Finnish    & ko   & Korean     & sr   & Serbian     \\
be   & Belarusian   & fr   & French     & lt   & Lithuanian & sv   & Swedish    \\
bg   & Bulgarian    & ga   & Irish      & lv   & Latvian    & sw   & Swahili    \\
ca   & Catalan      & gl   & Galician   & mr   & Marathi    & ta   & Tamil    \\
cs   & Czech        & he   & Hebrew     & nl   & Dutch      & th   & Thai      \\
da   & Danish       & hi   & Hindi      & no   & Norwegian  & tr   & Turkish  \\
de   & German       & hr   & Croatian   & pl   & Polish     & uk   & Ukranian     \\
el   & Greek        & hu   & Hungarian  & pt   & Portuguese & ur   & Urdu \\
en   & English      & hy   & Armenian   & ro   & Romanian   & vi   & Vietnamese   \\
es   & Spanish      & id   & Indonesian & ru   & Russian    & wo   & Wolof    \\
et   & Estonian     & it   & Italian    & sk   & Slovak     & zh   & Chinese     \\
\hline      
\end{tabular}
\caption{ISO 2-Letter language codes of the languages included in the challenge set}
\label{lang_codes}
\end{table}

\appendixsection{Permitted Unit Conversions}
\label{app:allowed-units}

The unit conversions permitted for the \textit{\hyperref[p:unit-conversion]{Hallucination - Unit Conversion}} challenge set are listed in Table~\ref{tab:unit_conversions}.

\begin{table*}[h!]
\centering
\small
\begin{tabular}{ll}
\toprule
\textbf{Distance}: & \textbf{Volume}: \\
\tabitem miles $\rightarrow$ metres & \tabitem barrels $\rightarrow$ gallons \\
\tabitem kilometres $\rightarrow$ miles & \tabitem barrels $\rightarrow$ litres \\
\tabitem kilometres $\rightarrow$ metres & \tabitem gallons $\rightarrow$ barrels \\
\tabitem metres $\rightarrow$ feet & \tabitem gallons $\rightarrow$ litres \\
\tabitem metres $\rightarrow$ yards &  \\
\tabitem feet $\rightarrow$ metres & \textbf{Weight}: \\
\tabitem feet $\rightarrow$ yards & \tabitem kilograms $\rightarrow$ grams \\
\tabitem centimetres $\rightarrow$ inches & \tabitem kilograms $\rightarrow$ pounds \\
\tabitem centimetres $\rightarrow$ millimetres & \tabitem grams $\rightarrow$ ounces \\
\tabitem inches $\rightarrow$ centimetres & \tabitem ounces $\rightarrow$ grams \\
\tabitem inches $\rightarrow$ millimetres &  \\
\tabitem millimetres $\rightarrow$ centimetres & \textbf{Time}: \\
\tabitem millimetres $\rightarrow$ inches & \tabitem hours $\rightarrow$ minutes \\
& \tabitem minutes $\rightarrow$ seconds \\
\textbf{Speed}: & \tabitem seconds $\rightarrow$ minutes \\
\tabitem miles per hour $\rightarrow$ kilometres per hour & \tabitem days $\rightarrow$ hours \\
\tabitem kilometres per hour $\rightarrow$ miles per hour & \tabitem months $\rightarrow$ weeks \\
\tabitem kilometres per second $\rightarrow$ miles per second & \tabitem weeks $\rightarrow$ days \\
\tabitem miles per second $\rightarrow$ kilometres per second &  \\
& \\
\textbf{Area}: & \\
\tabitem square kilometres $\rightarrow$ square miles & \\
\bottomrule
\end{tabular}
\caption{Permitted Unit Conversions}
\label{tab:unit_conversions}
\end{table*}

\appendixsection{Distribution of Examples Across Language Pairs}
\label{app:language_pair_matrix}
\begin{sidewaystable*}[]
\tiny
\setlength\tabcolsep{3.15 pt}
\renewcommand{\arraystretch}{1.15}
\begin{tabular}{ll|c|c|c|c|c|c|c|c|c|c|c|c|c|c|c|c|c|c|c|c|c|c|c|c|c|c|c|c|c|c|c|c|c|c|c|c|c|c|c|c|c|c|c|c|c|c|c|c|}
\multicolumn{1}{c}{} & \multicolumn{1}{c}{} & \multicolumn{48}{l}{\textbf{tgt $\rightarrow$}} \\

\multicolumn{1}{c}{} & \multicolumn{1}{c}{} & \multicolumn{1}{c}{af} & \multicolumn{1}{c}{ar} & \multicolumn{1}{c}{be} & \multicolumn{1}{c}{bg} & \multicolumn{1}{c}{ca} & \multicolumn{1}{c}{cs} & \multicolumn{1}{c}{da} & \multicolumn{1}{c}{de} & \multicolumn{1}{c}{el} & \multicolumn{1}{c}{en} & \multicolumn{1}{c}{es} & \multicolumn{1}{c}{et} & \multicolumn{1}{c}{fa} & \multicolumn{1}{c}{fi} & \multicolumn{1}{c}{fr} & \multicolumn{1}{c}{ga} & \multicolumn{1}{c}{gl} & \multicolumn{1}{c}{he} & \multicolumn{1}{c}{hi} & \multicolumn{1}{c}{hr} & \multicolumn{1}{c}{hu} & \multicolumn{1}{c}{hy} & \multicolumn{1}{c}{id} & \multicolumn{1}{c}{it} & \multicolumn{1}{c}{ja} & \multicolumn{1}{c}{ko} & \multicolumn{1}{c}{lt} & \multicolumn{1}{c}{lv} & \multicolumn{1}{c}{mr} & \multicolumn{1}{c}{nl} & \multicolumn{1}{c}{no} & \multicolumn{1}{c}{pl} & \multicolumn{1}{c}{pt} & \multicolumn{1}{c}{ro} & \multicolumn{1}{c}{ru} & \multicolumn{1}{c}{sk} & \multicolumn{1}{c}{sl} & \multicolumn{1}{c}{sr} & \multicolumn{1}{c}{sv} & \multicolumn{1}{c}{sw} & \multicolumn{1}{c}{ta} & \multicolumn{1}{c}{th} & \multicolumn{1}{c}{tr} & \multicolumn{1}{c}{uk} & \multicolumn{1}{c}{ur} & \multicolumn{1}{c}{vi} & \multicolumn{1}{c}{wo} & \multicolumn{1}{c}{zh}\\
\hhline{~~------------------------------------------------}
\textbf{src} & af &  &  &  &  &  &  &  &  &  & \cellcolor{red!10}{96} &  &  & \cellcolor{red!10}{25} &  &  &  &  &  &  &  &  &  &  &  &  &  &  &  &  &  &  &  &  &  &  &  &  &  &  &  &  &  &  &  &  &  &  & \\
\hhline{~~------------------------------------------------}
\textbf{$\downarrow$} & ar &  &  &  &  &  &  &  &  &  & \cellcolor{red!25}{361} &  &  &  &  & \cellcolor{red!25}{102} &  &  &  & \cellcolor{red!10}{17} &  &  &  &  &  &  &  &  &  &  &  &  &  &  &  &  &  &  &  &  &  &  &  &  &  &  &  &  & \\
\hhline{~~------------------------------------------------}
 & be &  &  &  &  &  &  &  &  &  & \cellcolor{red!10}{67} &  &  &  &  &  &  &  &  &  &  &  &  &  &  &  &  &  &  &  &  &  &  &  &  &  &  &  &  &  &  &  &  &  &  &  &  &  & \\
\hhline{~~------------------------------------------------}
 & bg &  &  &  &  &  &  &  &  &  & \cellcolor{red!25}{393} &  &  &  &  &  &  &  &  &  &  &  &  &  &  &  &  & \cellcolor{red!10}{40} &  &  &  &  &  &  &  &  &  &  &  &  &  &  &  &  &  &  &  &  & \\
\hhline{~~------------------------------------------------}
 & ca &  &  &  &  &  &  &  &  &  & \cellcolor{red!10}{79} & \cellcolor{red!25}{175} &  &  &  &  &  &  &  &  &  &  &  &  &  &  &  &  &  &  &  &  &  &  &  &  &  &  &  &  &  &  &  &  &  &  &  &  & \\
\hhline{~~------------------------------------------------}
 & cs &  &  &  &  &  &  &  &  &  & \cellcolor{red!10}{85} &  &  &  &  &  &  &  &  &  &  &  &  &  &  &  &  &  &  &  &  &  &  &  &  &  &  &  &  &  &  &  &  &  &  &  &  &  & \\
\hhline{~~------------------------------------------------}
 & da &  &  &  &  &  &  &  &  &  & \cellcolor{red!10}{83} &  &  &  &  &  &  &  &  &  &  &  &  &  &  &  &  &  &  &  &  &  &  &  &  &  &  &  &  &  &  &  &  &  &  &  &  &  & \\
\hhline{~~------------------------------------------------}
 & de &  &  &  &  &  &  &  &  &  & \cellcolor{red!40}{4163} & \cellcolor{red!10}{84} &  &  &  & \cellcolor{red!25}{394} &  &  &  &  &  &  &  &  &  & \cellcolor{red!25}{113} & \cellcolor{red!10}{63} &  &  &  &  &  &  &  &  & \cellcolor{red!25}{104} &  &  &  &  &  &  &  &  &  &  &  &  & \cellcolor{red!10}{75}\\
\hhline{~~------------------------------------------------}
 & el &  &  &  &  &  &  &  &  &  & \cellcolor{red!25}{387} &  &  &  &  &  &  &  &  &  &  &  &  &  &  &  &  &  &  &  &  &  &  &  &  &  &  &  &  &  &  &  &  &  &  &  &  &  & \\
\hhline{~~------------------------------------------------}
 & en & \cellcolor{red!10}{25} & \cellcolor{red!10}{5} & \cellcolor{red!10}{6} & \cellcolor{red!10}{15} & \cellcolor{red!25}{347} & \cellcolor{red!25}{368} & \cellcolor{red!10}{46} & \cellcolor{red!40}{6964} & \cellcolor{red!10}{21} &  & \cellcolor{red!25}{725} & \cellcolor{red!10}{25} & \cellcolor{red!10}{20} & \cellcolor{red!10}{12} & \cellcolor{red!25}{800} &  & \cellcolor{red!10}{16} & \cellcolor{red!10}{18} & \cellcolor{red!25}{343} & \cellcolor{red!10}{27} & \cellcolor{red!10}{44} & \cellcolor{red!10}{3} & \cellcolor{red!10}{31} & \cellcolor{red!10}{10} & \cellcolor{red!25}{430} & \cellcolor{red!25}{545} & \cellcolor{red!10}{17} & \cellcolor{red!10}{19} & \cellcolor{red!10}{52} & \cellcolor{red!10}{50} & \cellcolor{red!10}{53} & \cellcolor{red!25}{349} & \cellcolor{red!10}{44} & \cellcolor{red!10}{46} & \cellcolor{red!25}{698} & \cellcolor{red!10}{27} & \cellcolor{red!10}{45} & \cellcolor{red!10}{15} & \cellcolor{red!10}{39} &  & \cellcolor{red!10}{1} &  & \cellcolor{red!10}{10} & \cellcolor{red!10}{16} & \cellcolor{red!10}{10} & \cellcolor{red!10}{25} &  & \cellcolor{red!25}{333}\\
\hhline{~~------------------------------------------------}
 & es &  &  &  &  & \cellcolor{red!10}{88} &  &  & \cellcolor{red!10}{64} &  & \cellcolor{red!40}{1263} &  &  &  &  & \cellcolor{red!25}{125} &  &  &  &  &  &  &  &  &  & \cellcolor{red!25}{117} & \cellcolor{red!10}{74} &  &  &  &  &  &  &  &  &  &  &  &  &  &  &  &  &  &  &  &  &  & \cellcolor{red!10}{67}\\
\hhline{~~------------------------------------------------}
 & et &  &  &  &  &  &  &  &  &  & \cellcolor{red!10}{70} &  &  &  &  &  &  &  &  &  &  &  &  &  &  &  &  &  &  &  &  &  &  &  &  &  &  &  &  &  &  &  &  &  &  &  &  &  & \\
\hhline{~~------------------------------------------------}
 & fa & \cellcolor{red!10}{16} &  &  &  &  &  &  &  &  & \cellcolor{red!10}{85} &  &  &  &  &  &  &  &  &  &  &  &  &  &  &  &  &  &  &  &  &  &  &  &  &  &  &  &  &  &  &  &  &  &  &  &  &  & \\
\hhline{~~------------------------------------------------}
 & fi &  &  &  &  &  &  &  &  &  & \cellcolor{red!10}{79} &  &  &  &  &  &  &  &  &  &  &  &  &  &  &  &  &  &  &  &  &  &  &  &  &  &  &  &  &  &  &  &  &  &  &  &  &  & \\
\hhline{~~------------------------------------------------}
 & fr &  &  &  &  &  &  &  & \cellcolor{red!25}{683} &  & \cellcolor{red!40}{2868} & \cellcolor{red!10}{78} &  &  &  &  &  &  &  &  &  &  &  &  &  & \cellcolor{red!25}{403} & \cellcolor{red!10}{59} &  &  & \cellcolor{red!25}{344} &  &  &  &  &  & \cellcolor{red!10}{46} &  &  &  &  &  &  &  &  &  &  &  &  & \cellcolor{red!10}{61}\\
\hhline{~~------------------------------------------------}
 & ga &  &  &  &  &  &  &  &  &  & \cellcolor{red!10}{17} &  &  &  &  &  &  &  &  &  &  &  &  &  &  &  &  &  &  &  &  &  &  &  &  &  &  &  &  &  &  &  &  &  &  &  &  &  & \\
\hhline{~~------------------------------------------------}
 & gl &  &  &  &  &  &  &  &  &  & \cellcolor{red!10}{70} &  &  &  &  &  &  &  &  &  &  &  &  &  &  &  &  &  &  &  &  &  &  &  &  &  &  &  &  &  &  &  &  &  &  &  &  &  & \\
\hhline{~~------------------------------------------------}
 & he &  &  &  &  &  &  &  &  &  & \cellcolor{red!10}{59} &  &  &  &  &  &  &  &  &  &  &  &  &  &  &  &  &  &  &  &  &  &  &  &  &  &  &  &  & \cellcolor{red!10}{51} &  &  &  &  &  &  &  &  & \\
\hhline{~~------------------------------------------------}
 & hi &  & \cellcolor{red!10}{8} &  &  &  &  &  &  &  & \cellcolor{red!25}{367} &  &  &  &  &  &  &  &  &  &  &  &  &  &  &  &  &  &  &  &  &  &  &  &  &  &  &  &  &  &  &  &  &  &  &  &  &  & \\
\hhline{~~------------------------------------------------}
 & hr &  &  &  &  &  &  &  &  &  & \cellcolor{red!10}{81} &  &  &  &  &  &  &  &  &  &  &  &  &  &  &  &  &  & \cellcolor{red!10}{29} &  &  &  &  &  &  &  &  &  &  &  &  &  &  &  &  &  &  &  & \\
\hhline{~~------------------------------------------------}
 & hu &  &  &  &  &  &  &  &  &  & \cellcolor{red!10}{53} &  &  &  &  &  &  &  &  &  &  &  &  &  &  &  &  &  &  &  &  &  &  &  &  &  &  &  &  &  &  &  &  &  &  &  &  &  & \\
\hhline{~~------------------------------------------------}
 & hy &  &  &  &  &  &  &  &  &  & \cellcolor{red!10}{48} &  &  &  &  &  &  &  &  &  &  &  &  &  &  &  &  &  &  &  &  &  &  &  &  &  &  &  &  &  &  &  &  &  &  &  & \cellcolor{red!10}{13} &  & \\
\hhline{~~------------------------------------------------}
 & id &  &  &  &  &  &  &  &  &  & \cellcolor{red!10}{63} &  &  &  &  &  &  &  &  &  &  &  &  &  &  &  &  &  &  &  &  &  &  &  &  &  &  &  &  &  &  &  &  &  &  &  &  &  & \\
\hhline{~~------------------------------------------------}
 & it &  &  &  &  &  &  &  &  &  & \cellcolor{red!25}{801} &  &  &  &  &  &  &  &  &  &  &  &  &  &  &  &  &  &  &  &  &  &  &  &  &  &  &  &  &  &  &  &  &  &  &  &  &  & \\
\hhline{~~------------------------------------------------}
 & ja &  &  &  &  &  &  &  & \cellcolor{red!10}{60} &  & \cellcolor{red!25}{912} & \cellcolor{red!10}{67} &  &  &  & \cellcolor{red!25}{122} &  &  &  &  &  &  &  &  &  &  & \cellcolor{red!25}{163} &  &  &  &  &  &  &  &  &  &  &  &  &  &  &  &  &  &  &  &  &  & \cellcolor{red!10}{74}\\
\hhline{~~------------------------------------------------}
 & ko &  &  &  &  &  &  &  & \cellcolor{red!10}{70} &  & \cellcolor{red!40}{1004} & \cellcolor{red!10}{72} &  &  &  & \cellcolor{red!25}{110} &  &  &  &  &  &  &  &  &  & \cellcolor{red!25}{358} &  &  &  &  &  &  &  &  &  &  &  &  &  &  &  &  &  &  &  &  &  &  & \cellcolor{red!10}{73}\\
\hhline{~~------------------------------------------------}
 & lt &  &  &  & \cellcolor{red!10}{28} &  &  &  &  &  & \cellcolor{red!10}{68} &  &  &  &  &  &  &  &  &  &  &  &  &  &  &  &  &  &  &  &  &  &  &  &  &  &  &  &  &  &  &  &  &  &  &  &  &  & \\
\hhline{~~------------------------------------------------}
 & lv &  &  &  &  &  &  &  &  &  & \cellcolor{red!10}{61} &  &  &  &  &  &  &  &  &  & \cellcolor{red!10}{24} &  &  &  &  &  &  &  &  &  &  &  &  &  &  &  &  &  &  &  &  &  &  &  &  &  &  &  & \\
\hhline{~~------------------------------------------------}
 & mr &  &  &  &  &  &  &  &  &  & \cellcolor{red!10}{63} &  &  &  &  &  &  &  &  &  &  &  &  &  &  &  &  &  &  &  &  &  &  &  &  &  &  &  &  &  &  &  &  &  &  &  &  &  & \\
\hhline{~~------------------------------------------------}
 & nl &  &  &  &  &  &  &  &  &  & \cellcolor{red!10}{73} &  &  &  &  &  &  &  &  &  &  &  &  &  &  &  &  &  &  &  &  &  &  &  &  &  &  &  &  &  &  &  &  &  &  &  &  &  & \\
\hhline{~~------------------------------------------------}
 & no &  &  &  &  &  &  &  &  &  & \cellcolor{red!10}{53} &  &  &  &  &  &  &  &  &  &  &  &  &  &  &  &  &  &  &  &  &  &  &  &  &  &  &  &  &  &  &  &  &  &  &  &  &  & \\
\hhline{~~------------------------------------------------}
 & pl &  &  &  &  &  &  &  &  &  & \cellcolor{red!10}{65} &  &  &  &  &  &  &  &  &  &  &  &  &  &  &  &  &  &  & \cellcolor{red!25}{111} &  &  &  &  &  &  & \cellcolor{red!10}{58} &  &  &  &  &  &  &  &  &  &  &  & \\
\hhline{~~------------------------------------------------}
 & pt &  &  &  &  &  &  &  &  &  & \cellcolor{red!10}{89} &  &  &  &  &  &  &  &  &  &  &  &  &  &  &  &  &  &  &  &  &  &  &  &  &  &  &  & \cellcolor{red!10}{40} &  &  &  &  &  &  &  &  &  & \\
\hhline{~~------------------------------------------------}
 & ro &  &  &  &  &  &  &  &  &  & \cellcolor{red!10}{91} &  &  &  &  &  &  &  &  &  &  &  &  &  &  &  &  &  &  &  &  &  &  &  &  &  &  &  &  &  &  &  &  &  &  &  &  &  & \\
\hhline{~~------------------------------------------------}
 & ru &  &  &  &  &  &  &  & \cellcolor{red!25}{106} &  & \cellcolor{red!25}{472} & \cellcolor{red!10}{87} &  &  &  & \cellcolor{red!10}{42} &  &  &  &  &  &  &  &  &  &  &  &  &  &  &  &  &  &  &  &  &  &  &  &  &  &  &  &  &  &  &  &  & \\
\hhline{~~------------------------------------------------}
 & sk &  &  &  &  &  &  &  &  &  & \cellcolor{red!10}{54} &  &  &  &  &  &  &  &  &  &  &  &  &  &  &  &  &  &  &  &  &  & \cellcolor{red!10}{17} &  &  &  &  &  &  &  &  &  &  &  &  &  &  &  & \\
\hhline{~~------------------------------------------------}
 & sl &  &  &  &  &  &  &  &  &  & \cellcolor{red!10}{69} &  &  &  &  &  &  &  &  &  &  &  &  &  &  &  &  &  &  &  &  &  &  &  &  &  &  &  &  &  &  &  &  &  &  &  &  &  & \\
\hhline{~~------------------------------------------------}
 & sr &  &  &  &  &  &  &  &  &  & \cellcolor{red!10}{64} &  &  &  &  &  &  &  &  &  &  &  &  &  &  &  &  &  &  &  &  &  &  & \cellcolor{red!10}{54} &  &  &  &  &  &  &  &  &  &  &  &  &  &  & \\
\hhline{~~------------------------------------------------}
 & sv &  &  &  &  &  &  &  &  &  & \cellcolor{red!10}{79} &  &  &  &  &  &  &  & \cellcolor{red!10}{28} &  &  &  &  &  &  &  &  &  &  &  &  &  &  &  &  &  &  &  &  &  &  &  &  &  &  &  &  &  & \\
\hhline{~~------------------------------------------------}
 & sw &  &  &  &  &  &  &  &  &  & \cellcolor{red!25}{327} &  &  &  &  &  &  &  &  &  &  &  &  &  &  &  &  &  &  &  &  &  &  &  &  &  &  &  &  &  &  &  &  &  &  &  &  &  & \\
\hhline{~~------------------------------------------------}
 & ta &  &  &  &  &  &  &  &  &  & \cellcolor{red!10}{39} &  &  &  &  &  &  &  &  &  &  &  &  &  &  &  &  &  &  &  &  &  &  &  &  &  &  &  &  &  &  &  &  &  &  &  &  &  & \\
\hhline{~~------------------------------------------------}
 & th &  &  &  &  &  &  &  &  &  & \cellcolor{red!25}{299} &  &  &  &  &  &  &  &  &  &  &  &  &  &  &  &  &  &  &  &  &  &  &  &  &  &  &  &  &  &  &  &  &  &  &  &  &  & \\
\hhline{~~------------------------------------------------}
 & tr &  &  &  &  &  &  &  &  &  & \cellcolor{red!25}{386} &  &  &  &  &  &  &  &  &  &  &  &  &  &  &  &  &  &  &  &  &  &  &  &  &  &  &  &  &  &  &  &  &  &  &  &  &  & \\
\hhline{~~------------------------------------------------}
 & uk &  &  &  &  &  &  &  &  &  & \cellcolor{red!10}{77} &  &  &  &  &  &  &  &  &  &  &  &  &  &  &  &  &  &  &  &  &  &  &  &  &  &  &  &  &  &  &  &  &  &  &  &  &  & \\
\hhline{~~------------------------------------------------}
 & ur &  &  &  &  &  &  &  &  &  & \cellcolor{red!25}{372} &  &  &  &  &  &  &  &  &  &  &  &  &  &  &  &  &  &  &  &  &  &  &  &  &  &  &  &  &  &  &  &  &  &  &  &  &  & \\
\hhline{~~------------------------------------------------}
 & vi &  &  &  &  &  &  &  &  &  & \cellcolor{red!25}{391} &  &  &  &  &  &  &  &  &  &  &  & \cellcolor{red!10}{3} &  &  &  &  &  &  &  &  &  &  &  &  &  &  &  &  &  &  &  &  &  &  &  &  &  & \\
\hhline{~~------------------------------------------------}
 & wo &  &  &  &  &  &  &  &  &  & \cellcolor{red!10}{11} &  &  &  &  &  &  &  &  &  &  &  &  &  &  &  &  &  &  &  &  &  &  &  &  &  &  &  &  &  &  &  &  &  &  &  &  &  & \\
\hhline{~~------------------------------------------------}
 & zh &  &  &  &  &  &  &  & \cellcolor{red!25}{150} &  & \cellcolor{red!40}{1209} & \cellcolor{red!10}{59} &  &  &  & \cellcolor{red!25}{113} &  &  &  &  &  &  &  &  &  & \cellcolor{red!25}{128} & \cellcolor{red!10}{80} &  &  &  &  &  &  &  &  &  &  &  &  &  &  &  &  &  &  &  &  &  & \\
\hhline{~~------------------------------------------------}
\end{tabular}
\caption{Number of examples per language pair. Rows: source language; Columns: target language}
\label{lang_pair_matrix}
\end{sidewaystable*}

Table~\ref{lang_pair_matrix} contains the total number of examples per language pair in the challenge set. As can be seen in the table, the distribution of examples is variable across language pairs. The dominant language pairs are: en-de, de-en, and fr-en.

\appendixsection{Distribution of Language Pairs Across Phenomena}
\label{app:language_pair_phenomena}
\begin{sidewaystable*}[]
\tiny

\begin{tabular}{p{0.24\linewidth} | p{0.23\linewidth} | p{0.19\linewidth} | p{0.23\linewidth}}
\hline
phenomena &
  language pairs &
  phenomena &
  language pair \\ \hline
\begin{tabular}[c]{@{}l@{}}ambiguous-translation-wrong-\\ discourse-connective-since-causal\\
ambiguous-translation-wrong-\\ discourse-connective-since-temporal\\hallucination-unit-conversion-unit-matches-ref\end{tabular} &
  fr-en, de-en &
  hallucination-real-data-vs-ref-word &
  en-de, de-en, fr-de \\ \hline
ambiguous-translation-wrong-discourse-connective-while-contrast &
  fr-en &
  hallucination-real-data-vs-synonym &
  en-mr, de-en, en-de, fr-de \\ \hline
ambiguous-translation-wrong-discourse-connective-while-temporal &
  fr-en &
  untranslated-vs-ref-word &
  en-de, de-en, fr-de \\ \hline
ambiguous-translation-wrong-gender-female-anti &
  fr-en, de-en, it-en &
  untranslated-vs-synonym &
  en-de, de-en, fr-de \\ \hline
ambiguous-translation-wrong-gender-male-anti &
  fr-en, de-en, it-en &
  modal\_verb:deletion &
  de-en \\ \hline
ambiguous-translation-wrong-gender-male-pro &
  fr-en, de-en, it-en &
  modal\_verb:substitution &
  de-en \\ \hline
ambiguous-translation-wrong-sense-frequent &
  en-de, en-ru &
  nonsense &
  ko-en, ko-ja, en-ko, fr-ja, de-en \\ \hline
ambiguous-translation-wrong-sense-infrequent &
  en-de, en-ru &
  ordering-mismatch &
  en-de, de-en, fr-de \\ \hline
anaphoric\_group\_it-they:deletion &
  en-de &
  overly-literal-vs-correct-idiom &
  en-de, de-en \\ \hline
anaphoric\_group\_it-they:substitution &
  en-de &
  overly-literal-vs-explanation &
  en-de, de-en \\ \hline
anaphoric\_intra\_non-subject\_it:deletion &
  en-de &
  overly-literal-vs-ref-word &
  en-de, de-en, fr-de \\ \hline
anaphoric\_intra\_non-subject\_it:substitution &
  en-de &
  overly-literal-vs-synonym &
  en-mr, de-en, en-de, fr-de \\ \hline
anaphoric\_intra\_subject\_it:deletion &
  en-de &
  pleonastic\_it:deletion &
  en-de \\ \hline
anaphoric\_intra\_subject\_it:substitution &
  en-de &
  pleonastic\_it:substitution\_pro\_trans\_different\_to\_ref &
  en-de \\ \hline
anaphoric\_intra\_they:deletion &
  en-de &
  punctuation:deletion\_all &
  en-de \\ \hline
anaphoric\_intra\_they:substitution &
  en-de &
  punctuation:deletion\_commas &
  en-de \\ \hline
anaphoric\_singular\_they:deletion &
  en-de &
  punctuation:deletion\_quotes &
  en-de \\ \hline
anaphoric\_singular\_they:substitution &
  en-de &
  \begin{tabular}[c]{@{}l@{}}punctuation:statement-to-question\\ do-not-translate\end{tabular}
   &
  en-de \\ \hline
antonym-replacement &
  fr-en, ko-en, ja-en, es-en, zh-en, de-en &
  real-world-knowledge-entailment &
  en-de, de-en \\ \hline
similar-language-high &
  en-hi, en-cs, en-es &
  real-world-knowledge-hypernym-vs-distractor &
  en-de, de-en \\ \hline
similar-language-low &
  fr-mr, en-pl, en-ca &
  real-world-knowledge-hypernym-vs-hyponym &
  en-de, de-en \\ \hline
\begin{tabular}[c]{@{}l@{}}coreference-based-\\ on-commonsense\end{tabular} &
  en-de, en-ru, en-fr &
  real-world-knowledge-synonym-vs-antonym &
  en-de, de-en \\ \hline
\begin{tabular}[c]{@{}l@{}}hallucination-named-entity-level-1\\ hallucination-named-entity-level-2\\ hallucination-named-entity-level-3\\ hallucination-number-level-1\\ hallucination-number-level-2\\ hallucination-number-level-3\end{tabular} &
  en-de, ja-de, en-ko, de-zh, ja-en, es-de, fr-en, es-ko, ko-ja, es-ja, de-ja, zh-es, fr-zh, fr-ja, es-en, fr-ko, zh-en, ko-de, ko-es, de-ko, ko-en, fr-es, ja-es, ja-ko, zh-fr, en-es, de-en, ja-fr, ko-zh, en-fr, de-fr, ko-fr, es-fr, zh-ko, fr-de, ja-zh, de-es, es-zh, en-ja, zh-de, en-zh, zh-ja &
  \begin{tabular}[c]{@{}l@{}}undertranslation\\ overtranslation\end{tabular} &
  fr-en, ko-en, ja-en, es-en, zh-en, de-en \\ \hline
lexical-overlap &
  fr-en, en-fr, de-fr, ko-en, es-ja, ja-en, ko-fr, es-fr, ko-ja, de-ja, zh-en, ja-fr, zh-fr, en-ja, es-en, fr-ja, de-en, zh-ja &
  \begin{tabular}[c]{@{}l@{}}xnli-addition-contradiction\\ xnli-addition-neutral\\ xnli-omission-contradiction\\ xnli-omission-neutral\end{tabular} &
  fr-en, vi-en, sw-en, tr-en, zh-en, ru-en, bg-en, el-en, th-en, es-en, hi-en, de-en, ar-en, ur-en \\ \hline
\begin{tabular}[c]{@{}l@{}}hallucination-unit-conversion-amount-matches-ref\\ hallucination-unit-conversion-unit-matches-ref\end{tabular} &
  et-en, wo-en, da-en, no-en, uk-en, ta-en, fi-en, pl-en, ja-en, hy-en, ur-en, hr-en, fr-en, lt-en, tr-en, he-en, bg-en, ro-en, sv-en, ru-en, es-en, nl-en, zh-en, hu-en, be-en, lv-en, ko-en, ga-en, sk-en, af-en, sl-en, sr-en, ca-en, de-en, mr-en, id-en, vi-en, gl-en, pt-en, fa-en, hi-en, el-en, ar-en, it-en, cs-en &
  hallucination-date-time &
  en-de, et-en, ca-es, en-et, hr-lv, da-en, no-en, uk-en, fi-en, en-da, ta-en, pl-en, ja-en, en-hr, hy-en, ur-en, fr-en, hr-en, lt-en, sr-pt, en-sv, tr-en, en-no, en-sl, he-en, pl-sk, ru-en, ro-en, sv-en, en-lt, es-en, en-nl, nl-en, bg-en, he-sv, zh-en, hu-en, be-en, lv-hr, lv-en, bg-lt, en-ro, sk-pl, ko-en, ga-en, sk-en, af-en, sl-en, en-hu, sr-en, en-es, ca-en, en-sk, de-en, mr-en, id-en, vi-en, gl-en, en-fr, de-fr, pt-en, fr-de, en-pt, fa-en, hi-en, el-en, ar-en, it-en, en-pl, cs-en \\ \hline
\begin{tabular}[c]{@{}l@{}}commonsense-only-ref-ambiguous\\ commonsense-src-and-ref-ambiguous\end{tabular} &
  en-de, fr-en, ru-fr, en-fr, de-fr, ru-de, fr-de, ru-en, en-ru, fr-ru, de-ru, de-en &
  copy-source &
  ar-fr, ru-es, ur-en, fr-en, tr-en, zh-de, bg-en, ru-en, es-en, zh-en, sw-en, ja-ko, th-en, de-en, pl-mr, vi-en, hi-en, el-en, ar-en \\ \hline
\begin{tabular}[c]{@{}l@{}}addition\\ omission\end{tabular} &
  en-ca, en-el, en-et, en-ta, pl-en, hr-en, he-en, pl-sk, en-ar, ru-en, en-fi, zh-en, hu-en, be-en, lv-hr, en-he, ko-en, en-fa, sl-en, ca-en, en-gl, en-tr, en-sk, de-en, en-sr, fa-af, fa-en, ar-en, cs-en, en-de, en-hy, ar-hi, no-en, uk-en, fi-en, en-be, sr-pt, en-ru, sv-en, nl-en, sk-pl, en-hi, en-hu, mr-en, hi-ar, id-en, gl-en, en-fr, en-lv, fr-de, ca-es, en-uk, &
  \begin{tabular}[c]{@{}l@{}}addition\\ omission\end{tabular} &
  en-ur, en-hr, ur-en, en-no, en-sl, ro-en, en-vi, en-lt, es-en, en-nl, he-sv, en-it, en-ro, af-fa, en-id, lt-bg, en-af, af-en, es-ca, vi-en, sv-he, de-fr, pt-en, en-pl, et-en, hr-lv, wo-en, da-en, en-ko, en-da, ja-en, hy-en, pt-sr, hy-vi, fr-en, en-cs, lt-en, en-sv, tr-en, bg-en, lv-en, bg-lt, sr-en, en-es, en-bg, en-pt, hi-en, el-en, it-en \\ \hline
\end{tabular}
\caption{Collection of list of languages per phenomena}
\label{tab:lang-phenomena}

\end{sidewaystable*}

Table~\ref{tab:lang-phenomena} contains the list of language pairs per phenomena in the challenge set. As can be seen in the table, the distribution of language pairs is variable across phenomena. Addition and omission have the highest variety of language pairs. en-de is the most frequent language pair across all phenomena. 

\appendixsection{\textsc{ACES} Annotation Methods per Phenomena}
\label{app:automatic_annotations}
\begin{table*}[]
\footnotesize
\begin{center}
\begin{tabular}{l|l}
\hline
\textbf{Phenomenon} & \textbf{Annotation Method} \\ 
\hline
    addition & \hyperref[p:span_annotate_word]{addition/omissions} \\
    ambiguous-translation-wrong-discourse-connective-since-causal & \hyperref[p:span_diff_flexible]{word-lvl-compare-to-good} \\
    ambiguous-translation-wrong-discourse-connective-since-temporal & \hyperref[p:span_diff_flexible]{word-lvl-compare-to-good} \\
    ambiguous-translation-wrong-discourse-connective-while-contrast & \hyperref[p:span_diff_flexible]{word-lvl-compare-to-good} \\
    ambiguous-translation-wrong-discourse-connective-while-temporal & \hyperref[p:span_diff_flexible]{word-lvl-compare-to-good} \\
    ambiguous-translation-wrong-gender-female-anti & \hyperref[p:span_diff_flexible]{word-lvl-compare-to-good} \\
    ambiguous-translation-wrong-gender-female-pro & \hyperref[p:span_diff_flexible]{word-lvl-compare-to-good} \\
    ambiguous-translation-wrong-gender-male-anti & \hyperref[p:span_diff_flexible]{word-lvl-compare-to-good} \\
    ambiguous-translation-wrong-gender-male-pro & \hyperref[p:span_diff_flexible]{word-lvl-compare-to-good} \\
    ambiguous-translation-wrong-sense-frequent & \hyperref[p:span_diff_flexible]{word-lvl-compare-to-good} \\
    ambiguous-translation-wrong-sense-infrequent & \hyperref[p:span_diff_flexible]{word-lvl-compare-to-good} \\
    anaphoric\_group\_it-they:deletion & \hyperref[p:span_annotate_word]{addition/omissions} \\
    anaphoric\_group\_it-they:substitution & \hyperref[p:span_annotate_word]{addition/omissions} \\
    anaphoric\_intra\_non-subject\_it:deletion & \hyperref[p:span_annotate_word]{addition/omissions} \\
    anaphoric\_intra\_non-subject\_it:substitution &\hyperref[p:span_annotate_word]{addition/omissions}  \\
    anaphoric\_intra\_subject\_it:deletion & \hyperref[p:span_annotate_word]{addition/omissions} \\
    anaphoric\_intra\_subject\_it:substitution & \hyperref[p:span_annotate_word]{addition/omissions} \\
    anaphoric\_intra\_they:deletion & \hyperref[p:span_annotate_word]{addition/omissions} \\
    anaphoric\_intra\_they:substitution & \hyperref[p:span_annotate_word]{addition/omissions} \\
    anaphoric\_singular\_they:deletion & \hyperref[p:span_annotate_word]{addition/omissions} \\
    anaphoric\_singular\_they:substitution & \hyperref[p:span_annotate_word]{addition/omissions} \\
    antonym-replacement & \hyperref[p:span_REF_flexible]{word-lvl-compare-to-ref} \\
    commonsense-only-ref-ambiguous & \hyperref[p:span_diff_flexible]{word-lvl-compare-to-good} \\
    commonsense-src-and-ref-ambiguous & \hyperref[p:span_diff_flexible]{word-lvl-compare-to-good} \\
    copy-source & \hyperref[p:span_whole_sentence]{whole-sentence} \\
    coreference-based-on-commonsense & \hyperref[subsec:span_manual_annotation]{manual} \\
    do-not-translate & \hyperref[p:span_diff_flexible]{word-lvl-compare-to-good} \\
    hallucination-date-time & \hyperref[p:span_date]{date-time} \\
    hallucination-named-entity-level-1 & \hyperref[p:span_diff_flexible]{word-lvl-compare-to-good} \\
    hallucination-named-entity-level-2 & \hyperref[p:span_REF_flexible]{word-lvl-compare-to-ref} \\
    hallucination-named-entity-level-3 & \hyperref[p:span_REF_flexible]{word-lvl-compare-to-ref} \\
    hallucination-number-level-1 & \hyperref[p:span_diff_flexible]{word-lvl-compare-to-good} \\
    hallucination-number-level-2 & \hyperref[p:span_REF_flexible]{word-lvl-compare-to-ref} \\
    hallucination-number-level-3 & \hyperref[p:span_REF_flexible]{word-lvl-compare-to-ref} \\
    hallucination-real-data-vs-ref-word & \hyperref[subsec:span_manual_annotation]{manual} \\
    hallucination-real-data-vs-synonym & \hyperref[subsec:span_manual_annotation]{manual} \\
    hallucination-unit-conversion-amount-matches-ref & \hyperref[p:span_units]{unit-conversion} \\
    hallucination-unit-conversion-unit-matches-ref & \hyperref[p:span_units]{unit-conversion} \\
    hypernym-replacement & \hyperref[p:span_REF_flexible]{word-lvl-compare-to-ref} \\
    hyponym-replacement & \hyperref[p:span_REF_flexible]{word-lvl-compare-to-ref} \\
    lexical-overlap & \hyperref[subsec:span_manual_annotation]{manual} \\
    modal\_verb:deletion & \hyperref[p:span_annotate_word]{addition/omissions} \\
    modal\_verb:substitution & \hyperref[p:span_diff_flexible]{word-lvl-compare-to-good} \\
    nonsense & \hyperref[p:span_REF_flexible]{word-lvl-compare-to-ref} \\
    omission & \hyperref[p:span_annotate_word]{addition/omissions} \\
    ordering-mismatch & \hyperref[p:span_swap]{word-swap} \\
    overly-literal-vs-correct-idiom & \hyperref[p:span_diff_flexible]{word-lvl-compare-to-good} \\
    overly-literal-vs-explanation & \hyperref[p:span_diff_flexible]{word-lvl-compare-to-good} \\
    overly-literal-vs-ref-word & \hyperref[p:span_diff_flexible]{word-lvl-compare-to-good} \\
    overly-literal-vs-synonym & \hyperref[p:span_diff_flexible]{word-lvl-compare-to-good} \\
    pleonastic\_it:deletion & \hyperref[p:span_annotate_word]{addition/omissions} \\
    pleonastic\_it:substitution & \hyperref[p:span_annotate_word]{addition/omissions} \\
    punctuation:deletion\_all & \hyperref[p:span_annotate_word]{addition/omissions} \\
    punctuation:deletion\_commas & \hyperref[p:span_annotate_word]{addition/omissions} \\
    punctuation:deletion\_quotes & \hyperref[p:span_annotate_word]{addition/omissions} \\
    punctuation:statement-to-question & \hyperref[p:span_annotate_word]{addition/omissions} \\
    real-world-knowledge-entailment & \hyperref[p:span_diff_flexible]{word-lvl-compare-to-good} \\
    real-world-knowledge-hypernym-vs-distractor & \hyperref[p:span_diff_flexible]{word-lvl-compare-to-good} \\
    real-world-knowledge-hypernym-vs-hyponym & \hyperref[p:span_diff_flexible]{word-lvl-compare-to-good} \\
    real-world-knowledge-synonym-vs-antonym & \hyperref[p:span_diff_flexible]{word-lvl-compare-to-good} \\
    similar-language-high & \hyperref[p:span_whole_sentence]{whole-sentence} \\
    similar-language-low & \hyperref[p:span_whole_sentence]{whole-sentence} \\
    untranslated-vs-ref-word & \hyperref[p:span_diff_flexible]{word-lvl-compare-to-good} \\ 
    untranslated-vs-synonym & \hyperref[p:span_diff_flexible]{word-lvl-compare-to-good} \\
    xnli-addition-contradiction & \hyperref[p:span_whole_sentence]{whole-sentence} \\
    xnli-addition-neutral & \hyperref[p:span_whole_sentence]{whole-sentence} \\
    xnli-omission-contradiction & \hyperref[p:span_whole_sentence]{whole-sentence} \\
    xnli-omission-neutral & \hyperref[p:span_whole_sentence]{whole-sentence} \\

\end{tabular}
\caption{Methods used to annotate the error spans for each of the phenomena in \textsc{span-ACES}}
\label{tab:annotation_methods_for_phenomena}
\end{center}
\end{table*}

The methods used to annotate the error spans for each of the phenomena in \textsc{span-ACES} are listed in Table~\ref{tab:annotation_methods_for_phenomena}. 

\appendixsection{ACES Span Annotation Guidelines}
\label{app:annotation_guidelines}

\noindent\textbf{1. General guidelines}\\

\noindent Your task is to annotate spans of translation errors that match a specific error type: e.g. “word swap”, or ``overtranslation''. You are presented with two sentences (A and B) as well as a label denoting the error type that you should look for. You should compare translations A and B and mark any error spans of the specified type that occur in sentence B.\\

\noindent Please note that:
\begin{itemize}
    \setlength\itemsep{0.25em}
    \item You should annotate at the word level, not at the character level. I.e. in the case that the error is a misspelling (e.g. ``combuter'' instead of ``computer'') the complete word (``combuter'') should be marked.
    \item You should \textbf{\textit{only}} mark errors of the type specified by the error type label, and no other errors that may be present in sentence B.
    \item You are \textbf{\textit{not}} required to mark any errors that may be present in sentence A.
    \item Whilst the majority of sentences you will encounter will be fluent, some machine-generated sentences will contain disfluencies.
    \item In the examples in this document, errors are highlighted in bold text to help make the examples clearer. You do \textbf{\textit{not}} need to bold the error spans in your annotations.
    \item This document is intended to be comprehensive and cover the cases assigned across multiple annotators. As such, a batch that is assigned to you may contain only a subset of the error types listed in the \textit{Error type-specific } section (below).
    \item You should only mark punctuation as part of error spans if it is part of the error (e.g. added as part of an addition operation or changed as part of a substitution operation).
\end{itemize}

\noindent Please read the guidelines thoroughly before you start the annotation task. Once you have finished, please make a second pass to identify and correct any mistakes that you may have made. Please also make a note of any examples that you were unsure how to annotate e.g. the example ID and a brief note.\\

\noindent All error spans should be marked with open and closing tags (e.g. <error span>). Errors of specific types may be formed by addition, substitution, deletion or reordering operations. For deletion operations, you should insert an empty pair of tags <> where content is missing in sentence B.\\

\noindent\textbf{Whitespace:} Error tags should \textbf{\textit{not}} contain leading (e.g. \textbf{< error span>}) or trailing (e.g. \textbf{<error span >}) whitespace.\\

\noindent\textbf{Addition:} a text span that is not present in sentence A is included in sentence B.

\begin{quote}
Sentence A: The cat is a species of small carnivorous mammal.\\
Sentence B: The cat is a \textbf{<domestic>} species of small carnivorous mammal.
\end{quote}

\noindent\textbf{Substitution:} a text span in sentence A is substituted with a different text span in sentence B.

\begin{quote}
Sentence A: Female domestic cats can have kittens from spring to late autumn.\\
Sentence B: Female domestic cats can have kittens from \textbf{<May>} to \textbf{<December>}.
\end{quote}

\noindent\textbf{Deletion:} a text span that is present in sentence A is omitted from sentence B. Note that when marking a deletion, care should be taken to ensure that no extra whitespace is inserted into the sentence. Tags marking the deletion should be inserted after the space separating the two words where the deletion occurred.

\begin{quote}
Sentence A: Feral cats are domestic cats that were born in or have reverted to a wild state.\\
Sentence B: Feral cats are domestic cats \textbf{<>}or have reverted to a wild state.
\end{quote}

\noindent\textbf{Reordering:} a text span in sentence A that appears in a different position in sentence B, as though the sentence has been reordered.

\begin{quote}
Sentence A: Montreal is the second most populous city in Canada and the most populous city in the province of Quebec.\\
Sentence B: Montreal is the \textbf{<>}most populous city in Canada and the \textbf{<second>} most populous city in the province of Quebec.
\end{quote}

\noindent Note: reordering operations can be viewed as a combination of a \textit{deletion} and an \textit{addition} operation to change the order of elements of a sentence.\\

\noindent\textbf{Example 1:} Marking a single error span of a specified error type; ignoring other error types\\

\noindent In this example, the aim is to mark ``overtranslation'' type errors, i.e. where translation B is more specific than translation A:

\begin{quote}
Sentence A: The festival in Houston took place in the summer.\\
Sentence B: The festival in took place in August.
\end{quote}

\noindent The error span is ``August'', which is more specific than ``the summer'' - the information that the event took place in August has been ``hallucinated''.

\begin{quote}
Annotated B: The Republican National Convention in was in \textbf{<August>}.
\end{quote}

\noindent Note that the missing information in sentence B (``Houston'') can be ignored because it is an ``omission'' error not an ``overtranslation'' error. Other examples of errors that can be ignored include e.g. agreement errors in German.\\

\noindent\textbf{Example 2:} Marking multiple error spans in the same example\\

\noindent If there are multiple errors of the specified type present in sentence B, you should mark each error span individually. For example, if the error label is ``omission'' you should mark the two spans of omitted text in sentence B:

\begin{quote}
Sentence A: Like the other planets in the Solar System, Mars was formed 4.5 billion years ago.\\
Sentence B: Like the other planets, Mars was formed 4.5 years ago.
\end{quote}

\begin{quote}
Annotated B: Like the other planets \textbf{<>}, Mars was formed 4.5 \textbf{<>}years ago.
\end{quote}

\noindent\textbf{2. Error type-specific guidelines}\\

\noindent In your annotations, you will only encounter three specific error types. Additional guidelines are provided below for these error types - hallucination, word swap and coreference.\\

\noindent\textbf{Hallucination}\\

\noindent In a \textit{hallucination} example, text that is not present in sentence A is observed in sentence B or word in sentence A is replaced by a more frequent or orthographically similar word in sentence B. I.e. hallucination can be an ``addition'' or a ``substitution'' case. This may result in a change of meaning in sentence B. You should mark the ``hallucinated'' text in sentence B.

\begin{quote}
Sentence A: The official languages of Scotland are: English, Scots, and Scottish Gaelic.\\
Sentence B: The official languages of Scotland are: English, Welsh, French, Scots, and Scottish Garlic.
\end{quote}

The information that Welsh and French are official languages of Scotland has been hallucinated and inserted into sentence B. Additionally, ``Gaelic'' has been hallucinated as ``Garlic''. This should be annotated as:

\begin{quote}
Annotated B: The official languages of Scotland are: English, \textbf{<Welsh, French,>} Scots, and Scottish \textbf{<Garlic>}.
\end{quote}

\noindent\textbf{Word Swap}\\

\noindent In a \textit{word swap} example the position of a word or a span of text in sentence A appears swapped in sentence B. This may result in sentence B being factually incorrect. You should mark (in sentence B) the spans of text that have been swapped.

\begin{quote}
Sentence A: Their music is considered by many as an alternative metal with rap metal and industrial metal influences, which according to previous interviews call themselves ``murder - rock''.\\
Sentence B: Their music is considered by many as industrial metal with rap metal and alternative metal influences. According to previous interviews, they consider themselves ``murder rock''.
\end{quote}

\noindent The position of the words ``alternative'' and ``industrial'' is different in sentence A, compared with sentence B and should be annotated as follows:

\begin{quote}
Annotated B: Their music is considered by many as \textbf{<industrial>} metal with rap metal and \textbf{<alternative>} metal influences. According to previous interviews, they consider themselves ``murder rock''.
\end{quote}

\noindent\textbf{Coreference}\\

\noindent In a \textit{coreference} example a pronoun in sentence A is replaced with a (potentially) inappropriate noun-phrase in sentence B. You should mark the relevant noun-phrase in sentence B.\\

\noindent Example:

\begin{quote}
Sentence A: The cat had caught the mouse and it was trying to wriggle free.\\
Sentence B: The cat had caught the mouse and the cat was trying to wriggle free.
\end{quote}

\noindent The pronoun ``it'' has been replaced with the noun-phrase ``the cat'', resulting in a change in meaning. This should be annotated as:

\begin{quote}
Annotated B: The cat had caught the mouse and \textbf{<the cat>} was trying to wriggle free.
\end{quote}

\appendixsection{Prompt for LLMs for MT evaluation}
For reference-based evaluation, we used the following prompt:\\\
\noindent Score the following translation with respect to human reference on a continuous scale of 0 to 100 where score of zero means ``no meaning preserved'' and score of one hundred means ``perfect meaning and grammar''. Only output an integer between 0 to 100.\\
\noindent Source: \textit{source sentence here} \\
\noindent    Human Reference: \textit{reference sentence here} \\
\noindent    Translation: \textit{candidate translation} \\

For reference-free evaluation, we excluded the ``with respect to human reference'' and 
``Human Reference'' from the prompt.

\appendixsection{Importance of source}
We report the results on the \hyperref[sec:real-world-knowledge]{real-world knowledge commonsense challenge set} in Table ~\ref{tab:corr_gain}. Reference-based metrics tend to disregard the information in the source.


\begin{table}[]
\centering
\begin{tabular}{@{}lr|lr@{}}
\toprule
Reference-based & \multicolumn{1}{l}{corr-gain} & Reference-free & \multicolumn{1}{l}{corr-gain} \\ \toprule
BERTScore &\phantom{-}0.002 & COMET-QE &\phantom{-}0.018 \\
COMET-20 &\phantom{-}0.06 & Cross-QE &\phantom{-}0.292 \\
COMET-22 &\phantom{-}0.19 & HWTSC-Teacher-Sim &\phantom{-}0.154 \\
metricx\_xxl\_DA\_2019 &\phantom{-}0.012 & KG-BERTScore &\phantom{-}0.154 \\
metricx\_xxl\_MQM\_2020 & -0.016 & MS-COMET-QE-22 &\phantom{-}0.196 \\
MS-COMET-22 &\phantom{-}0.05 & UniTE-src &\phantom{-}0.216 \\
UniTE &\phantom{-}0.042 & cometoid22-wmt23 &\phantom{-}0.138 \\
COMET-22 &\phantom{-}0.042 & CometKiwi &\phantom{-}0.454 \\
MetricX-23 &\phantom{-}0.004 & CometKiwi-XL &\phantom{-}0.148 \\
MetricX-23-b & -0.002 & GEMBA-MQM & 1.107 \\
MetricX-23-c &\phantom{-}0.008 & KG-BERTScore &\phantom{-}0.436 \\
XCOMET-Ensemble &\phantom{-}0.162 & MS-COMET-QE-22 &\phantom{-}0.198 \\
XCOMET-XL &\phantom{-}0.11 & MetricX-23-QE-b &\phantom{-}0.296 \\
XCOMET-XXL &\phantom{-}0.016 & XCOMET-QE-Ensemble &\phantom{-}0.112 \\
 & \multicolumn{1}{l}{} & XLsimQE &\phantom{-}0.184 \\ \bottomrule
\end{tabular}
\caption{Results on the \hyperref[sec:real-world-knowledge]{real-world knowledge commonsense challenge set} with reference-based metrics in the left block and reference-free metrics in the right block. The numbers are computed as the difference between the correlation with the subordinate clause in the source and the correlation without the subordinate clause in the source. Largest gains are bolded.}
\label{tab:corr_gain}
\end{table}

\appendixsection{Phenomena-level Metric Sensitivity Scores}
\label{app:phenomena-level-sensitivities}
\begin{sidewaystable*}[ht]
 \small 
 \setlength{\tabcolsep}{3.75pt} 
 \centering 
  \resizebox{\textwidth}{!}{

 \begin{tabular}{@{}lcccccccccc@{}} 
 \\\toprule 
 & \hyperref[sec:addition-omission]{\textbf{addition}} & \hyperref[sec:addition-omission]{\textbf{omission}} & \hyperref[sec:source-disambig]{\textbf{mistranslation}} & \hyperref[sec:untranslated]{\textbf{untranslated}} & \hyperref[sec:do-not-translate]{\textbf{do not}} & \hyperref[sec:overtranslation_undertranslation]{\textbf{overtranslation}} & \hyperref[sec:overtranslation_undertranslation]{\textbf{undertranslation}} & \hyperref[sec:real-world-knowledge]{\textbf{real-world}} & \hyperref[sec:wrong_language]{\textbf{wrong}} & \hyperref[sec:punctuation]{\textbf{punctuation}} \\
 &  &  &  &  & \hyperref[sec:do-not-translate]{\textbf{translate}} &  &  & \hyperref[sec:real-world-knowledge]{\textbf{knowledge}} & \hyperref[sec:wrong_language]{\textbf{language}} & \\
\midrule
\textit{\textbf{Examples}}  & \textit{931} & \textit{951} & \textit{22530} & \textit{1187} & \textit{76} & \textit{962} & \textit{967} & \textit{2924} & \textit{1840} & \textit{1449}\\ 
 \midrule
BLEU					&	\colorbox[HTML]{B2EAB1}{\textbf{\phantom{-}0.222}}	&	\phantom{-}0.253	&	-0.078	&	\phantom{-}1.213	&	\phantom{-}0.314	&	-1.093	&	-1.096	&	-0.293	&	\phantom{-}0.655	&	\phantom{-}0.365 \\ 
f101spBLEU					&	\phantom{-}0.136	&	\phantom{-}0.186	&	-0.071	&	\phantom{-}0.911	&	\phantom{-}0.348	&	-0.563	&	-0.608	&	-0.160	&	\phantom{-}0.503	&	\phantom{-}0.211	 \\ 
f200spBLEU					&	\phantom{-}0.131	&	\phantom{-}0.181	&	-0.072	&	\phantom{-}0.892	&	\phantom{-}0.356	&	-0.529	&	-0.553	&	-0.159	&	\phantom{-}0.496	&	\phantom{-}0.192	 \\ 
chrF					&	\phantom{-}0.061	&	\phantom{-}0.253	&	-0.073	&	\phantom{-}1.535	&	\phantom{-}0.289	&	-0.286	&	-0.222	&	-0.087	&	\phantom{-}0.656	&	\phantom{-}0.107	\\ 
BERTScore					&	\phantom{-}0.115	&	\phantom{-}0.182	&	-0.018	&	\phantom{-}1.373	&	\phantom{-}0.304	&	-0.018	&	-0.037	&	-0.026	&	\phantom{-}0.375	&	\colorbox[HTML]{B2EAB1}{\textbf{\phantom{-}0.493}}	 \\ 
BLEURT-20					&	\phantom{-}0.106	&	\phantom{-}0.355	&	\phantom{-}0.200	&	\phantom{-}1.743	&	\phantom{-}0.398	&	\phantom{-}0.142	&	-0.002	&	\phantom{-}0.055	&	\colorbox[HTML]{B2EAB1}{\textbf{\phantom{-}0.826}}	&	\phantom{-}0.318	\\ 
COMET-20					&	\phantom{-}0.073	&	\phantom{-}0.410	&	\phantom{-}0.262	&	\phantom{-}1.486	&	\phantom{-}0.312	&	\phantom{-}0.150	&	\phantom{-}0.061	&	\phantom{-}0.051	&	-0.322	&	\phantom{-}0.229	 \\ 
COMET-QE					&	-0.103	&	\phantom{-}0.117	&	\phantom{-}0.072	&	-0.126	&	\phantom{-}0.049	&	\phantom{-}0.286	&	\phantom{-}0.196	&	\phantom{-}0.096	&	-0.310	&	\phantom{-}0.026		 \\ 
YISI-1					&	\phantom{-}0.118	&	\phantom{-}0.293	&	\phantom{-}0.075	&	\colorbox[HTML]{B2EAB1}{\textbf{\phantom{-}2.575}}	&	\phantom{-}0.294	&	-0.036	&	-0.049	&	-0.044	&	\phantom{-}0.190	&	\phantom{-}0.376 \\ 
\midrule 
COMET-22					&	\phantom{-}0.045	&	\phantom{-}0.250	&	\phantom{-}0.292	&	\phantom{-}0.399	&	\phantom{-}0.318	&	\phantom{-}0.352	&	\phantom{-}0.216	&	\phantom{-}0.207	&	-0.497	&	\phantom{-}0.217	\\ 
metricx\_xl\_DA\_2019					&	\phantom{-}0.100	&	\phantom{-}0.447	&	\phantom{-}0.496	&	\phantom{-}1.772	&	\phantom{-}0.559	&	\phantom{-}0.689	&	\phantom{-}0.366	&	\phantom{-}0.498	&	\phantom{-}0.752	&	\phantom{-}0.302 \\ 
metricx\_xl\_MQM\_2020					&	-0.056	&	\phantom{-}0.361	&	\colorbox[HTML]{B2EAB1}{\textbf{\phantom{-}0.651}}	&	\phantom{-}0.422	&	\colorbox[HTML]{B2EAB1}{\textbf{\phantom{-}0.697}}	&	\colorbox[HTML]{B2EAB1}{\textbf{\phantom{-}1.000}}	&	\colorbox[HTML]{B2EAB1}{\textbf{\phantom{-}0.654}}	&	\colorbox[HTML]{B2EAB1}{\textbf{\phantom{-}0.740}}	&	-0.560	&	\phantom{-}0.331	\\ 
metricx\_xxl\_DA\_2019					&	\phantom{-}0.085	&	\phantom{-}0.411	&	\phantom{-}0.532	&	\phantom{-}1.679	&	\phantom{-}0.543	&	\phantom{-}0.547	&	\phantom{-}0.189	&	\phantom{-}0.423	&	\phantom{-}0.088	&	\phantom{-}0.244 \\ 
metricx\_xxl\_MQM\_2020					&	-0.008	&	\phantom{-}0.294	&	\phantom{-}0.550	&	\phantom{-}0.649	&	\phantom{-}0.688	&	\phantom{-}0.826	&	\phantom{-}0.485	&	\phantom{-}0.629	&	-0.768	&	\phantom{-}0.225	\\ 
MS-COMET-22					&	-0.048	&	\phantom{-}0.351	&	\phantom{-}0.133	&	\phantom{-}0.987	&	\phantom{-}0.183	&	\phantom{-}0.222	&	\phantom{-}0.119	&	\phantom{-}0.059	&	-0.159	&	\phantom{-}0.192	 \\ 
UniTE					&	\phantom{-}0.113	&	\phantom{-}0.420	&	\phantom{-}0.321	&	\phantom{-}0.534	&	\phantom{-}0.353	&	\phantom{-}0.338	&	\phantom{-}0.152	&	\phantom{-}0.189	&	-0.439	&	\phantom{-}0.204	\\ 
UniTE-ref					&	\phantom{-}0.078	&	\phantom{-}0.385	&	\phantom{-}0.304	&	\phantom{-}0.081	&	\phantom{-}0.294	&	\phantom{-}0.378	&	\phantom{-}0.182	&	\phantom{-}0.187	&	-0.436	&	\phantom{-}0.168	 \\ 
\midrule 
COMETKiwi					&	\phantom{-}0.126	&	\phantom{-}0.441	&	\phantom{-}0.594	&	\phantom{-}0.699	&	\phantom{-}0.272	&	\phantom{-}0.572	&	\phantom{-}0.358	&	\phantom{-}0.337	&	-0.559	&	\phantom{-}0.247	 \\ 
Cross-QE					&	\phantom{-}0.104	&	\phantom{-}0.422	&	\phantom{-}0.599	&	-0.285	&	\phantom{-}0.055	&	\phantom{-}0.703	&	\phantom{-}0.456	&	\phantom{-}0.225	&	-0.510	&	\phantom{-}0.109  \\ 
HWTSC-Teacher-Sim					&	-0.005	&	\phantom{-}0.140	&	\phantom{-}0.163	&	-0.574	&	\phantom{-}0.200	&	\phantom{-}0.224	&	\phantom{-}0.174	&	\phantom{-}0.055	&	-0.063	&	\phantom{-}0.239	\\ 
HWTSC-TLM					&	-0.095	&	\phantom{-}0.148	&	\phantom{-}0.120	&	\phantom{-}0.104	&	-0.019	&	\phantom{-}0.293	&	\phantom{-}0.241	&	\phantom{-}0.062	&	-0.158	&	\phantom{-}0.463	\\ 
KG-BERTScore					&	\phantom{-}0.139	&	\phantom{-}0.236	&	\phantom{-}0.428	&	-0.786	&	\phantom{-}0.598	&	\phantom{-}0.186	&	\phantom{-}0.124	&	\phantom{-}0.079	&	-0.170	&	\phantom{-}0.366	\\ 
MS-COMET-QE-22					&	-0.038	&	\phantom{-}0.369	&	\phantom{-}0.243	&	\phantom{-}2.564	&	\phantom{-}0.119	&	\phantom{-}0.241	&	\phantom{-}0.174	&	\phantom{-}0.090	&	-0.616	&	\phantom{-}0.212 \\ 
UniTE-src					&	\phantom{-}0.096	&	\colorbox[HTML]{B2EAB1}{\textbf{\phantom{-}0.524}}	&	\phantom{-}0.484	&	-0.782	&	\phantom{-}0.318	&	\phantom{-}0.394	&	\phantom{-}0.254	&	\phantom{-}0.191	&	-0.552	&	\phantom{-}0.196 \\ 
\midrule 
Average					&	\phantom{-}0.062	&	\phantom{-}0.309	&	\phantom{-}0.259	&	\phantom{-}0.794	&	\phantom{-}0.327	&	\phantom{-}0.209	&	\phantom{-}0.076	&	\phantom{-}0.142	&	-0.066	&	\phantom{-}0.251	\\

 \bottomrule
\end{tabular}}
\caption{Metric sensitivity scores (scaled by WMT scores, then avg(good - bad)) of metrics submitted to WMT 2022 for the nine top level categories in the \textsc{ACES} ontology, plus the additional fluency category: punctuation.   The horizontal lines delimit baseline metrics (top), participating reference-based metrics (middle) and participating reference-free metrics (bottom). The best result for each category is denoted by bold text with a green highlight. Note that \textit{Average} is an average over averages.}
\label{tab:analysis_overview_sensitivities_2022}
\end{sidewaystable*}
\afterpage{\clearpage}
\begin{sidewaystable*}[ht]
 \small 
 \setlength{\tabcolsep}{3.75pt} 
 \centering 
  \resizebox{\textwidth}{!}{
 \begin{tabular}{@{}lcccccccccc@{}} 
 \\\toprule 
 & \hyperref[sec:addition-omission]{\textbf{addition}} & \hyperref[sec:addition-omission]{\textbf{omission}} & \hyperref[sec:source-disambig]{\textbf{mistranslation}} & \hyperref[sec:untranslated]{\textbf{untranslated}} & \hyperref[sec:do-not-translate]{\textbf{do not}} & \hyperref[sec:overtranslation_undertranslation]{\textbf{overtranslation}} & \hyperref[sec:overtranslation_undertranslation]{\textbf{undertranslation}} & \hyperref[sec:real-world-knowledge]{\textbf{real-world}} & \hyperref[sec:wrong_language]{\textbf{wrong}} & \hyperref[sec:punctuation]{\textbf{punctuation}}  \\
 &  &  &  &  & \hyperref[sec:do-not-translate]{\textbf{translate}} &  &  & \hyperref[sec:real-world-knowledge]{\textbf{knowledge}} & \hyperref[sec:wrong_language]{\textbf{language}} &  \\
\midrule
\textit{\textbf{Examples}}  & \textit{931} & \textit{951} & \textit{22530} & \textit{1187} & \textit{76} & \textit{962} & \textit{967} & \textit{2924} & \textit{1840} & \textit{1449}\\ 
 \midrule
BERTScore					&	\phantom{-}0.119	&	\phantom{-}0.189	&	-0.010	&	\phantom{-}1.410	&	\phantom{-}0.311	&	-0.047	&	-0.066	&	-0.031	&	\phantom{-}0.384	&	\colorbox[HTML]{B2EAB1}{\textbf{\phantom{-}0.505}}	\\ 
BLEU					&	\phantom{-}0.131	&	\phantom{-}0.149	&	-0.046	&	\phantom{-}0.716	&	\phantom{-}0.185	&	-0.646	&	-0.647	&	-0.173	&	\phantom{-}0.387	&	\phantom{-}0.216	\\ 
BLEURT-20					&	\phantom{-}0.094	&	\phantom{-}0.314	&	\phantom{-}0.177	&	\phantom{-}1.545	&	\phantom{-}0.353	&	\phantom{-}0.126	&	-0.002	&	\phantom{-}0.048	&	\phantom{-}0.732	&	\phantom{-}0.281	 \\ 
chrF					&	\phantom{-}0.053	&	\phantom{-}0.220	&	-0.064	&	\phantom{-}1.338	&	\phantom{-}0.252	&	-0.249	&	-0.194	&	-0.076	&	\phantom{-}0.572	&	\phantom{-}0.094 \\ 
COMET-22					&	\phantom{-}0.036	&	\phantom{-}0.320	&	\phantom{-}0.199	&	\phantom{-}0.846	&	\phantom{-}0.272	&	\phantom{-}0.250	&	\phantom{-}0.127	&	\phantom{-}0.107	&	-0.213	&	\phantom{-}0.196	\\ 
COMETKiwi					&	\phantom{-}0.196	&	\phantom{-}0.618	&	\phantom{-}0.721	&	-0.180	&	\phantom{-}0.285	&	\phantom{-}0.719	&	\phantom{-}0.439	&	\phantom{-}0.316	&	-0.767	&	\phantom{-}0.218	 \\ 
f200spBLEU					&	\phantom{-}0.121	&	\phantom{-}0.167	&	-0.066	&	\phantom{-}0.824	&	\phantom{-}0.329	&	-0.489	&	-0.511	&	-0.147	&	\phantom{-}0.458	&	\phantom{-}0.177	 \\ 
MS-COMET-QE-22					&	-0.040	&	\phantom{-}0.391	&	\phantom{-}0.258	&	\colorbox[HTML]{B2EAB1}{\textbf{\phantom{-}2.730}}	&	\phantom{-}0.126	&	\phantom{-}0.257	&	\phantom{-}0.185	&	\phantom{-}0.095	&	-0.654	&	\phantom{-}0.226	\\ 
Random-sysname					&	-0.003	&	\phantom{-}0.016	&	\phantom{-}0.003	&	\phantom{-}0.013	&	\phantom{-}0.015	&	\phantom{-}0.023	&	-0.004	&	-0.008	&	\phantom{-}0.003	&	-0.013	\\ 
YISI-1					&	\phantom{-}0.114	&	\phantom{-}0.283	&	\phantom{-}0.072	&	\phantom{-}2.489	&	\phantom{-}0.284	&	-0.034	&	-0.048	&	-0.043	&	\phantom{-}0.183	&	\phantom{-}0.365	\\ 
\midrule 
eBLEU					&	\phantom{-}0.070	&	\phantom{-}0.135	&	\phantom{-}0.014	&	\phantom{-}1.358	&	\phantom{-}0.169	&	-0.199	&	-0.194	&	-0.068	&	\colorbox[HTML]{B2EAB1}{\textbf{\phantom{-}0.986}}	&	\phantom{-}0.065	 \\ 
embed\_llama					&	\phantom{-}0.190	&	\phantom{-}0.324	&	\phantom{-}0.027	&	\phantom{-}0.962	&	\phantom{-}0.262	&	-0.217	&	-0.851	&	-0.360	&	\phantom{-}0.139	&	\phantom{-}0.130	 \\ 
MetricX-23					&	\phantom{-}0.001	&	\phantom{-}0.184	&	\phantom{-}0.407	&	-0.022	&	\phantom{-}0.363	&	\phantom{-}0.675	&	\phantom{-}0.367	&	\phantom{-}0.491	&	-0.618	&	\phantom{-}0.151	\\ 
MetricX-23-b					&	-0.029	&	\phantom{-}0.231	&	\phantom{-}0.375	&	\phantom{-}0.135	&	\phantom{-}0.385	&	\phantom{-}0.601	&	\phantom{-}0.336	&	\phantom{-}0.460	&	-0.696	&	\phantom{-}0.140	\\ 
MetricX-23-c					&	\phantom{-}0.021	&	\phantom{-}0.413	&	\phantom{-}0.645	&	\phantom{-}0.334	&	\phantom{-}0.399	&	\phantom{-}0.593	&	\phantom{-}0.330	&	\phantom{-}0.766	&	-0.728	&	\phantom{-}0.131	\\ 
tokengram\_F					&	\phantom{-}0.054	&	\phantom{-}0.214	&	-0.058	&	\phantom{-}1.335	&	\phantom{-}0.280	&	-0.281	&	-0.231	&	-0.080	&	\phantom{-}0.603	&	\phantom{-}0.208	 \\ 
XCOMET-Ensemble					&	\phantom{-}0.070	&	\phantom{-}0.342	&	\phantom{-}0.434	&	\phantom{-}0.208	&	\phantom{-}0.249	&	\phantom{-}0.462	&	\phantom{-}0.308	&	\phantom{-}0.358	&	-0.713	&	\phantom{-}0.151 \\ 
XCOMET-XL					&	\phantom{-}0.047	&	\phantom{-}0.244	&	\phantom{-}0.303	&	-0.125	&	\phantom{-}0.227	&	\phantom{-}0.310	&	\phantom{-}0.198	&	\phantom{-}0.288	&	-0.650	&	\phantom{-}0.075	\\ 
XCOMET-XXL					&	\phantom{-}0.057	&	\phantom{-}0.260	&	\phantom{-}0.349	&	-0.197	&	\phantom{-}0.210	&	\phantom{-}0.476	&	\phantom{-}0.375	&	\phantom{-}0.298	&	-0.690	&	\phantom{-}0.119	\\ 
XLsim					&	\phantom{-}0.085	&	\phantom{-}0.220	&	\phantom{-}0.126	&	\phantom{-}1.442	&	\phantom{-}0.372	&	-0.129	&	-0.158	&	-0.063	&	\phantom{-}0.192	&	\phantom{-}0.365	\\ 
\midrule 
cometoid22-wmt21					&	-0.063	&	\phantom{-}0.233	&	\phantom{-}0.207	&	-0.513	&	\phantom{-}0.138	&	\phantom{-}0.399	&	\phantom{-}0.260	&	\phantom{-}0.147	&	-0.528	&	\phantom{-}0.215	 \\ 
cometoid22-wmt22					&	-0.058	&	\phantom{-}0.243	&	\phantom{-}0.221	&	-0.567	&	\phantom{-}0.134	&	\phantom{-}0.396	&	\phantom{-}0.260	&	\phantom{-}0.148	&	-0.545	&	\phantom{-}0.205	\\ 
cometoid22-wmt23					&	-0.044	&	\phantom{-}0.270	&	\phantom{-}0.239	&	-0.479	&	\phantom{-}0.155	&	\phantom{-}0.304	&	\phantom{-}0.198	&	\phantom{-}0.106	&	-0.600	&	\phantom{-}0.198	\\ 
CometKiwi-XL					&	\phantom{-}0.094	&	\phantom{-}0.581	&	\phantom{-}0.703	&	\phantom{-}2.047	&	\phantom{-}0.264	&	\phantom{-}0.459	&	\phantom{-}0.284	&	\phantom{-}0.486	&	-0.614	&	\phantom{-}0.197	\\ 
CometKiwi-XXL					&	\phantom{-}0.118	&	\phantom{-}0.519	&	\phantom{-}0.713	&	\phantom{-}2.038	&	\phantom{-}0.197	&	\phantom{-}0.550	&	\phantom{-}0.306	&	\phantom{-}0.611	&	-0.733	&	\phantom{-}0.187	\\ 
GEMBA-MQM					&	\colorbox[HTML]{B2EAB1}{\textbf{\phantom{-}0.584}}	&	\colorbox[HTML]{B2EAB1}{\textbf{\phantom{-}1.132}}	&	\colorbox[HTML]{B2EAB1}{\textbf{\phantom{-}1.566}}	&	\phantom{-}2.380	&	\colorbox[HTML]{B2EAB1}{\textbf{\phantom{-}0.719}}	&	\colorbox[HTML]{B2EAB1}{\textbf{\phantom{-}1.646}}	&	\colorbox[HTML]{B2EAB1}{\textbf{\phantom{-}0.976}}	&	\colorbox[HTML]{B2EAB1}{\textbf{\phantom{-}1.814}}	&	\phantom{-}0.328	&	\phantom{-}0.226	 \\ 
KG-BERTScore					&	\phantom{-}0.175	&	\phantom{-}0.550	&	\phantom{-}0.638	&	-0.181	&	\phantom{-}0.341	&	\phantom{-}0.639	&	\phantom{-}0.388	&	\phantom{-}0.277	&	-0.685	&	\phantom{-}0.173 \\ 
MetricX-23-QE					&	\phantom{-}0.001	&	\phantom{-}0.410	&	\phantom{-}0.903	&	\phantom{-}0.100	&	\phantom{-}0.309	&	\phantom{-}0.995	&	\phantom{-}0.660	&	\phantom{-}0.955	&	-1.163	&	\phantom{-}0.135	\\ 
MetricX-23-QE-b					&	\phantom{-}0.001	&	\phantom{-}0.441	&	\phantom{-}0.836	&	\phantom{-}0.145	&	\phantom{-}0.269	&	\phantom{-}0.764	&	\phantom{-}0.527	&	\phantom{-}0.930	&	-1.123	&	\phantom{-}0.124	 \\ 
MetricX-23-QE-c					&	-0.011	&	\phantom{-}0.291	&	\phantom{-}0.609	&	-0.072	&	\phantom{-}0.169	&	\phantom{-}0.695	&	\phantom{-}0.485	&	\phantom{-}0.963	&	-0.759	&	\phantom{-}0.098	\\ 
XCOMET-QE-Ensemble					&	\phantom{-}0.080	&	\phantom{-}0.373	&	\phantom{-}0.517	&	\phantom{-}0.322	&	\phantom{-}0.196	&	\phantom{-}0.491	&	\phantom{-}0.335	&	\phantom{-}0.389	&	-0.734	&	\phantom{-}0.112	\\ 
\midrule 
Average					&	\phantom{-}0.073	&	\phantom{-}0.332	&	\phantom{-}0.355	&	\phantom{-}0.722	&	\phantom{-}0.265	&	\phantom{-}0.308	&	\phantom{-}0.143	&	\phantom{-}0.290	&	-0.266	&	\phantom{-}0.183	\\

\bottomrule
\end{tabular}}
\caption{Metric sensitivity scores (scaled by WMT scores, then avg(good - bad)) of metrics submitted to WMT 2023 for the nine top level categories in the \textsc{ACES} ontology, plus the additional fluency category: punctuation.   The horizontal lines delimit baseline metrics (top), participating reference-based metrics (middle) and participating reference-free metrics (bottom). The best result for each category is denoted by bold text with a green highlight. Note that \textit{Average} is an average over averages.}
\label{tab:analysis_overview_sensitivities_2023}
\end{sidewaystable*}
\afterpage{\clearpage}
Tables~\ref{tab:analysis_overview_sensitivities_2022} and ~\ref{tab:analysis_overview_sensitivities_2023} contain the average sensitivity scores for each high-level phenomena of the metrics submitted to WMT 2022 and WMT 2023 respectively.

\starttwocolumn
\bibliography{compling_style,anthology,custom}

\begin{thebibliography}{103}
\expandafter\ifx\csname natexlab\endcsname\relax\def\natexlab#1{#1}\fi

\bibitem[{Alves et~al.(2022)Alves, Rei, Farinha, C.~de Souza, and
  Martins}]{alves-etal-2022-robust}
Alves, Duarte, Ricardo Rei, Ana~C Farinha, Jos{\'e}~G. C.~de Souza, and
  Andr{\'e} F.~T. Martins. 2022.
\newblock Robust {MT} evaluation with sentence-level multilingual augmentation.
\newblock In \emph{Proceedings of the Seventh Conference on Machine Translation
  (WMT)}, pages 469--478, Association for Computational Linguistics, Abu Dhabi,
  United Arab Emirates (Hybrid).

\bibitem[{Amrhein, Moghe, and Guillou(2022)}]{amrhein-etal-2022-aces}
Amrhein, Chantal, Nikita Moghe, and Liane Guillou. 2022.
\newblock {ACES}: Translation accuracy challenge sets for evaluating machine
  translation metrics.
\newblock In \emph{Proceedings of the Seventh Conference on Machine Translation
  (WMT)}, pages 479--513, Association for Computational Linguistics, Abu Dhabi,
  United Arab Emirates (Hybrid).

\bibitem[{Amrhein, Moghe, and Guillou(2023)}]{amrhein-moghe-guillou:2023:WMT}
Amrhein, Chantal, Nikita Moghe, and Liane Guillou. 2023.
\newblock Aces: Translation accuracy challenge sets at wmt 2023.
\newblock In \emph{Proceedings of the Eighth Conference on Machine
  Translation}, pages 693--710, Association for Computational Linguistics,
  Singapore.

\bibitem[{Amrhein and Sennrich(2022)}]{amrhein2022identifying}
Amrhein, Chantal and Rico Sennrich. 2022.
\newblock Identifying weaknesses in machine translation metrics through minimum
  bayes risk decoding: A case study for {COMET}.
\newblock In \emph{2nd Conference of the Asia-Pacific Chapter of the
  Association for Computational Linguistics and the 12th International Joint
  Conference on Natural Language Processing}, Association for Computational
  Linguistics, Online.

\bibitem[{Avramidis and
  Macketanz(2022)}]{avramidis-macketanz-2022-linguistically}
Avramidis, Eleftherios and Vivien Macketanz. 2022.
\newblock Linguistically motivated evaluation of machine translation metrics
  based on a challenge set.
\newblock In \emph{Proceedings of the Seventh Conference on Machine Translation
  (WMT)}, pages 514--529, Association for Computational Linguistics, Abu Dhabi,
  United Arab Emirates (Hybrid).

\bibitem[{Avramidis et~al.(2018)Avramidis, Macketanz, Lommel, and
  Uszkoreit}]{avramidis-etal-2018-fine}
Avramidis, Eleftherios, Vivien Macketanz, Arle Lommel, and Hans Uszkoreit.
  2018.
\newblock Fine-grained evaluation of quality estimation for machine translation
  based on a linguistically motivated test suite.
\newblock In \emph{Proceedings of the {AMTA} 2018 Workshop on Translation
  Quality Estimation and Automatic Post-Editing}, pages 243--248, Association
  for Machine Translation in the Americas, Boston, MA.

\bibitem[{Avramidis et~al.(2023)Avramidis, Manakhimova, Macketanz, and
  M{\"o}ller}]{avramidis-etal-2023-challenging}
Avramidis, Eleftherios, Shushen Manakhimova, Vivien Macketanz, and Sebastian
  M{\"o}ller. 2023.
\newblock Challenging the state-of-the-art machine translation metrics from a
  linguistic perspective.
\newblock In \emph{Proceedings of the Eighth Conference on Machine
  Translation}, pages 713--729, Association for Computational Linguistics,
  Singapore.

\bibitem[{Bentivogli et~al.(2016)Bentivogli, Bisazza, Cettolo, and
  Federico}]{bentivogli-etal-2016-neural}
Bentivogli, Luisa, Arianna Bisazza, Mauro Cettolo, and Marcello Federico. 2016.
\newblock Neural versus phrase-based machine translation quality: a case study.
\newblock In \emph{Proceedings of the 2016 Conference on Empirical Methods in
  Natural Language Processing}, pages 257--267, Association for Computational
  Linguistics, Austin, Texas.

\bibitem[{Bojar et~al.(2016)Bojar, Chatterjee, Federmann, Graham, Haddow, Huck,
  Jimeno~Yepes, Koehn, Logacheva, Monz, Negri, N{\'e}v{\'e}ol, Neves, Popel,
  Post, Rubino, Scarton, Specia, Turchi, Verspoor, and
  Zampieri}]{bojar-etal-2016-findings}
Bojar, Ond{\v{r}}ej, Rajen Chatterjee, Christian Federmann, Yvette Graham,
  Barry Haddow, Matthias Huck, Antonio Jimeno~Yepes, Philipp Koehn, Varvara
  Logacheva, Christof Monz, Matteo Negri, Aur{\'e}lie N{\'e}v{\'e}ol, Mariana
  Neves, Martin Popel, Matt Post, Raphael Rubino, Carolina Scarton, Lucia
  Specia, Marco Turchi, Karin Verspoor, and Marcos Zampieri. 2016.
\newblock Findings of the 2016 conference on machine translation.
\newblock In \emph{Proceedings of the First Conference on Machine Translation:
  Volume 2, Shared Task Papers}, pages 131--198, Association for Computational
  Linguistics, Berlin, Germany.

\bibitem[{Brown et~al.(2020)Brown, Mann, Ryder, Subbiah, Kaplan, Dhariwal,
  Neelakantan, Shyam, Sastry, Askell, Agarwal, Herbert{-}Voss, Krueger,
  Henighan, Child, Ramesh, Ziegler, Wu, Winter, Hesse, Chen, Sigler, Litwin,
  Gray, Chess, Clark, Berner, McCandlish, Radford, Sutskever, and
  Amodei}]{DBLP:journals/corr/abs-2005-14165}
Brown, Tom~B., Benjamin Mann, Nick Ryder, Melanie Subbiah, Jared Kaplan,
  Prafulla Dhariwal, Arvind Neelakantan, Pranav Shyam, Girish Sastry, Amanda
  Askell, Sandhini Agarwal, Ariel Herbert{-}Voss, Gretchen Krueger, Tom
  Henighan, Rewon Child, Aditya Ramesh, Daniel~M. Ziegler, Jeffrey Wu, Clemens
  Winter, Christopher Hesse, Mark Chen, Eric Sigler, Mateusz Litwin, Scott
  Gray, Benjamin Chess, Jack Clark, Christopher Berner, Sam McCandlish, Alec
  Radford, Ilya Sutskever, and Dario Amodei. 2020.
\newblock Language models are few-shot learners.
\newblock \emph{CoRR}, abs/2005.14165.

\bibitem[{Campolungo et~al.(2022)Campolungo, Martelli, Saina, and
  Navigli}]{campolungo-etal-2022-dibimt}
Campolungo, Niccol{\`o}, Federico Martelli, Francesco Saina, and Roberto
  Navigli. 2022.
\newblock {D}i{B}i{MT}: A novel benchmark for measuring {W}ord {S}ense
  {D}isambiguation biases in {M}achine {T}ranslation.
\newblock In \emph{Proceedings of the 60th Annual Meeting of the Association
  for Computational Linguistics (Volume 1: Long Papers)}, pages 4331--4352,
  Association for Computational Linguistics, Dublin, Ireland.

\bibitem[{Carlini et~al.(2020)Carlini, Tram{\`e}r, Wallace, Jagielski,
  Herbert-Voss, Lee, Roberts, Brown, Song, Erlingsson, Oprea, and
  Raffel}]{Carlini2020ExtractingTD}
Carlini, Nicholas, Florian Tram{\`e}r, Eric Wallace, Matthew Jagielski, Ariel
  Herbert-Voss, Katherine Lee, Adam Roberts, Tom~B. Brown, Dawn~Xiaodong Song,
  {\'U}lfar Erlingsson, Alina Oprea, and Colin Raffel. 2020.
\newblock Extracting training data from large language models.
\newblock In \emph{USENIX Security Symposium}.

\bibitem[{Castilho et~al.(2017)Castilho, Moorkens, Gaspari, Calixto, Tinsley,
  and Way}]{article}
Castilho, Sheila, Joss Moorkens, Federico Gaspari, Iacer Calixto, John Tinsley,
  and Andy Way. 2017.
\newblock Is neural machine translation the new state of the art?
\newblock \emph{The Prague Bulletin of Mathematical Linguistics}, 108:109--120.

\bibitem[{Chen et~al.(2022)Chen, Wei, Shang, Li, Wu, Yu, Zhu, Zhu, Xie, Lei,
  Tao, Yang, and Qin}]{chen-etal-2022-exploring}
Chen, Xiaoyu, Daimeng Wei, Hengchao Shang, Zongyao Li, Zhanglin Wu, Zhengzhe
  Yu, Ting Zhu, Mengli Zhu, Ning Xie, Lizhi Lei, Shimin Tao, Hao Yang, and Ying
  Qin. 2022.
\newblock Exploring robustness of machine translation metrics: A study of
  twenty-two automatic metrics in the {WMT}22 metric task.
\newblock In \emph{Proceedings of the Seventh Conference on Machine Translation
  (WMT)}, pages 530--540, Association for Computational Linguistics, Abu Dhabi,
  United Arab Emirates (Hybrid).

\bibitem[{Chia et~al.(2023)Chia, Hong, Bing, and Poria}]{chia2023instructeval}
Chia, Yew~Ken, Pengfei Hong, Lidong Bing, and Soujanya Poria. 2023.
\newblock Instructeval: Towards holistic evaluation of instruction-tuned large
  language models.

\bibitem[{Chung et~al.(2022)Chung, Hou, Longpre, Zoph, Tay, Fedus, Li, Wang,
  Dehghani, Brahma, Webson, Gu, Dai, Suzgun, Chen, Chowdhery, Castro-Ros,
  Pellat, Robinson, Valter, Narang, Mishra, Yu, Zhao, Huang, Dai, Yu, Petrov,
  Chi, Dean, Devlin, Roberts, Zhou, Le, and Wei}]{chung2022scaling}
Chung, Hyung~Won, Le~Hou, Shayne Longpre, Barret Zoph, Yi~Tay, William Fedus,
  Yunxuan Li, Xuezhi Wang, Mostafa Dehghani, Siddhartha Brahma, Albert Webson,
  Shixiang~Shane Gu, Zhuyun Dai, Mirac Suzgun, Xinyun Chen, Aakanksha
  Chowdhery, Alex Castro-Ros, Marie Pellat, Kevin Robinson, Dasha Valter,
  Sharan Narang, Gaurav Mishra, Adams Yu, Vincent Zhao, Yanping Huang, Andrew
  Dai, Hongkun Yu, Slav Petrov, Ed~H. Chi, Jeff Dean, Jacob Devlin, Adam
  Roberts, Denny Zhou, Quoc~V. Le, and Jason Wei. 2022.
\newblock Scaling instruction-finetuned language models.

\bibitem[{Conneau et~al.(2018)Conneau, Rinott, Lample, Williams, Bowman,
  Schwenk, and Stoyanov}]{conneau-etal-2018-xnli}
Conneau, Alexis, Ruty Rinott, Guillaume Lample, Adina Williams, Samuel Bowman,
  Holger Schwenk, and Veselin Stoyanov. 2018.
\newblock {XNLI}: Evaluating cross-lingual sentence representations.
\newblock In \emph{Proceedings of the 2018 Conference on Empirical Methods in
  Natural Language Processing}, pages 2475--2485, Association for Computational
  Linguistics, Brussels, Belgium.

\bibitem[{Dale et~al.(2023)Dale, Voita, Barrault, and
  Costa-juss{\`a}}]{dale-etal-2023-detecting}
Dale, David, Elena Voita, Loic Barrault, and Marta~R. Costa-juss{\`a}. 2023.
\newblock Detecting and mitigating hallucinations in machine translation: Model
  internal workings alone do well, sentence similarity {E}ven better.
\newblock In \emph{Proceedings of the 61st Annual Meeting of the Association
  for Computational Linguistics (Volume 1: Long Papers)}, pages 36--50,
  Association for Computational Linguistics, Toronto, Canada.

\bibitem[{Devlin et~al.(2019)Devlin, Chang, Lee, and
  Toutanova}]{devlin-etal-2019-bert}
Devlin, Jacob, Ming-Wei Chang, Kenton Lee, and Kristina Toutanova. 2019.
\newblock {BERT}: Pre-training of deep bidirectional transformers for language
  understanding.
\newblock In \emph{Proceedings of the 2019 Conference of the North {A}merican
  Chapter of the Association for Computational Linguistics: Human Language
  Technologies, Volume 1 (Long and Short Papers)}, pages 4171--4186,
  Association for Computational Linguistics, Minneapolis, Minnesota.

\bibitem[{Dréano, Molloy, and
  Murphy(2023{\natexlab{a}})}]{dreano-molloy-murphy:2023:WMT2}
Dréano, Sören, Derek Molloy, and Noel Murphy. 2023{\natexlab{a}}.
\newblock Embed\_llama: Using llm embeddings for the metrics shared task.
\newblock In \emph{Proceedings of the Eighth Conference on Machine
  Translation}, pages 736--743, Association for Computational Linguistics,
  Singapore.

\bibitem[{Dréano, Molloy, and
  Murphy(2023{\natexlab{b}})}]{dreano-molloy-murphy:2023:WMT1}
Dréano, Sören, Derek Molloy, and Noel Murphy. 2023{\natexlab{b}}.
\newblock Tokengram\_f, a fast and accurate token-based chrf++ derivative.
\newblock In \emph{Proceedings of the Eighth Conference on Machine
  Translation}, pages 728--735, Association for Computational Linguistics,
  Singapore.

\bibitem[{Dziri et~al.(2023)Dziri, Lu, Sclar, Li, Jian, Lin, West, Bhagavatula,
  Bras, Hwang, Sanyal, Welleck, Ren, Ettinger, Harchaoui, and
  Choi}]{Dziri2023FaithAF}
Dziri, Nouha, Ximing Lu, Melanie Sclar, Xiang~Lorraine Li, Liwei Jian,
  Bill~Yuchen Lin, Peter West, Chandra Bhagavatula, Ronan~Le Bras, Jena~D.
  Hwang, Soumya Sanyal, Sean Welleck, Xiang Ren, Allyson Ettinger, Za{\"i}d
  Harchaoui, and Yejin Choi. 2023.
\newblock Faith and fate: Limits of transformers on compositionality.
\newblock \emph{ArXiv}, abs/2305.18654.

\bibitem[{ElNokrashy and Kocmi(2023)}]{elnokrashy-kocmi:2023:WMT}
ElNokrashy, Muhammad and Tom Kocmi. 2023.
\newblock ebleu: Unexpectedly good machine translation evaluation using simple
  word embeddings.
\newblock In \emph{Proceedings of the Eighth Conference on Machine
  Translation}, pages 744--748, Association for Computational Linguistics,
  Singapore.

\bibitem[{Emelin and Sennrich(2021)}]{emelin-sennrich-2021-wino}
Emelin, Denis and Rico Sennrich. 2021.
\newblock Wino-{X}: Multilingual {W}inograd schemas for commonsense reasoning
  and coreference resolution.
\newblock In \emph{Proceedings of the 2021 Conference on Empirical Methods in
  Natural Language Processing}, pages 8517--8532, Association for Computational
  Linguistics, Online and Punta Cana, Dominican Republic.

\bibitem[{Fan et~al.(2021)Fan, Bhosale, Schwenk, Ma, El-Kishky, Goyal, Baines,
  Celebi, Wenzek, Chaudhary, Goyal, Birch, Liptchinsky, Edunov, Auli, and
  Joulin}]{fan2021beyond}
Fan, Angela, Shruti Bhosale, Holger Schwenk, Zhiyi Ma, Ahmed El-Kishky,
  Siddharth Goyal, Mandeep Baines, Onur Celebi, Guillaume Wenzek, Vishrav
  Chaudhary, Naman Goyal, Tom Birch, Vitaliy Liptchinsky, Sergey Edunov,
  Michael Auli, and Armand Joulin. 2021.
\newblock Beyond english-centric multilingual machine translation.
\newblock \emph{Journal of Machine Learning Research}, 22(107):1--48.

\bibitem[{Fernandes et~al.(2023)Fernandes, Deutsch, Finkelstein, Riley,
  Martins, Neubig, Garg, Clark, Freitag, and Firat}]{fernandes-EtAl:2023:WMT}
Fernandes, Patrick, Daniel Deutsch, Mara Finkelstein, Parker Riley, Andr{\'e}
  Martins, Graham Neubig, Ankush Garg, Jonathan Clark, Markus Freitag, and
  Orhan Firat. 2023.
\newblock The devil is in the errors: Leveraging large language models for
  fine-grained machine translation evaluation.
\newblock In \emph{Proceedings of the Eighth Conference on Machine
  Translation}, pages 1066--1083, Association for Computational Linguistics,
  Singapore.

\bibitem[{Freitag et~al.(2021{\natexlab{a}})Freitag, Foster, Grangier,
  Ratnakar, Tan, and Macherey}]{freitag-etal-2021-experts}
Freitag, Markus, George Foster, David Grangier, Viresh Ratnakar, Qijun Tan, and
  Wolfgang Macherey. 2021{\natexlab{a}}.
\newblock Experts, errors, and context: A large-scale study of human evaluation
  for machine translation.
\newblock \emph{Transactions of the Association for Computational Linguistics},
  9:1460--1474.

\bibitem[{Freitag et~al.(2023)Freitag, Mathur, Lo, Avramidis, Rei, Thompson,
  Kocmi, Blain, Deutsch, Stewart, Zerva, Castilho, Lavie, and
  Foster}]{freitag-etal-2023-results}
Freitag, Markus, Nitika Mathur, Chi-kiu Lo, Eleftherios Avramidis, Ricardo Rei,
  Brian Thompson, Tom Kocmi, Frederic Blain, Daniel Deutsch, Craig Stewart,
  Chrysoula Zerva, Sheila Castilho, Alon Lavie, and George Foster. 2023.
\newblock Results of {WMT}23 metrics shared task: Metrics might be guilty but
  references are not innocent.
\newblock In \emph{Proceedings of the Eighth Conference on Machine
  Translation}, pages 578--628, Association for Computational Linguistics,
  Singapore.

\bibitem[{Freitag et~al.(2022)Freitag, Rei, Mathur, Lo, Stewart, Avramidis,
  Kocmi, Foster, Lavie, and Martins}]{freitag-etal-2022-results}
Freitag, Markus, Ricardo Rei, Nitika Mathur, Chi-kiu Lo, Craig Stewart,
  Eleftherios Avramidis, Tom Kocmi, George Foster, Alon Lavie, and Andr{\'e}
  F.~T. Martins. 2022.
\newblock Results of {WMT}22 metrics shared task: Stop using {BLEU} {--} neural
  metrics are better and more robust.
\newblock In \emph{Proceedings of the Seventh Conference on Machine Translation
  (WMT)}, pages 46--68, Association for Computational Linguistics, Abu Dhabi,
  United Arab Emirates (Hybrid).

\bibitem[{Freitag et~al.(2021{\natexlab{b}})Freitag, Rei, Mathur, Lo, Stewart,
  Foster, Lavie, and Bojar}]{freitag-etal-2021-results}
Freitag, Markus, Ricardo Rei, Nitika Mathur, Chi-kiu Lo, Craig Stewart, George
  Foster, Alon Lavie, and Ond{\v{r}}ej Bojar. 2021{\natexlab{b}}.
\newblock Results of the {WMT}21 metrics shared task: Evaluating metrics with
  expert-based human evaluations on {TED} and news domain.
\newblock In \emph{Proceedings of the Sixth Conference on Machine Translation},
  pages 733--774, Association for Computational Linguistics, Online.

\bibitem[{Gowda, Kocmi, and
  Junczys-Dowmunt(2023)}]{gowda-kocmi-junczysdowmunt:2023:WMT}
Gowda, Thamme, Tom Kocmi, and Marcin Junczys-Dowmunt. 2023.
\newblock Cometoid: Distilling strong reference-based machine translation
  metrics into even stronger quality estimation metrics.
\newblock In \emph{Proceedings of the Eighth Conference on Machine
  Translation}, pages 749--753, Association for Computational Linguistics,
  Singapore.

\bibitem[{Goyal et~al.(2021)Goyal, Du, Ott, Anantharaman, and
  Conneau}]{DBLP:journals/corr/abs-2105-00572}
Goyal, Naman, Jingfei Du, Myle Ott, Giri Anantharaman, and Alexis Conneau.
  2021.
\newblock Larger-scale transformers for multilingual masked language modeling.
\newblock pages 29--33.

\bibitem[{Goyal et~al.(2022)Goyal, Gao, Chaudhary, Chen, Wenzek, Ju, Krishnan,
  Ranzato, Guzm{\'a}n, and Fan}]{goyal-etal-2022-flores}
Goyal, Naman, Cynthia Gao, Vishrav Chaudhary, Peng-Jen Chen, Guillaume Wenzek,
  Da~Ju, Sanjana Krishnan, Marc{'}Aurelio Ranzato, Francisco Guzm{\'a}n, and
  Angela Fan. 2022.
\newblock The {F}lores-101 evaluation benchmark for low-resource and
  multilingual machine translation.
\newblock \emph{Transactions of the Association for Computational Linguistics},
  10:522--538.

\bibitem[{Guerreiro et~al.(2023)Guerreiro, Rei, van Stigt, Coheur, Colombo, and
  Martins}]{guerreiro2023xcomet}
Guerreiro, Nuno~M., Ricardo Rei, Daan van Stigt, Luisa Coheur, Pierre Colombo,
  and André F.~T. Martins. 2023.
\newblock xcomet: Transparent machine translation evaluation through
  fine-grained error detection.

\bibitem[{Guillou and Hardmeier(2016)}]{guillou-hardmeier-2016-protest}
Guillou, Liane and Christian Hardmeier. 2016.
\newblock {PROTEST}: A test suite for evaluating pronouns in machine
  translation.
\newblock In \emph{Proceedings of the Tenth International Conference on
  Language Resources and Evaluation ({LREC}'16)}, pages 636--643, European
  Language Resources Association (ELRA), Portoro{\v{z}}, Slovenia.

\bibitem[{Guillou et~al.(2018)Guillou, Hardmeier, Lapshinova-Koltunski, and
  Lo{\'a}iciga}]{guillou-etal-2018-pronoun}
Guillou, Liane, Christian Hardmeier, Ekaterina Lapshinova-Koltunski, and Sharid
  Lo{\'a}iciga. 2018.
\newblock A pronoun test suite evaluation of the {E}nglish{--}{G}erman {MT}
  systems at {WMT} 2018.
\newblock In \emph{Proceedings of the Third Conference on Machine Translation:
  Shared Task Papers}, pages 570--577, Association for Computational
  Linguistics, Belgium, Brussels.

\bibitem[{Hanna and Bojar(2021)}]{hanna-bojar-2021-fine}
Hanna, Michael and Ond{\v{r}}ej Bojar. 2021.
\newblock A fine-grained analysis of {BERTS}core.
\newblock In \emph{Proceedings of the Sixth Conference on Machine Translation},
  pages 507--517, Association for Computational Linguistics, Online.

\bibitem[{Isabelle, Cherry, and Foster(2017)}]{isabelle-etal-2017-challenge}
Isabelle, Pierre, Colin Cherry, and George Foster. 2017.
\newblock A challenge set approach to evaluating machine translation.
\newblock In \emph{Proceedings of the 2017 Conference on Empirical Methods in
  Natural Language Processing}, pages 2486--2496, Association for Computational
  Linguistics, Copenhagen, Denmark.

\bibitem[{Jia and Liang(2017)}]{jia-liang-2017-adversarial}
Jia, Robin and Percy Liang. 2017.
\newblock Adversarial examples for evaluating reading comprehension systems.
\newblock In \emph{Proceedings of the 2017 Conference on Empirical Methods in
  Natural Language Processing}, pages 2021--2031, Association for Computational
  Linguistics, Copenhagen, Denmark.

\bibitem[{Juraska et~al.(2023)Juraska, Finkelstein, Deutsch, Siddhant,
  Mirzazadeh, and Freitag}]{juraska-EtAl:2023:WMT}
Juraska, Juraj, Mara Finkelstein, Daniel Deutsch, Aditya Siddhant, Mehdi
  Mirzazadeh, and Markus Freitag. 2023.
\newblock Metricx-23: The google submission to the wmt 2023 metrics shared
  task.
\newblock In \emph{Proceedings of the Eighth Conference on Machine
  Translation}, pages 754--765, Association for Computational Linguistics,
  Singapore.

\bibitem[{Karpinska et~al.(2022)Karpinska, Raj, Thai, Song, Gupta, and
  Iyyer}]{karpinska-etal-2022-demetr}
Karpinska, Marzena, Nishant Raj, Katherine Thai, Yixiao Song, Ankita Gupta, and
  Mohit Iyyer. 2022.
\newblock {DEMETR}: Diagnosing evaluation metrics for translation.
\newblock In \emph{Proceedings of the 2022 Conference on Empirical Methods in
  Natural Language Processing}, pages 9540--9561, Association for Computational
  Linguistics, Abu Dhabi, United Arab Emirates.

\bibitem[{Khashabi et~al.(2018)Khashabi, Chaturvedi, Roth, Upadhyay, and
  Roth}]{khashabi-etal-2018-looking}
Khashabi, Daniel, Snigdha Chaturvedi, Michael Roth, Shyam Upadhyay, and Dan
  Roth. 2018.
\newblock Looking beyond the surface: A challenge set for reading comprehension
  over multiple sentences.
\newblock In \emph{Proceedings of the 2018 Conference of the North {A}merican
  Chapter of the Association for Computational Linguistics: Human Language
  Technologies, Volume 1 (Long Papers)}, pages 252--262, Association for
  Computational Linguistics, New Orleans, Louisiana.

\bibitem[{King and Falkedal(1990)}]{king-falkedal-1990-using}
King, Margaret and Kirsten Falkedal. 1990.
\newblock Using test suites in evaluation of machine translation systems.
\newblock In \emph{{COLING} 1990 Volume 2: Papers presented to the 13th
  International Conference on Computational Linguistics}.

\bibitem[{Kocmi and Federmann(2023{\natexlab{a}})}]{kocmi-federmann:2023:WMT}
Kocmi, Tom and Christian Federmann. 2023{\natexlab{a}}.
\newblock Gemba-mqm: Detecting translation quality error spans with gpt-4.
\newblock In \emph{Proceedings of the Eighth Conference on Machine
  Translation}, pages 766--773, Association for Computational Linguistics,
  Singapore.

\bibitem[{Kocmi and Federmann(2023{\natexlab{b}})}]{kocmi-federmann-2023-large}
Kocmi, Tom and Christian Federmann. 2023{\natexlab{b}}.
\newblock Large language models are state-of-the-art evaluators of translation
  quality.
\newblock In \emph{Proceedings of the 24th Annual Conference of the European
  Association for Machine Translation}, pages 193--203, European Association
  for Machine Translation, Tampere, Finland.

\bibitem[{Kocmi et~al.(2021)Kocmi, Federmann, Grundkiewicz, Junczys-Dowmunt,
  Matsushita, and Menezes}]{kocmi-etal-2021-ship}
Kocmi, Tom, Christian Federmann, Roman Grundkiewicz, Marcin Junczys-Dowmunt,
  Hitokazu Matsushita, and Arul Menezes. 2021.
\newblock To ship or not to ship: An extensive evaluation of automatic metrics
  for machine translation.
\newblock In \emph{Proceedings of the Sixth Conference on Machine Translation},
  pages 478--494, Association for Computational Linguistics, Online.

\bibitem[{Kocmi, Matsushita, and Federmann(2022)}]{MS-COMET:WMT22}
Kocmi, Tom, Hitokazu Matsushita, and Christian Federmann. 2022.
\newblock {MS-COMET}: {M}ore and {B}etter {H}uman {J}udgements {I}mprove
  {M}etric {P}erformance.
\newblock In \emph{Proceedings of the Seventh Conference on Machine
  Translation}, Association for Computational Linguistics, Abu Dhabi.

\bibitem[{Koehn(2005)}]{koehn-2005-europarl}
Koehn, Philipp. 2005.
\newblock {E}uroparl: A parallel corpus for statistical machine translation.
\newblock In \emph{Proceedings of Machine Translation Summit X: Papers}, pages
  79--86, Phuket, Thailand.

\bibitem[{Koehn and Monz(2006)}]{koehn-monz-2006-manual}
Koehn, Philipp and Christof Monz. 2006.
\newblock Manual and automatic evaluation of machine translation between
  {E}uropean languages.
\newblock In \emph{Proceedings on the Workshop on Statistical Machine
  Translation}, pages 102--121, Association for Computational Linguistics, New
  York City.

\bibitem[{Kudo and Richardson(2018)}]{kudo-richardson-2018-sentencepiece}
Kudo, Taku and John Richardson. 2018.
\newblock {S}entence{P}iece: A simple and language independent subword
  tokenizer and detokenizer for neural text processing.
\newblock In \emph{Proceedings of the 2018 Conference on Empirical Methods in
  Natural Language Processing: System Demonstrations}, pages 66--71,
  Association for Computational Linguistics, Brussels, Belgium.

\bibitem[{Laali and Kosseim(2017)}]{laali-kosseim-2017-improving}
Laali, Majid and Leila Kosseim. 2017.
\newblock Improving discourse relation projection to build discourse annotated
  corpora.
\newblock In \emph{Proceedings of the International Conference Recent Advances
  in Natural Language Processing, {RANLP} 2017}, pages 407--416, INCOMA Ltd.,
  Varna, Bulgaria.

\bibitem[{Lapshinova-Koltunski, Hardmeier, and
  Krielke(2018)}]{lapshinova-koltunski-etal-2018-parcorfull}
Lapshinova-Koltunski, Ekaterina, Christian Hardmeier, and Pauline Krielke.
  2018.
\newblock {P}ar{C}or{F}ull: a parallel corpus annotated with full coreference.
\newblock In \emph{Proceedings of the Eleventh International Conference on
  Language Resources and Evaluation ({LREC} 2018)}, European Language Resources
  Association (ELRA), Miyazaki, Japan.

\bibitem[{Li, Cohn, and Baldwin(2017)}]{li-etal-2017-bibi}
Li, Yitong, Trevor Cohn, and Timothy Baldwin. 2017.
\newblock {BIBI} system description: Building with {CNN}s and breaking with
  deep reinforcement learning.
\newblock In \emph{Proceedings of the First Workshop on Building Linguistically
  Generalizable {NLP} Systems}, pages 27--32, Association for Computational
  Linguistics, Copenhagen, Denmark.

\bibitem[{Liu et~al.(2022)Liu, Qiao, Wu, Chang, Zhang, Zhao, Song~Peng, Yang,
  Qin, Guo, Wang, Li, Li, and Zhao}]{HWTSC-Metrics:WMT22}
Liu, Yilun, Xiaosong Qiao, Zhanglin Wu, Su~Chang, Min Zhang, Yanqing Zhao,
  shimin~tao Song~Peng, Hao Yang, Ying Qin, Jiaxin Guo, Minghan Wang, Yinglu
  Li, Peng Li, and Xiaofeng Zhao. 2022.
\newblock {P}artial {C}ould {B}e {B}etter {T}han {W}hole: {HW-TSC} 2022
  {S}ubmission for the {M}etrics {S}hared {T}ask.
\newblock In \emph{Proceedings of the Seventh Conference on Machine
  Translation}, Association for Computational Linguistics, Abu Dhabi.

\bibitem[{Lo(2019)}]{lo-2019-yisi}
Lo, Chi-kiu. 2019.
\newblock {Y}i{S}i - a unified semantic {MT} quality evaluation and estimation
  metric for languages with different levels of available resources.
\newblock In \emph{Proceedings of the Fourth Conference on Machine Translation
  (Volume 2: Shared Task Papers, Day 1)}, pages 507--513, Association for
  Computational Linguistics, Florence, Italy.

\bibitem[{Lo, Larkin, and Knowles(2023)}]{lo-larkin-knowles:2023:WMT}
Lo, Chi-kiu, Samuel Larkin, and Rebecca Knowles. 2023.
\newblock Metric score landscape challenge (mslc23): Understanding metrics'
  performance on a wider landscape of translation quality.
\newblock In \emph{Proceedings of the Eighth Conference on Machine
  Translation}, pages 774--797, Association for Computational Linguistics,
  Singapore.

\bibitem[{Lommel, Burchardt, and Uszkoreit(2014)}]{lommel2014}
Lommel, Arle, Aljoscha Burchardt, and Hans Uszkoreit. 2014.
\newblock Multidimensional quality metrics (mqm): A framework for declaring and
  describing translation quality metrics.
\newblock \emph{Tradumàtica: tecnologies de la traducció}, 0:455--463.

\bibitem[{Lu et~al.(2023)Lu, Qiu, Ding, Xie, and Tao}]{lu2023error}
Lu, Qingyu, Baopu Qiu, Liang Ding, Liping Xie, and Dacheng Tao. 2023.
\newblock Error analysis prompting enables human-like translation evaluation in
  large language models: A case study on chatgpt.

\bibitem[{Mahler et~al.(2017)Mahler, Cheung, Elsner, King, de~Marneffe, Shain,
  Stevens-Guille, and White}]{mahler-etal-2017-breaking}
Mahler, Taylor, Willy Cheung, Micha Elsner, David King, Marie-Catherine
  de~Marneffe, Cory Shain, Symon Stevens-Guille, and Michael White. 2017.
\newblock Breaking {NLP}: Using morphosyntax, semantics, pragmatics and world
  knowledge to fool sentiment analysis systems.
\newblock In \emph{Proceedings of the First Workshop on Building Linguistically
  Generalizable {NLP} Systems}, pages 33--39, Association for Computational
  Linguistics, Copenhagen, Denmark.

\bibitem[{McCoy and Linzen(2019)}]{mccoy2019non}
McCoy, Richard~T and Tal Linzen. 2019.
\newblock Non-entailed subsequences as a challenge for natural language
  inference.
\newblock \emph{Proceedings of the Society for Computation in Linguistics
  (SCiL)}, pages 358--360.

\bibitem[{Moghe et~al.(2023)Moghe, Sherborne, Steedman, and
  Birch}]{moghe-etal-2023-extrinsic}
Moghe, Nikita, Tom Sherborne, Mark Steedman, and Alexandra Birch. 2023.
\newblock Extrinsic evaluation of machine translation metrics.
\newblock In \emph{Proceedings of the 61st Annual Meeting of the Association
  for Computational Linguistics (Volume 1: Long Papers)}, pages 13060--13078,
  Association for Computational Linguistics, Toronto, Canada.

\bibitem[{Mukherjee and Shrivastava(2022)}]{REUSE:WMT22}
Mukherjee, Ananya and Manish Shrivastava. 2022.
\newblock {REUSE}: {RE}ference-free {U}n{S}upervised quality {E}stimation
  {M}etric.
\newblock In \emph{Proceedings of the Seventh Conference on Machine
  Translation}, Association for Computational Linguistics, Abu Dhabi.

\bibitem[{Mukherjee and Shrivastava(2023)}]{mukherjee-shrivastava:2023:WMT2}
Mukherjee, Ananya and Manish Shrivastava. 2023.
\newblock Mee4 and xlsim : Iiit hyd's submissions' for wmt23 metrics shared
  task.
\newblock In \emph{Proceedings of the Eighth Conference on Machine
  Translation}, pages 798--803, Association for Computational Linguistics,
  Singapore.

\bibitem[{Neubig(2022)}]{neubig2022is}
Neubig, Graham. 2022.
\newblock Is my nlp model working? the answer is harder than you think.

\bibitem[{{NLLB Team} et~al.(2022){NLLB Team}, Costa-jussà, Cross, Çelebi,
  Elbayad, Heafield, Heffernan, Kalbassi, Lam, Licht, Maillard, Sun, Wang,
  Wenzek, Youngblood, Akula, Barrault, Gonzalez, Hansanti, Hoffman, Jarrett,
  Sadagopan, Rowe, Spruit, Tran, Andrews, Ayan, Bhosale, Edunov, Fan, Gao,
  Goswami, Guzmán, Koehn, Mourachko, Ropers, Saleem, Schwenk, and
  Wang}]{flores-200}
{NLLB Team}, Marta~R. Costa-jussà, James Cross, Onur Çelebi, Maha Elbayad,
  Kenneth Heafield, Kevin Heffernan, Elahe Kalbassi, Janice Lam, Daniel Licht,
  Jean Maillard, Anna Sun, Skyler Wang, Guillaume Wenzek, Al~Youngblood, Bapi
  Akula, Loic Barrault, Gabriel~Mejia Gonzalez, Prangthip Hansanti, John
  Hoffman, Semarley Jarrett, Kaushik~Ram Sadagopan, Dirk Rowe, Shannon Spruit,
  Chau Tran, Pierre Andrews, Necip~Fazil Ayan, Shruti Bhosale, Sergey Edunov,
  Angela Fan, Cynthia Gao, Vedanuj Goswami, Francisco Guzmán, Philipp Koehn,
  Alexandre Mourachko, Christophe Ropers, Safiyyah Saleem, Holger Schwenk, and
  Jeff Wang. 2022.
\newblock No language left behind: Scaling human-centered machine translation.

\bibitem[{Papineni et~al.(2002)Papineni, Roukos, Ward, and
  Zhu}]{papineni-etal-2002-bleu}
Papineni, Kishore, Salim Roukos, Todd Ward, and Wei-Jing Zhu. 2002.
\newblock {B}leu: a method for automatic evaluation of machine translation.
\newblock In \emph{Proceedings of the 40th Annual Meeting of the Association
  for Computational Linguistics}, pages 311--318, Association for Computational
  Linguistics, Philadelphia, Pennsylvania, USA.

\bibitem[{Perrella et~al.(2022{\natexlab{a}})Perrella, Proietti, Scir{\`e},
  Campolungo, and Navigli}]{perrella-etal-2022-matese}
Perrella, Stefano, Lorenzo Proietti, Alessandro Scir{\`e}, Niccol{\`o}
  Campolungo, and Roberto Navigli. 2022{\natexlab{a}}.
\newblock {M}a{TES}e: Machine translation evaluation as a sequence tagging
  problem.
\newblock In \emph{Proceedings of the Seventh Conference on Machine Translation
  (WMT)}, pages 569--577, Association for Computational Linguistics, Abu Dhabi,
  United Arab Emirates (Hybrid).

\bibitem[{Perrella et~al.(2022{\natexlab{b}})Perrella, Proietti, Scirè,
  Campolungo, and Navigli}]{MATESE:WMT22}
Perrella, Stefano, Lorenzo Proietti, Alessandro Scirè, Niccolò Campolungo,
  and Roberto Navigli. 2022{\natexlab{b}}.
\newblock {M}achine {T}ranslation {E}valuation as a {S}equence {T}agging
  {P}roblem.
\newblock In \emph{Proceedings of the Seventh Conference on Machine
  Translation}, Association for Computational Linguistics, Abu Dhabi.

\bibitem[{Popovi{\'c}(2017)}]{popovic-2017-chrf}
Popovi{\'c}, Maja. 2017.
\newblock chr{F}++: words helping character n-grams.
\newblock In \emph{Proceedings of the Second Conference on Machine
  Translation}, pages 612--618, Association for Computational Linguistics,
  Copenhagen, Denmark.

\bibitem[{Popovi{\'c} and Castilho(2019)}]{popovic-castilho-2019-challenge}
Popovi{\'c}, Maja and Sheila Castilho. 2019.
\newblock Challenge test sets for {MT} evaluation.
\newblock In \emph{Proceedings of Machine Translation Summit XVII: Tutorial
  Abstracts}, European Association for Machine Translation, Dublin, Ireland.

\bibitem[{Raganato, Scherrer, and Tiedemann(2019)}]{raganato-etal-2019-mucow}
Raganato, Alessandro, Yves Scherrer, and J{\"o}rg Tiedemann. 2019.
\newblock The {M}u{C}o{W} test suite at {WMT} 2019: Automatically harvested
  multilingual contrastive word sense disambiguation test sets for machine
  translation.
\newblock In \emph{Proceedings of the Fourth Conference on Machine Translation
  (Volume 2: Shared Task Papers, Day 1)}, pages 470--480, Association for
  Computational Linguistics, Florence, Italy.

\bibitem[{Ravichander et~al.(2021)Ravichander, Dalmia, Ryskina, Metze, Hovy,
  and Black}]{ravichander-etal-2021-noiseqa}
Ravichander, Abhilasha, Siddharth Dalmia, Maria Ryskina, Florian Metze, Eduard
  Hovy, and Alan~W Black. 2021.
\newblock {N}oise{QA}: Challenge set evaluation for user-centric question
  answering.
\newblock In \emph{Proceedings of the 16th Conference of the European Chapter
  of the Association for Computational Linguistics: Main Volume}, pages
  2976--2992, Association for Computational Linguistics, Online.

\bibitem[{Rei et~al.(2023)Rei, Guerreiro, Treviso, Coheur, Lavie, and
  Martins}]{rei-etal-2023-inside}
Rei, Ricardo, Nuno~M. Guerreiro, Marcos Treviso, Luisa Coheur, Alon Lavie, and
  Andr{\'e} Martins. 2023.
\newblock The inside story: Towards better understanding of machine translation
  neural evaluation metrics.
\newblock In \emph{Proceedings of the 61st Annual Meeting of the Association
  for Computational Linguistics (Volume 2: Short Papers)}, pages 1089--1105,
  Association for Computational Linguistics, Toronto, Canada.

\bibitem[{Rei et~al.(2022)Rei, de~Souza, Alves, Zerva, Farinha, Glushkova,
  Lavie, Coheur, and Martins}]{COMET:WMT22}
Rei, Ricardo, José G.~C. de~Souza, Duarte Alves, Chrysoula Zerva, Ana~C
  Farinha, Taisiya Glushkova, Alon Lavie, Luisa Coheur, and André F.~T.
  Martins. 2022.
\newblock {COMET-22:} {U}nbabel-{IST} 2022 {S}ubmission for the {M}etrics
  {S}hared {T}ask.
\newblock In \emph{Proceedings of the Seventh Conference on Machine
  Translation}, Association for Computational Linguistics, Abu Dhabi.

\bibitem[{Rei et~al.(2020)Rei, Stewart, Farinha, and
  Lavie}]{rei-etal-2020-comet}
Rei, Ricardo, Craig Stewart, Ana~C Farinha, and Alon Lavie. 2020.
\newblock {COMET}: A neural framework for {MT} evaluation.
\newblock In \emph{Proceedings of the 2020 Conference on Empirical Methods in
  Natural Language Processing (EMNLP)}, pages 2685--2702, Association for
  Computational Linguistics, Online.

\bibitem[{Rimell, Clark, and Steedman(2009)}]{rimell-etal-2009-unbounded}
Rimell, Laura, Stephen Clark, and Mark Steedman. 2009.
\newblock Unbounded dependency recovery for parser evaluation.
\newblock In \emph{Proceedings of the 2009 Conference on Empirical Methods in
  Natural Language Processing}, pages 813--821, Association for Computational
  Linguistics, Singapore.

\bibitem[{Rios, M{\"u}ller, and Sennrich(2018)}]{rios-etal-2018-word}
Rios, Annette, Mathias M{\"u}ller, and Rico Sennrich. 2018.
\newblock The word sense disambiguation test suite at {WMT}18.
\newblock In \emph{Proceedings of the Third Conference on Machine Translation:
  Shared Task Papers}, pages 588--596, Association for Computational
  Linguistics, Belgium, Brussels.

\bibitem[{Rocchietti et~al.(2021)Rocchietti, Achena, Marziano, Salaris, and
  Lenci}]{Rocchietti2021FANCYAD}
Rocchietti, Guido, Flavia Achena, Giuseppe Marziano, Sara Salaris, and
  Alessandro Lenci. 2021.
\newblock Fancy: A diagnostic data-set for nli models.
\newblock In \emph{Proceedings of the Eighth Italian Conference on
  Computational Linguistics (CLiC-it)}.

\bibitem[{Rudinger, May, and Van~Durme(2017)}]{rudinger-etal-2017-social}
Rudinger, Rachel, Chandler May, and Benjamin Van~Durme. 2017.
\newblock Social bias in elicited natural language inferences.
\newblock In \emph{Proceedings of the First {ACL} Workshop on Ethics in Natural
  Language Processing}, pages 74--79, Association for Computational
  Linguistics, Valencia, Spain.

\bibitem[{Scao et~al.(2022)Scao, Fan, Akiki, Pavlick, Ilic, Hesslow,
  Castagn{\'{e}}, Luccioni, Yvon, Gall{\'{e}}, Tow, Rush, Biderman, Webson,
  Ammanamanchi, Wang, Sagot, Muennighoff, del Moral, Ruwase, Bawden, Bekman,
  McMillan{-}Major, Beltagy, Nguyen, Saulnier, Tan, Suarez, Sanh,
  Lauren{\c{c}}on, Jernite, Launay, Mitchell, Raffel, Gokaslan, Simhi, Soroa,
  Aji, Alfassy, Rogers, Nitzav, Xu, Mou, Emezue, Klamm, Leong, van Strien,
  Adelani, and et~al.}]{DBLP:journals/corr/abs-2211-05100}
Scao, Teven~Le, Angela Fan, Christopher Akiki, Ellie Pavlick, Suzana Ilic,
  Daniel Hesslow, Roman Castagn{\'{e}}, Alexandra~Sasha Luccioni,
  Fran{\c{c}}ois Yvon, Matthias Gall{\'{e}}, Jonathan Tow, Alexander~M. Rush,
  Stella Biderman, Albert Webson, Pawan~Sasanka Ammanamanchi, Thomas Wang,
  Beno{\^{\i}}t Sagot, Niklas Muennighoff, Albert~Villanova del Moral, Olatunji
  Ruwase, Rachel Bawden, Stas Bekman, Angelina McMillan{-}Major, Iz~Beltagy,
  Huu Nguyen, Lucile Saulnier, Samson Tan, Pedro~Ortiz Suarez, Victor Sanh,
  Hugo Lauren{\c{c}}on, Yacine Jernite, Julien Launay, Margaret Mitchell, Colin
  Raffel, Aaron Gokaslan, Adi Simhi, Aitor Soroa, Alham~Fikri Aji, Amit
  Alfassy, Anna Rogers, Ariel~Kreisberg Nitzav, Canwen Xu, Chenghao Mou, Chris
  Emezue, Christopher Klamm, Colin Leong, Daniel van Strien, David~Ifeoluwa
  Adelani, and et~al. 2022.
\newblock {BLOOM:} {A} 176b-parameter open-access multilingual language model.
\newblock \emph{CoRR}, abs/2211.05100.

\bibitem[{Sellam et~al.(2020)Sellam, Pu, Chung, Gehrmann, Tan, Freitag, Das,
  and Parikh}]{sellam-etal-2020-learning}
Sellam, Thibault, Amy Pu, Hyung~Won Chung, Sebastian Gehrmann, Qijun Tan,
  Markus Freitag, Dipanjan Das, and Ankur Parikh. 2020.
\newblock Learning to evaluate translation beyond {E}nglish: {BLEURT}
  submissions to the {WMT} metrics 2020 shared task.
\newblock In \emph{Proceedings of the Fifth Conference on Machine Translation},
  pages 921--927, Association for Computational Linguistics, Online.

\bibitem[{Sennrich, Haddow, and Birch(2016)}]{sennrich-etal-2016-neural}
Sennrich, Rico, Barry Haddow, and Alexandra Birch. 2016.
\newblock Neural machine translation of rare words with subword units.
\newblock In \emph{Proceedings of the 54th Annual Meeting of the Association
  for Computational Linguistics (Volume 1: Long Papers)}, pages 1715--1725,
  Association for Computational Linguistics, Berlin, Germany.

\bibitem[{Sinha et~al.(2021)Sinha, Jia, Hupkes, Pineau, Williams, and
  Kiela}]{sinha-etal-2021-masked}
Sinha, Koustuv, Robin Jia, Dieuwke Hupkes, Joelle Pineau, Adina Williams, and
  Douwe Kiela. 2021.
\newblock Masked language modeling and the distributional hypothesis: Order
  word matters pre-training for little.
\newblock In \emph{Proceedings of the 2021 Conference on Empirical Methods in
  Natural Language Processing}, pages 2888--2913, Association for Computational
  Linguistics, Online and Punta Cana, Dominican Republic.

\bibitem[{Smith(2012)}]{DBLP:journals/corr/abs-1207-0245}
Smith, Noah~A. 2012.
\newblock Adversarial evaluation for models of natural language.
\newblock \emph{CoRR}, abs/1207.0245.

\bibitem[{Specia et~al.(2020)Specia, Li, Pino, Chaudhary, Guzm{\'a}n, Neubig,
  Durrani, Belinkov, Koehn, Sajjad, Michel, and Li}]{specia-etal-2020-findings}
Specia, Lucia, Zhenhao Li, Juan Pino, Vishrav Chaudhary, Francisco Guzm{\'a}n,
  Graham Neubig, Nadir Durrani, Yonatan Belinkov, Philipp Koehn, Hassan Sajjad,
  Paul Michel, and Xian Li. 2020.
\newblock Findings of the {WMT} 2020 shared task on machine translation
  robustness.
\newblock In \emph{Proceedings of the Fifth Conference on Machine Translation},
  pages 76--91, Association for Computational Linguistics, Online.

\bibitem[{Stali{\=u}nait{\.e} and
  Bonfil(2017)}]{staliunaite-bonfil-2017-breaking}
Stali{\=u}nait{\.e}, Ieva and Ben Bonfil. 2017.
\newblock Breaking sentiment analysis of movie reviews.
\newblock In \emph{Proceedings of the First Workshop on Building Linguistically
  Generalizable {NLP} Systems}, pages 61--64, Association for Computational
  Linguistics, Copenhagen, Denmark.

\bibitem[{Stanovsky, Smith, and
  Zettlemoyer(2019)}]{stanovsky-etal-2019-evaluating}
Stanovsky, Gabriel, Noah~A. Smith, and Luke Zettlemoyer. 2019.
\newblock Evaluating gender bias in machine translation.
\newblock In \emph{Proceedings of the 57th Annual Meeting of the Association
  for Computational Linguistics}, pages 1679--1684, Association for
  Computational Linguistics, Florence, Italy.

\bibitem[{Tao et~al.(2022)Tao, Chang, Miaomiao, Yang, Geng, Huang, Zhang, Guo,
  Wang, and Li}]{tao-etal-2022-crossqe}
Tao, Shimin, Su~Chang, Ma~Miaomiao, Hao Yang, Xiang Geng, Shujian Huang, Min
  Zhang, Jiaxin Guo, Minghan Wang, and Yinglu Li. 2022.
\newblock {C}ross{QE}: {HW}-{TSC} 2022 submission for the quality estimation
  shared task.
\newblock In \emph{Proceedings of the Seventh Conference on Machine Translation
  (WMT)}, pages 646--652, Association for Computational Linguistics, Abu Dhabi,
  United Arab Emirates (Hybrid).

\bibitem[{Taori et~al.(2023)Taori, Gulrajani, Zhang, Dubois, Li, Guestrin,
  Liang, and Hashimoto}]{alpaca}
Taori, Rohan, Ishaan Gulrajani, Tianyi Zhang, Yann Dubois, Xuechen Li, Carlos
  Guestrin, Percy Liang, and Tatsunori~B. Hashimoto. 2023.
\newblock Stanford alpaca: An instruction-following llama model.
\newblock \url{https://github.com/tatsu-lab/stanford_alpaca}.

\bibitem[{Toral and
  S{\'a}nchez-Cartagena(2017)}]{toral-sanchez-cartagena-2017-multifaceted}
Toral, Antonio and V{\'\i}ctor~M. S{\'a}nchez-Cartagena. 2017.
\newblock A multifaceted evaluation of neural versus phrase-based machine
  translation for 9 language directions.
\newblock In \emph{Proceedings of the 15th Conference of the {E}uropean Chapter
  of the Association for Computational Linguistics: Volume 1, Long Papers},
  pages 1063--1073, Association for Computational Linguistics, Valencia, Spain.

\bibitem[{Touvron et~al.(2023)Touvron, Martin, Stone, Albert, Almahairi,
  Babaei, Bashlykov, Batra, Bhargava, Bhosale, Bikel, Blecher, Ferrer, Chen,
  Cucurull, Esiobu, Fernandes, Fu, Fu, Fuller, Gao, Goswami, Goyal, Hartshorn,
  Hosseini, Hou, Inan, Kardas, Kerkez, Khabsa, Kloumann, Korenev, Koura,
  Lachaux, Lavril, Lee, Liskovich, Lu, Mao, Martinet, Mihaylov, Mishra,
  Molybog, Nie, Poulton, Reizenstein, Rungta, Saladi, Schelten, Silva, Smith,
  Subramanian, Tan, Tang, Taylor, Williams, Kuan, Xu, Yan, Zarov, Zhang, Fan,
  Kambadur, Narang, Rodriguez, Stojnic, Edunov, and Scialom}]{llama2}
Touvron, Hugo, Louis Martin, Kevin Stone, Peter Albert, Amjad Almahairi,
  Yasmine Babaei, Nikolay Bashlykov, Soumya Batra, Prajjwal Bhargava, Shruti
  Bhosale, Dan Bikel, Lukas Blecher, Cristian~Canton Ferrer, Moya Chen, Guillem
  Cucurull, David Esiobu, Jude Fernandes, Jeremy Fu, Wenyin Fu, Brian Fuller,
  Cynthia Gao, Vedanuj Goswami, Naman Goyal, Anthony Hartshorn, Saghar
  Hosseini, Rui Hou, Hakan Inan, Marcin Kardas, Viktor Kerkez, Madian Khabsa,
  Isabel Kloumann, Artem Korenev, Punit~Singh Koura, Marie-Anne Lachaux,
  Thibaut Lavril, Jenya Lee, Diana Liskovich, Yinghai Lu, Yuning Mao, Xavier
  Martinet, Todor Mihaylov, Pushkar Mishra, Igor Molybog, Yixin Nie, Andrew
  Poulton, Jeremy Reizenstein, Rashi Rungta, Kalyan Saladi, Alan Schelten, Ruan
  Silva, Eric~Michael Smith, Ranjan Subramanian, Xiaoqing~Ellen Tan, Binh Tang,
  Ross Taylor, Adina Williams, Jian~Xiang Kuan, Puxin Xu, Zheng Yan, Iliyan
  Zarov, Yuchen Zhang, Angela Fan, Melanie Kambadur, Sharan Narang, Aurelien
  Rodriguez, Robert Stojnic, Sergey Edunov, and Thomas Scialom. 2023.
\newblock Llama 2: Open foundation and fine-tuned chat models.

\bibitem[{Vamvas and Sennrich(2021)}]{vamvas-sennrich-2021-contrastive}
Vamvas, Jannis and Rico Sennrich. 2021.
\newblock Contrastive conditioning for assessing disambiguation in {MT}: {A}
  case study of distilled bias.
\newblock In \emph{Proceedings of the 2021 Conference on Empirical Methods in
  Natural Language Processing}, pages 10246--10265, Association for
  Computational Linguistics, Online and Punta Cana, Dominican Republic.

\bibitem[{Vamvas and Sennrich(2022)}]{vamvas-sennrich-2022-little}
Vamvas, Jannis and Rico Sennrich. 2022.
\newblock As little as possible, as much as necessary: Detecting over- and
  undertranslations with contrastive conditioning.
\newblock In \emph{Proceedings of the 60th Annual Meeting of the Association
  for Computational Linguistics (Volume 2: Short Papers)}, pages 490--500,
  Association for Computational Linguistics, Dublin, Ireland.

\bibitem[{Vieira, O'Hagan, and
  O'Sullivan(2021)}]{doi:10.1080/1369118X.2020.1776370}
Vieira, Lucas~Nunes, Minako O'Hagan, and Carol O'Sullivan. 2021.
\newblock Understanding the societal impacts of machine translation: a critical
  review of the literature on medical and legal use cases.
\newblock \emph{Information, Communication \& Society}, 24(11):1515--1532.

\bibitem[{Wan et~al.(2022{\natexlab{a}})Wan, Bao, Liu, Yang, Wong, Chao, Lei,
  and Xie}]{UNITE:WMT22}
Wan, Yu, Keqin Bao, Dayiheng Liu, Baosong Yang, Derek~F. Wong, Lidia~S. Chao,
  Wenqiang Lei, and Jun Xie. 2022{\natexlab{a}}.
\newblock {A}libaba-{T}ranslate {C}hina's {S}ubmission for {WMT2022} {M}etrics
  {S}hared {T}ask.
\newblock In \emph{Proceedings of the Seventh Conference on Machine
  Translation}, Association for Computational Linguistics, Abu Dhabi.

\bibitem[{Wan et~al.(2022{\natexlab{b}})Wan, Liu, Yang, Zhang, Chen, Wong, and
  Chao}]{wan-etal-2022-unite}
Wan, Yu, Dayiheng Liu, Baosong Yang, Haibo Zhang, Boxing Chen, Derek Wong, and
  Lidia Chao. 2022{\natexlab{b}}.
\newblock {U}ni{TE}: Unified translation evaluation.
\newblock In \emph{Proceedings of the 60th Annual Meeting of the Association
  for Computational Linguistics (Volume 1: Long Papers)}, pages 8117--8127,
  Association for Computational Linguistics, Dublin, Ireland.

\bibitem[{Wu and Dredze(2019)}]{wu-dredze-2019-beto}
Wu, Shijie and Mark Dredze. 2019.
\newblock Beto, bentz, becas: The surprising cross-lingual effectiveness of
  {BERT}.
\newblock In \emph{Proceedings of the 2019 Conference on Empirical Methods in
  Natural Language Processing and the 9th International Joint Conference on
  Natural Language Processing (EMNLP-IJCNLP)}, pages 833--844, Association for
  Computational Linguistics, Hong Kong, China.

\bibitem[{Wu et~al.(2023)Wu, Liu, Zhang, Zhao, Zhu, Zhu, Qiao, Zhang, Miaomiao,
  Yanqing, Peng, tao, Yang, and Jiang}]{wu-EtAl:2023:WMT4}
Wu, Zhanglin, Yilun Liu, Min Zhang, Xiaofeng Zhao, Junhao Zhu, Ming Zhu,
  Xiaosong Qiao, Jingfei Zhang, Ma~Miaomiao, Zhao Yanqing, Song Peng, shimin
  tao, Hao Yang, and Yanfei Jiang. 2023.
\newblock Empowering a metric with llm-assisted named entity annotation:
  Hw-tsc's submission to the wmt23 metrics shared task.
\newblock In \emph{Proceedings of the Eighth Conference on Machine
  Translation}, pages 820--826, Association for Computational Linguistics,
  Singapore.

\bibitem[{Xu et~al.(2023)Xu, Wang, Pan, Song, Freitag, Wang, and
  Li}]{xu-etal-2023-instructscore}
Xu, Wenda, Danqing Wang, Liangming Pan, Zhenqiao Song, Markus Freitag, William
  Wang, and Lei Li. 2023.
\newblock {INSTRUCTSCORE}: Towards explainable text generation evaluation with
  automatic feedback.
\newblock In \emph{Proceedings of the 2023 Conference on Empirical Methods in
  Natural Language Processing}, pages 5967--5994, Association for Computational
  Linguistics, Singapore.

\bibitem[{Yang et~al.(2019)Yang, Zhang, Tar, and
  Baldridge}]{yang-etal-2019-paws}
Yang, Yinfei, Yuan Zhang, Chris Tar, and Jason Baldridge. 2019.
\newblock {PAWS}-{X}: A cross-lingual adversarial dataset for paraphrase
  identification.
\newblock In \emph{Proceedings of the 2019 Conference on Empirical Methods in
  Natural Language Processing and the 9th International Joint Conference on
  Natural Language Processing (EMNLP-IJCNLP)}, pages 3687--3692, Association
  for Computational Linguistics, Hong Kong, China.

\bibitem[{Zhang et~al.(2020)Zhang, Kishore, Wu, Weinberger, and
  Artzi}]{DBLP:conf/iclr/ZhangKWWA20}
Zhang, Tianyi, Varsha Kishore, Felix Wu, Kilian~Q. Weinberger, and Yoav Artzi.
  2020.
\newblock Bertscore: Evaluating text generation with {BERT}.
\newblock In \emph{8th International Conference on Learning Representations,
  {ICLR} 2020, Addis Ababa, Ethiopia, April 26-30, 2020}, OpenReview.net.

\bibitem[{Zhao et~al.(2018)Zhao, Wang, Yatskar, Ordonez, and
  Chang}]{zhao-etal-2018-gender}
Zhao, Jieyu, Tianlu Wang, Mark Yatskar, Vicente Ordonez, and Kai-Wei Chang.
  2018.
\newblock Gender bias in coreference resolution: Evaluation and debiasing
  methods.
\newblock In \emph{Proceedings of the 2018 Conference of the North {A}merican
  Chapter of the Association for Computational Linguistics: Human Language
  Technologies, Volume 2 (Short Papers)}, pages 15--20, Association for
  Computational Linguistics, New Orleans, Louisiana.

\bibitem[{Zhou, Gong, and Bhat(2021)}]{zhou-etal-2021-pie}
Zhou, Jianing, Hongyu Gong, and Suma Bhat. 2021.
\newblock {PIE}: A parallel idiomatic expression corpus for idiomatic sentence
  generation and paraphrasing.
\newblock In \emph{Proceedings of the 17th Workshop on Multiword Expressions
  (MWE 2021)}, pages 33--48, Association for Computational Linguistics, Online.

\end{thebibliography}


\end{document}